\documentclass[journal,twoside, final]{IEEEtran}
\usepackage{tabularx}
\newcolumntype{L}[1]{>{\raggedright\arraybackslash}p{#1}} % linksbündig mit Breitenangabe
\newcolumntype{C}[1]{>{\centering\arraybackslash}p{#1}} % zentriert mit Breitenangabe
\newcolumntype{R}[1]{>{\raggedleft\arraybackslash}p{#1}} % rechtsbündig mit Breitenangabe

\ifCLASSINFOpdf
\else
\fi
\usepackage{array}

\usepackage{import, graphicx}
\usepackage{xcolor}
\usepackage{cite}
\usepackage{subcaption}
\usepackage{amsmath, amssymb}
\usepackage{mathtools}
\usepackage{bm}
\usepackage{bbm}
\usepackage{hhline}
\usepackage{makecell}

\usepackage{multirow}

\DeclareMathOperator*{\argmin}{argmin}
\DeclareMathOperator*{\placeholder}{\,\cdot\,}
\newcommand{\newDatasetName}{SplicedColorPipeline}
\newcommand{\noisep}{NP}
\newcommand{\ffn}{FFN}
\newcommand{\forsim}{FS}
\newcommand{\lcr}{LCR}

\newcommand{\edited}[1]{{\color{black}{#1}}}
\newcommand{\reedited}[1]{{\color{black}{#1}}}

\begin{document}
%
% paper title
% Titles are generally capitalized except for words such as a, an, and, as,
% at, but, by, for, in, nor, of, on, or, the, to and up, which are usually
% not capitalized unless they are the first or last word of the title.
% Linebreaks \\ can be used within to get better formatting as desired.
% Do not put math or special symbols in the title.
%\title{Exposing Forgeries In Low Quality Images\\With Deep Metric Color Features}
%\title{Robust Forgery Localization In Severely Degraded Images With Deep Metric Color Features}
\title{Deep Metric Color Embeddings for Splicing Localization in Severely Degraded Images}
%
%
% author names and IEEE memberships
% note positions of commas and nonbreaking spaces ( ~ ) LaTeX will not break
% a structure at a ~ so this keeps an author's name from being broken across
% two lines.
% use \thanks{} to gain access to the first footnote area
% a separate \thanks must be used for each paragraph as LaTeX2e's \thanks
% was not built to handle multiple paragraphs
%

\author{Benjamin~Hadwiger, Christian~Riess,~\IEEEmembership{Senior~Member,~IEEE}}
\maketitle

% As a general rule, do not put math, special symbols or citations
% in the abstract or keywords.
\begin{abstract}
One common task in image forensics is to detect spliced images, where multiple
source images are composed to one output image.
Most of the currently best performing splicing detectors leverage high-frequency artifacts. % that carry information about the processing history of an image. 
%To cope with the vast amounts of multimedia content shared on social media every day, online platforms usually post-process images during upload to reduce file sizes, using e.g.\ JPEG compression and downsampling.
However, after an image underwent strong compression, most of the high
frequency artifacts are not available anymore.
%While largely maintaining the visual quality of an image, these operations often heavily attenuate the visually imperceptible signal components. In this way, the performance of most existing forensics algorithms severely deteriorates on post-processed images.
%

In this work, we explore an alternative approach to splicing detection, which
is potentially better suited for images in-the-wild, subject to strong
compression and downsampling. Our proposal is to model the color formation of
an image. The color formation largely depends on variations at the scale of
scene objects, and is hence much less dependent on high-frequency artifacts.
We learn a deep metric space that is on one hand sensitive to illumination
color and camera white-point estimation, but on the other hand insensitive to
variations in object color. Large distances in the embedding space indicate
that two image regions either stem from different scenes or different cameras.
In our evaluation, we show that the proposed embedding space outperforms the
state of the art on images that have been subject to strong compression and
downsampling. We confirm in two further experiments the dual nature of the
metric space, namely to both characterize the acquisition camera and the scene
illuminant color. As such, this work resides at the intersection of
physics-based and statistical forensics with benefits from both sides.  
%, both on existing benchmark datasets and on a newly crafted test dataset.
%
%We also show that the embeddings can be directly marginalized to two very
%This confirms that the embeddings %are indeed closely related to
%indeed actively embrace 
%color formation traces. 
\end{abstract}

% Note that keywords are not normally used for peerreview papers.

% For peer review papers, you can put extra information on the cover
% page as needed:
% \ifCLASSOPTIONpeerreview
% \begin{center} \bfseries EDICS Category: 3-BBND \end{center}
% \fi
%
% For peerreview papers, this IEEEtran command inserts a page break and
% creates the second title. It will be ignored for other modes.
\IEEEpeerreviewmaketitle

\section{Introduction}
% The very first letter is a 2 line initial drop letter followed
% by the rest of the first word in caps.
% 
% form to use if the first word consists of a single letter:
% \IEEEPARstart{A}{demo} file is ....
% 
% form to use if you need the single drop letter followed by
% normal text (unknown if ever used by the IEEE):
% \IEEEPARstart{A}{}demo file is ....
% 
% Some journals put the first two words in caps:
% \IEEEPARstart{T}{his demo} file is ....
% 
% Here we have the typical use of a "T" for an initial drop letter
% and "HIS" in caps to complete the first word.
\IEEEPARstart{I}{mage}-based communication became commonplace with the rise of
social networks and the broad availability of network-connected acquisition
devices.
Images effectively transport messages and emotions, and are widely accepted as
proof for an event.
%
%Among millions of images are uploaded on the internet, ranging from personal memories, such as holiday photographs, to images reaching a broad audience, for instance in commercials, news reports, or political campaigns.
%
%
However, sophisticated image processing software makes it increasingly easy to
realistically manipulate images. Manipulations with malicious intent change the
message of an image, with the purpose to deceive the viewer.

One important goal of image forensics is to develop methods for detecting image manipulations.
Over the past two decades, various methods have been proposed for exposing image forgeries.
A general overview can be found in recent books and
surveys~\cite{piva2013overview, farid2016photo, verdoliva2020media}. 

\begin{figure}[t]
	\centering
	%\resizebox{\linewidth}{!}{\import{figures/}{turtle_splices.pdf}}\\[2.0ex]
	%\resizebox{\linewidth}{!}{\import{figures/}{example_splices_persons_var_quals_hor.pdf_tex}}
	%\includegraphics[width=\linewidth]{figures/example_splices_var_quals_hor.pdf}\\[1.0ex]	%\includegraphics[width=\linewidth]{figures/example_splices_persons_var_quals_hor.pdf}
	\includegraphics[width=\linewidth]{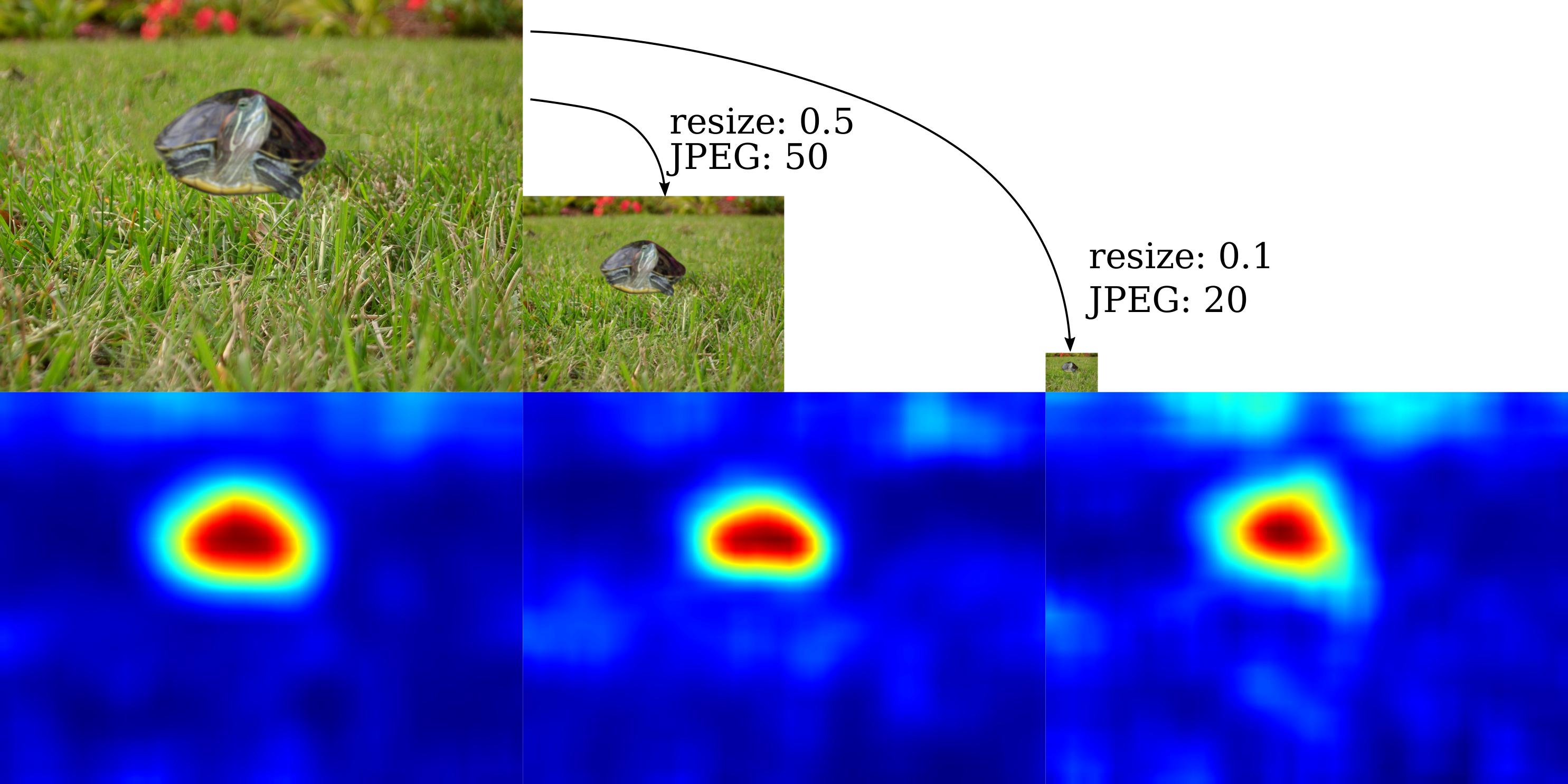}
	\caption{Spliced tortoise (manipulation from NIMBLE 2017
	dataset), and detection result under challenging downsampling and
	compression factors as they may occur during distribution over the
	internet. The detection heatmaps in the bottom row are barely affected even
	by JPEG quality 20 and downsampling to 10\% of the original size.}
	\label{fig:example_analyses}
	
\end{figure} 

Most of the current research concentrates on statistical methods. Examples
are traces from the digital image formation, such as fixed-pattern sensor noise~\cite{DBLP:conf/sswmc/LukasFG06} or noise residuals~\cite{DBLP:conf/icassp/MayerS18, DBLP:journals/tifs/CozzolinoV20}.
Other statistical methods search for potential tampering cues such a double JPEG compression~\cite{amerini2017localization, barni2017aligned} or artifacts from computer-generated images~\cite{lyu20deepfake, xuan2019generalization, li2020identification, barni2020cnn, bonettini2020use}.
%{\color{magenta}{tactical citations: Siwei Lyu, Milano, Siena, Firenze}. Siwei Lyu: Deepfakes~\cite{yang2019exposing, lyu20deepfake}. Jiwu Huang: GAN-generated image detection~\cite{li2020identification}. Bo Peng: Optimized illuminant estimation from faces~\cite{peng2016optimized}, perspective constraints~\cite{peng2017image}, GAN generated images~\cite{xuan2019generalization}. Irene Amerini: Localization of Double JPEG Compression~\cite{amerini2017localization}. Mauro Barni: GAN generated face image detection~\cite{barni2020cnn}, adversary-aware CNN for low-level manipulation detection~\cite{barni2019adversarial}. Mauro Barni, Paolo Bestagini: Double JPEG detection with CNNs~\cite{barni2017aligned}. Bestagini, Tubaro: GAN-generated face image detection~\cite{bonettini2020use}.}
Another branch is physics-based forensics, which validates for example the
lighting direction~\cite{johnson2007exposing, DBLP:conf/wifs/KeeF10,
peng2016optimized, DBLP:journals/tifs/MaternRS20}, lighting
color~\cite{DBLP:journals/tifs/CarvalhoRAPR13, DBLP:conf/cvpr/StantonHM19,
DBLP:conf/icassp/HadwigerBPR19}, shadows~\cite{DBLP:journals/tifs/LiuCDG11,
DBLP:journals/tog/KeeOF13}, or perspective constraints~\cite{Iuliani2015Image,
peng2017image, yang2019exposing}.

Both families of approaches have unique advantages and limitations. Statistical
methods achieve unparalleled performance on high-quality images, while
physics-based methods can oftentimes better process strongly compressed or
downsampled low-quality images. Statistical methods have the added benefit
to allow batch processing on images with arbitrary content, while
physics-based methods oftentimes require specific image
content~\cite{DBLP:conf/wifs/KeeF10, DBLP:journals/tifs/CarvalhoRAPR13}, and
they also oftentimes require manual annotations by an
analyst~\cite{DBLP:journals/tifs/LiuCDG11, DBLP:journals/tog/KeeOF13,
DBLP:conf/wifs/KeeF10}.

In this work, we combine concepts from statistical and physics-based methods to
harvest benefits from both. We propose a forensic descriptor to characterize
the color formation in an image. Color is impacted by scene objects,
illumination, and the in-camera processing pipeline. We present a metric space
that co-locates similar illumination and in-camera processing artifacts while
largely ignoring scene objects.

The distances in this space can be used to detect image splices.  One concrete
example is
shown in Fig.~\ref{fig:example_analyses}, where the tortoise is spliced in the
image (top). This insertion can be detected on a heatmap from our proposed
method (bottom). The proposed method shares with physics-based methods the
benefit of being remarkably robust to compression and downsampling, which is
illustrated by the three variants of the same picture in horizontal direction
and their associated heatmaps. At the same time, the method can run in a fully
automated fashion, and is as such less dependent on user input than classical
physics-based methods.

This work was preceded by a conference paper from our group on supervised
learning of color differences~\cite{DBLP:conf/icassp/HadwigerBPR19}. That
previous work used a complicated acquisition protocol for ground-truth training
data with a Macbeth color checker to train a regression network. In contrast,
this work proposes a \emph{metric space} without any need for Macbeth ground
truth data.  This enables the use of significantly more data for training with
a much larger diversity. The resulting method performs considerably better, and
outperforms the state of the art on strongly compressed images.

\begin{figure}[tb]
	\centering
	\includegraphics[width=\linewidth]{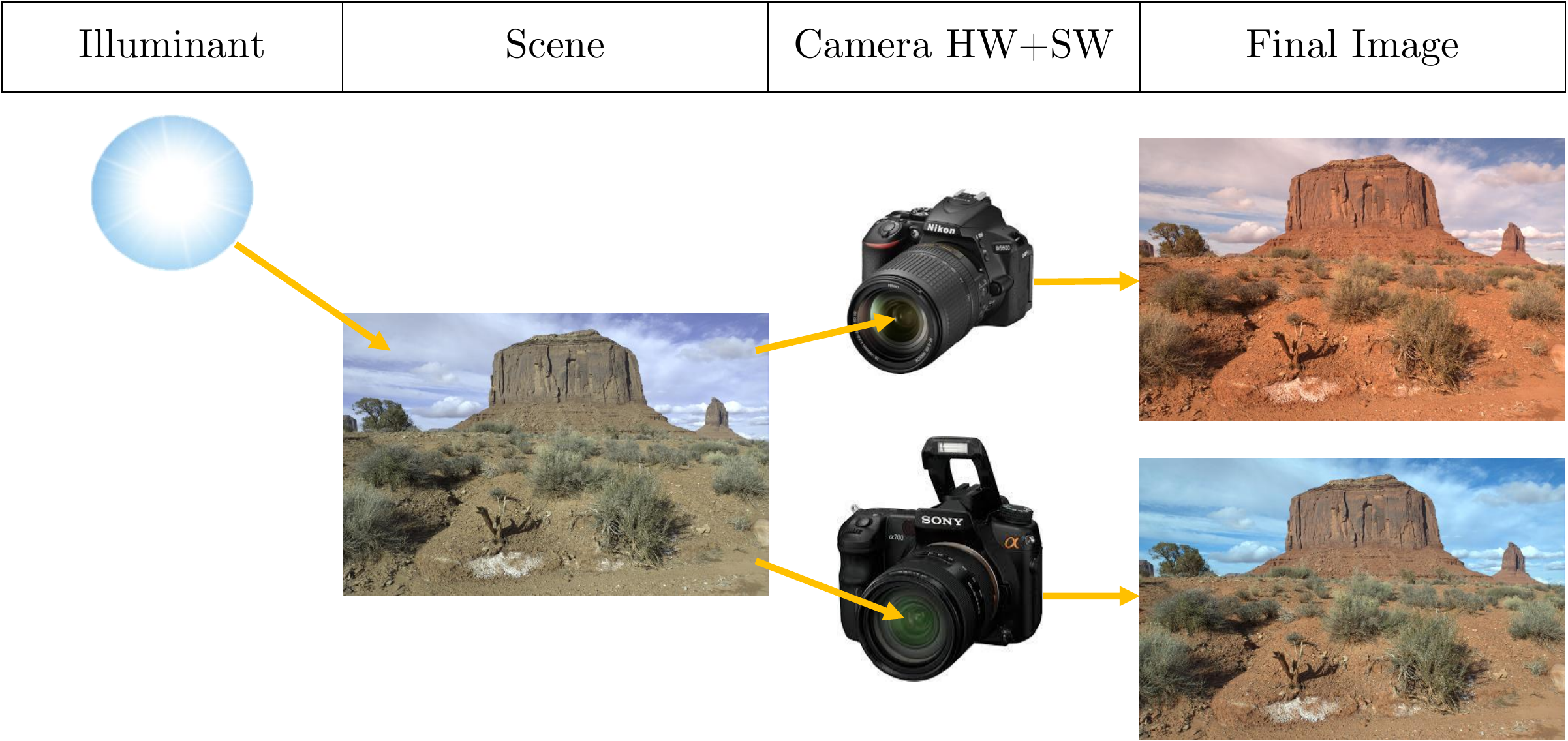}
	\caption{Illustration of the color image formation. Image colors are determined by the scene illuminant, object reflectances, and the camera color processing in hardware (HW) and software (SW). Our goal is to learn invariance to object reflectances, while preserving sensitivity to differences in illumination and camera processing.
%	Thus, different cameras yield different color images even for identical scenes. RAW image from~\cite{DBLP:conf/cvpr/BychkovskyPCD11}, processed as described in Sec.~\ref{sec:data_simulation}.
}
	\label{fig:color_imaging_pipeline}

\end{figure}

%In this work, we propose a method to learn both robust and versatile forensic features drawing inspiration from camera- and illuminant-based works.

%Our approach builds on the observation that an image depends not only on the illumination properties and content of the imaged scene, but also the camera pipeline that recorded it~\cite{DBLP:conf/icassp/HadwigerBPR19}. In this way, different images of identical scenes usually diverge, depending on the color processing applied in the cameras, see Fig.~\ref{fig:color_imaging_pipeline}. 

%More specifically, we propose a deep metric learning approach for embeddings with distances representative of the consistency of the color imaging conditions. Forgeries are exposed by locating remote image regions in this metric space. We create images of identical scenes and different color pipelines for training and evaluation. Then, we demonstrate robustness of the learned embeddings against compression, and probe our algorithm for various compression strengths on existing benchmarks and a newly created dataset. Finally, we investigate the information extracted by the network. In detail, our contributions are as follows.
The main contributions of this work are:
\begin{itemize}
%	\item a method to create images of identical scenes imaged with different color pipelines for training and validation
	\item A practical, scalable method to learn \edited{distances in a metric space that represent the similarity of color imaging conditions.}
	%a mapping from image patches into a metric space.
	%The distances in that space represent the similarity of color imaging
	%conditions.
	\item An extensive performance evaluation that demonstrates the usefulness
	of this color representation for strongly compressed and downsampled
	images, where other forensic methods are oftentimes challenged.
	\item Evaluations on white-point estimation and camera model identification to demonstrate that the learned feature space generalizes to camera identification and illuminant estimation.
%	These learned color embeddings are demonstrated to be robust against JPEG compression and resizing. When applied to splicing detection and localization, the proposed embeddings outperform the state-of-the-art for strongly compressed images from existing datasets and a newly created dataset. 
%	%\item Creating a new benchmark dataset for splicing detection and localization in which the manipulations are based on locally varying the color processing of the imaged scenes
%	%\item Outperforming the state-of-the-art for splicing detection and localization for manipulations explicitly involving the camera pipeline, and generally for strong compressions
%	\item We show applicability to open-set camera classification and competitive performance for white-point estimation
\end{itemize}

The paper is organized as follows: in Section~\ref{sec:related_work}, we review
the related literature.  Section~\ref{sec:proposed_method} presents the
theoretical background, our proposed method, and its application to exposing
image manipulations. In Section~\ref{sec:experimental_results}, we evaluate the
robustness of the learned embeddings, benchmark our algorithm for forgery
detection and localization, and further analyze the transferability of the
learned features to other vision tasks. Section~\ref{sec:conclusion} concludes
this work.

\section{Related Work}
\label{sec:related_work}
Many forensic approaches aim to expose image manipulations by validating a
particular consistency property. This property is assumed to be satisfied by
pristine images. A violation of this property is considered as an indicator of
manipulation. In this spirit, this work proposes a learned color-based
descriptor to establish a consistency criterion that jointly includes camera
properties, camera settings, and scene properties. We hence review in this
section two directions of related works, namely on camera forensics and
color-based forensics.

\subsection{Forensic Characterization of the Acquisition Device}

%One group of methods assumes the consistency of camera traces within pristine images.
The first forensic criteria on camera consistency were formed via analytic
models of specific processing steps in the imaging pipeline.
For example, manipulations can be exposed by examining color filter array
interpolation~\cite{popescu2005exposing} or lateral chromatic aberration, which
particularly applies to low-quality lenses~\cite{DBLP:conf/mmsec/JohnsonF06}.
Arguably the most impactful analytic approach is based on photo-response
non-uniformity (PRNU). PRNU is a fixed-noise pattern of the camera sensor that
allows for associating an image to its acquisition
device~\cite{DBLP:conf/sswmc/LukasFG06} and exposing local
manipulations~\cite{chen2008determining}.

Analytic approaches are appealing to forensic practitioners due to the inherent
explainability of observations. However, analytic approaches are difficult to
adapt to complicated processing chains. Such complications arise for example on
computational images from modern smartphones and with social network
post-processing.  As a consequence, recent methods learn consistency criteria
directly from the data.
For example, Bondi~\textit{et al.} propose a convolutional neural network (CNN)
to predict the acquisition camera from image
patches~\cite{DBLP:journals/spl/BondiBGBDT17}. The CNN is trained on a closed
set of known cameras to locate inserted patches from another camera.
For manipulation detection on an open set of cameras, Mayer and Stamm train a
Siamese network to compare image regions~\cite{DBLP:conf/icassp/MayerS18}. This
work has been further extended
to also operate on post-processed images~\cite{DBLP:journals/tifs/MayerS20}.
Cozzolino and Verdoliva propose an image-to-image mapping to obtain a camera
model fingerprint of high-frequency
artifacts~\cite{DBLP:journals/tifs/CozzolinoV20}.  This fingerprint directly
exposes statistical inconsistencies in manipulated regions.

These learning-based methods have in common that they exploit the camera
provenance of an image, and hence require training data with source camera
labels. Huh~\textit{et al.} relax the need for labeled data via a metadata-based
self-supervised learning approach~\cite{DBLP:conf/eccv/HuhLOE18}. Here,
pair-wise patch consistency is learned, and manipulations can be exposed by
aggregating all pair-wise consistencies.  The metadata enables the
network to implicitly learn a discriminator which also uses traces other than
camera provenance.
Our approach is in some sense similar to that work, but we consider the
\textit{scene provenance} of patches instead of image metadata. Additionally,
we explicitly include the effect of in-camera color processing as a consistency
criterion. Note that in-camera color processing is different from other camera
provenance-based approaches since it does not require data with provenance
labels, which potentially allows to use very large datasets for training.

\subsection{Forensic Characterization of Scene Colors}

The first forensic criteria on scene color consistency were formed via
analytic, physics-based models of the illuminant color.  Gholap and Bora
validate the presence of a global scene illuminant via the intersection of
dichromatic lines on specular objects~\cite{gholap2008illuminant}. Riess and
Angelopoulou calculate an illuminant map from local illuminant color estimates,
which can subsequently be analyzed by an expert~\cite{DBLP:conf/ih/RiessA10}.

As both methods require manual interaction, they are not applicable to a fully
automatic analysis. Carvalho~\textit{et al.} limit the analysis of illuminant
maps to human faces to automate the analysis~\cite{DBLP:journals/tifs/CarvalhoRAPR13}. 
%
%Deep learning based approaches such as the proposed one,
%however, can be fully automated and hence generally do not require user
%interaction.
%
Stanton~\textit{et al.} use a white point in the image as reference to assess the consistency of local illuminant estimates with a CNN~\cite{DBLP:conf/cvpr/StantonHM19}.
%
%As features, they form a histogram of local deviations of these two quantities for each superpixel in an image.
%
%Then, they train a CNN to decide whether an image is manipulated based on such a histogram.
%
Pomari~\textit{et al.} train a CNN for fake image classification on illuminant maps~\cite{DBLP:conf/icip/PomariRRRC18}.
Here, the manipulation is localized by propagating the classifier decision
backwards through the CNN to the input layer.  However, such discriminatively
trained algorithms require pristine~\emph{and} manipulated images for training,
which makes them susceptible to unseen manipulations. 

In an earlier conference paper, we propose a two-stage approach to expose
manipulations from inconsistent color imaging conditions, specifically from the
illuminant color and camera color
processing~\cite{DBLP:conf/icassp/HadwigerBPR19}.  A CNN is trained to locally
estimate a color descriptor and a classifier for pairwise consistency. However,
the supervised training of the CNN depends on ground-truth images with Macbeth
color charts, which are extremely tedious to acquire. Hence, the amount of
available training data is severely limited, which implicitly also limits the
method performance. 

The work in this paper significantly improves our earlier work, by approaching
color inconsistencies from a different angle: We propose an unsupervised deep
metric learning approach that \textit{disentangles illumination and
camera-based color traces from scene content using arbitrary RAW images}. This
allows to train a neural network on considerably larger datasets to improve the
performance.

As a sidenote, some computer vision methods apply related problem statements to
different applications.  For example, Lalonde and Efros investigate color
distributions of composite images to assess their perceived
realism~\cite{DBLP:conf/iccv/LalondeE07}. Gao~\textit{et al.} model the
in-camera color pipeline for visually plausible insertion of full 3-D computer
graphics models~\cite{gao2019mimicking}. Our method aims at more subtle,
imperceptible color mismatches of inserted content that might deceive the
viewer. As such, our method can potentially help to raise the bar for methods
that primarily aim at visually plausible splicings.
%these works aim at the  \emph{visually plausible} creation
%of content, whereas our proposed method 
%
%As a related note outside of the field of forensics, Gao~\textit{et al.} model
%the in-camera color pipeline to render 3-D objects into images.
%While
%this approach 
%, which could be regarded as a counter measure to our approach.
%
%scene color is also used in rendering In other 
%Outside of the field of forensics, Other works outside of forensics leverage color distributions as consistency criteria, as well.
%In~\cite{gao2019mimicking}, Gao~\textit{et al.} model the in-camera color pipeline to consistently insert objects into images, which could be regarded as a counter measure to our approach.
%However, such a synthesis requires a virtual 3D-model of the to-be-inserted object, which restricts it to special cases.
%
%Our approach, on the other hand, is directed to forensic application and more subtle, imperceptible color mismatches, as well.

%\subsection{Color Consistency Outside Forensics}

% if have a single appendix:
%\appendix[Proof of the Zonklar Equations]
% or
%\appendix  % for no appendix heading
% do not use \section anymore after \appendix, only \section*
% is possibly needed

% use appendices with more than one appendix
% then use \section to start each appendix
% you must declare a \section before using any
% \subsection or using \label (\appendices by itself
% starts a section numbered zero.)
%

\section{Proposed Method}
\label{sec:proposed_method}

This Section consists of five parts. In Sec.~\ref{sec:theo_background}, we
introduce a model for color image formation to derive our approach.
Section~\ref{sec:related_tasks} relates this model to the tasks of estimating
the illuminant color and camera identification.
Section~\ref{sec:data_simulation} introduces the creation of training and validation
data of identical scenes imaged with different color pipelines.
Section~\ref{sec:learning_metric_color_features} presents the details of how to
practically learn the metric color features.
Section~\ref{sec:application_to_manipulation_detection} presents the
bottom-up aggregation of patch-wise decisions for exposing image manipulations.

\subsection{Theoretical Background}
\label{sec:theo_background}

\paragraph{Image Formation} We denote a position in the image domain $\mathcal{X}\subset\mathbb{R}^2$ as
$\bm{x}\in\mathcal{X}$ and a wavelength in the electromagnetic spectrum
$\Lambda\subset\mathbb{R}_+$ as $\lambda\in\Lambda$.
Let $\bm{I}^{c,s}(\bm{x})$ denote
the RGB intensities of an image
$\bm{I}^{c,s}:\mathcal{X}\rightarrow\mathbb{R}_{+}^3$ at position $\bm{x}$,
where $s \in \mathcal{S}$ denotes one specific scene out of the set of all
scenes $\mathcal{S}$, and $c \in \mathcal{C}$ denotes one specific camera color
processing pipeline out of the set of all pipelines $\mathcal{C}$. Then, the
image formation is modeled as \edited{(compare~\cite{gijsenij2011computational})}

%Neglecting other influences like finite exposure times,

\begin{equation}
\label{eq:color_imaging_model}
\bm{I}^{c,s}(\bm{x}) = \bm{\Omega}_{c}\left(\int\limits_{\Lambda} e_{s}(\lambda, \bm{x})\cdot r_{s}(\lambda, \bm{x}) \cdot\bm{c}_{c}(\lambda)\, \mathrm{d}\lambda \right)\enspace.
\end{equation}

Here, the illuminant spectral density $e_s:\Lambda\times\mathcal{X}\rightarrow\mathbb{R}_+$ models the potentially spatially varying power density of the scene illuminants in dependence of $\lambda$. The term
$r_s = \tilde{r}_s \cdot \tilde{m}_s : \Lambda\times\mathcal{X}\rightarrow\mathbb{R}_+$ denotes the reflectance. It consists 
%Here, we summarize the combined geometric and spectral reflectance with 
%.
of the spatially varying spectral reflectance
$\tilde{r}_s:\Lambda\times\mathcal{X}\rightarrow\mathbb{R}_+$ that determines
what fraction of the light striking a surface is reflected for a given
$\lambda$, and $\tilde{m}_s:\mathcal{X}\rightarrow\mathbb{R}_+$ \edited{that determines the ratio of reflected light in dependence of the surface geometry.}
%The spectral distribution of light that falls onto a camera sensor is a
%combination of the illumination $e_s$ and the scene reflectance $r_s$.
%Processing 
%
%%
%% WICHTIGE ANMERKUNG, ABER VIELLEICHT SPAETER:
%%
% When recording an image, a camera captures and processes light reflected off a scene impinging on its sensor.
% Thereby, the final image is determined by the scene itself $(e_s, r_s)$,
% i.e.\ the illuminant conditions and the scene content, and the camera
% pipeline $(\bm{\Omega}_c, \bm{c}_c)$, recall
% Fig.~\ref{fig:color_imaging_pipeline}.
%
The vector-valued spectral camera sensitivity
$\bm{c}_c:\Lambda\rightarrow\mathbb{R}_+^3$ determines the fraction of light of
wavelength $\lambda$ that passes through the red, green, and blue (RGB) color
filters.
The subsequent and potentially nonlinear \emph{in-camera} color processing operations, such as white balancing or color space transformation, are modeled by the vector-valued function $\bm{\Omega}_{c}:\mathbb{R}_+^3\rightarrow\mathbb{R}_+^3$. \reedited{Note that, although the argument of $\bm{\Omega}_{c}$ depends on $\bm{x}$, the actual function $\bm{\Omega}_{c}$ is spatially invariant, and only determined by the camera itself.}
The positions $\bm{x}=(\frac{m}{M},\frac{n}{N})$ are discretized by the sensor with $m\in\{0,\dots,M-1\}$ and $n\in\{0,\dots,N-1\}$. A sampling point $\bm{x}$ yields a 3-dimensional RGB-vector $\bm{I}^{c,s}(\bm{x})\in\mathbb{R}_+^3$.

%%
%% WICHTIGE ANMERKUNG, ABER VIELLEICHT SPAETER:
%%
% Note that, $e_{s}$ and $r_{s}$ are determined by the scene $s$ only, while
% $\bm{\Omega_{c}}$ and $\bm{c}_{c}$ depend solely on the camera pipeline $c$.
% Further, generally $r_s$ and potentially $e_s$ are spatially varying and thus
% depend on $\bm{x}$, while $\bm{\Omega}_c$ and $\bm{c}_c$ are assumed to be
% image-global.
%
%
With this definition, a discretized image $\bm{I}^{c,s}$ is represented as
element of $\mathbb{R}_+^{M\times N\times 3}$.
An image patch $\bm{p}_{i}^{c,s}\in\mathbb{R}^{M^{\prime}\times N^{\prime}\times 3}$ collects all RGB-samples in a rectangular neighborhood around its center $\bm{x}_i$ with $M^{\prime}\leq M$ and $N^{\prime}\leq N$.
We denote by $\mathcal{I}^{c,s}$ the set of all indices of patches from image
$\bm{I}^{c,s}$, and use $i \in \mathcal{I}^{c,s}$ as index for individual patches.

%Note that, $e_{s}$ and $r_{s}$ are determined by the scene $s$, while
%$\bm{\Omega_{c}}$ and $\bm{c}_{c}$ are determined by the camera pipeline $c$.
%In general, $r_s$ (and to a lesser extend also $e_s$) are spatially varying,
%and thus depend on $\bm{x}$. In contrast, $\bm{\Omega}_c$ and $\bm{c}_c$ are
%assumed to be image-global.

%
%This leads us to train a CNN to learn features that allow characterizing %dissimilarities in 
%the illuminant conditions $e_s$ and the color handling within the camera $(\bm{\Omega}_c, \bm{c}_c)$ locally, i.e.\ .

%The key difference to our earlier work~\cite{DBLP:conf/icassp/HadwigerBPR19} is
%to avoid supervised learning. Instead, we propose a metric learning approach,
%which mitigates the
%difficult task of creating appropriate labeled training data: instead of
%demanding each embedding to be characteristic in itself (as required for
%supervised learning), metric learning only requires \emph{pairwise distances} of
%learned embeddings to describe the semantic similarity. This relaxation allows
%to train on more diverse datasets and thereby to improve the descriptiveness of
%the learned embeddings.

\paragraph{Learning Invariances and Covariances}
Our fundamental assumption is that a manipulated image contains parts from
multiple sources, and hence exhibits traces of different color imaging
conditions $(e_s, \bm{\Omega}_c, \bm{c}_c)$. These traces are captured in local
features that are learned by a CNN, i.e., from image patches
$\bm{p}_{i}^{c,s}$.

The proposed CNN is represented by the parametric family of functions
$f_{\bm{\theta}}:\mathcal{P}\rightarrow\mathcal{M}$ with parameters
$\bm{\theta}$.
It maps patches $\bm{p}_i^{c,s}\in\mathcal{P}\subseteq\mathbb{R}^{M^{\prime}\times N^{\prime}\times 3}$ to embeddings $\bm{y}_i^{c,s}=f_{\bm{\theta}}(\bm{p}_i^{c,s})\in\mathcal{M}\subseteq\mathbb{R}^{q}$ in a metric space $(\mathcal{M}, d)$. We use the metric $d:\mathcal{M}\times\mathcal{M}\rightarrow [0,1]$, where
\begin{equation}
\label{eq:cos_distance}
	d(\bm{y}_{i_0}, \bm{y}_{i_1}) = \frac{1}{2}\cdot\left(1-s(\bm{y}_{i_0}, \bm{y}_{i_1})\right)\enspace.
\end{equation}
Note that we simplify the notation by omitting the indices $c$ and $s$ when
they are not required for disambiguation.
In Eqn.~(\ref{eq:cos_distance}), $s:\mathcal{M}\times\mathcal{M}\rightarrow[-1,1]$
denotes the cosine similarity
\begin{equation}
\label{eq:cos_similarity}
	s(\bm{y}_{i_0}, \bm{y}_{i_1}) = \frac{\bm{y}_{i_0}^\top\bm{y}_{i_1}^{}}{\lVert\bm{y}_{i_0}\rVert\cdot\lVert\bm{y}_{i_1}\rVert}\enspace.
\end{equation}
The normalization discards differences in length, such that only the
angular distance is measured. This allows similar vectors to cluster
radially.
%For parallel (i.e. most similar) embeddings, it follows $d(\bm{y}_{i_1},\bm{y}_{i_2}) = 0$, while $d(\bm{y}_{i_1},\bm{y}_{i_2}) = 1$ indicates anti-parallelity, and $d(\bm{y}_{i_1},\bm{y}_{i_2}) = 0.5$ holds for orthogonality. 

Our goal is to learn an embedding space where the embedding distances
\mbox{$d(\bm{y}_{i_0}^{c_0,s_0}, \bm{y}_{i_1}^{c_1,s_1})$} characterize the
degree of agreement or discrepancy in two color imaging conditions $(e_{s_0},
\bm{\Omega}_{c_0}, \bm{c}_{c_0})$ and $(e_{s_1}, \bm{\Omega}_{c_1},
\bm{c}_{c_1})$. To this end, we observe that
in Eqn.~(\ref{eq:color_imaging_model}), the camera properties $(\bm{\Omega}_c,
\bm{c}_c)$ are independent from the scene $s$, and the scene properties
$(e_s,r_s)$ are independent from the camera $c$. \reedited{Using these independencies, we argue that the following two \textit{additional} conditions suffice to learn the desired embedding}:
%The the pairwise distances \mbox{$d(\bm{y}_{i_1}^{c_1,s_1}, \bm{y}_{i_2}^{c_2,s_2})$} between the embeddings are characteristic for the color imaging conditions $(e_{s_1},\bm{\Omega}_{c_1},\bm{c}_{c_1})$ and $(e_{s_2},\bm{\Omega}_{c_2},\bm{c}_{c_2})$, if the CNN
\begin{enumerate}
	\item $f_{\bm{\theta}}$ maps two patches from the same image closer than two patches from the same scene but different camera color pipelines. Formally, if $c_0, c_1 \in \mathcal{C}$, $c_0 \neq c_1$ denote two different camera color pipelines, and $s \in S$ one arbitrary scene, then
	\begin{equation}
	\label{eq:rel_dist_same_scene_diff_pipel}
		\begin{split}
%			\forall c_1\neq c_2,\, c_1,c_2\in\mathcal{C} \quad \forall s_1\in\mathcal{S}\\
			\forall i_0,i_1\in\mathcal{I}^{c_0,s} &\enspace \forall i_2\in\mathcal{I}^{c_1,s}:\\
			&\quad d(\bm{y}_{i_0}^{c_0,s},\bm{y}_{i_1}^{c_0,s}) < d(\bm{y}_{i_0}^{c_0,s},\bm{y}_{i_2}^{c_1,s})\enspace.
		\end{split}
	\end{equation}
	\item $f_{\bm{\theta}}$ maps two patches from the same image closer than two patches from different scenes regardless of the camera pipeline. Formally, if $c_0, c_1 \in \mathcal{C}$ are any two camera color pipelines, and $s_0, s_1 \in S$, $s_0 \neq s_1$ are two different scenes, then
	\begin{equation}
	\label{eq:rel_dist_diff_scene_diff_pipel}
		\begin{split}
		%	\forall c_1,c_2\in\mathcal{C}\enspace \forall s_1\neq s_2,\,  s_1,s_2\in\mathcal{S}\\
			&\forall i_0,i_1\in\mathcal{I}^{c_0,s_0} \enspace \forall i_2\in\mathcal{I}^{c_1,s_1}:\\
			&\quad\quad\quad\quad\quad d(\bm{y}_{i_0}^{c_0,s_0}, \bm{y}_{i_1}^{c_0,s_0}) < d(\bm{y}_{i_0}^{c_0,s_0}, \bm{y}_{i_2}^{c_1,s_1})\enspace.
		\end{split}
	\end{equation}
\end{enumerate}
Both of these conditions enforce small pairwise distances within the same image.
For such within-image patch pairs, the camera parameters $(\bm{\Omega}_{c}, \bm{c}_{c})$ are identical, but the spatially variant scene parameters $(e_{s}(\bm{x}),r_{s}(\bm{x}))$ differ if $\bm{x}_{i_0}\neq\bm{x}_{i_1}$.
These pairs encourage the CNN to learn an \emph{invariance} against within-scene reflectance variations $\Delta r_s(\placeholder, \Delta \bm{x})$ and within-scene illuminant variations $\Delta e_s(\placeholder, \Delta \bm{x})$.
%As illumination differences $\Delta e_s(\placeholder, \Delta \bm{x})$ typically vary more slowly than reflectance differences $\Delta r_s(\placeholder, \Delta \bm{x})$, the CNN can exploit correlations in the color appearance of different objects under similar illuminants for a particular camera pipeline. 

The inequality in Eqn.~(\ref{eq:rel_dist_same_scene_diff_pipel}) encourages large pairwise distances between patches from the same scene $s$ and different camera pipelines $c_0$ and $c_1$.
We implicitly use the within-scene invariance to reflectance $\Delta
r_s(\placeholder, \Delta\bm{x})$ due to varying image content. Hence,
Eqn.~(\ref{eq:rel_dist_same_scene_diff_pipel}) induces a \emph{covariance} with
between-camera variations of the parameters $\Delta\bm{\Omega}_{\Delta c}$, i.e., 
only the variations in the camera parameters contribute to this larger
distances.

The inequality in Eqn.~(\ref{eq:rel_dist_diff_scene_diff_pipel}) encourages
long distances between clusters of embeddings from different scenes.
Using again implicitly the within-scene invariance to reflectance $\Delta
r_s(\placeholder, \Delta\bm{x})$,
Eqn.~(\ref{eq:rel_dist_diff_scene_diff_pipel}) induces a \emph{covariance} with
respect to between-scene illuminant variations $\Delta e_{\Delta s}$ with the additional (mild) assumption that illuminant variations across scenes $\Delta e_{\Delta s}$ are
larger than illuminant variations within the same scene $\Delta
e_s(\placeholder,\Delta \bm{x})$.

In summary, both conditions together aim to learn embeddings with invariance
against within-scene illuminant and reflectance variations, and covariance with
between-scene illuminant variations and between-camera parameter variations.
These two conditions are enforced during training: \edited{they dictate the boolean ground truth matrices $\bm{B}_k$, that determine whether distances between pairs of embeddings are minimized or maximized, respectively}, as described in
Sec.~\ref{sec:learning_metric_color_features}.
The distances between these embeddings characterize discrepancies in the color
imaging conditions $(e_s, \bm{\Omega}_c, \bm{c}_c)$.

\begin{figure}[t]
	\centering
	\includegraphics[width=.8\linewidth]{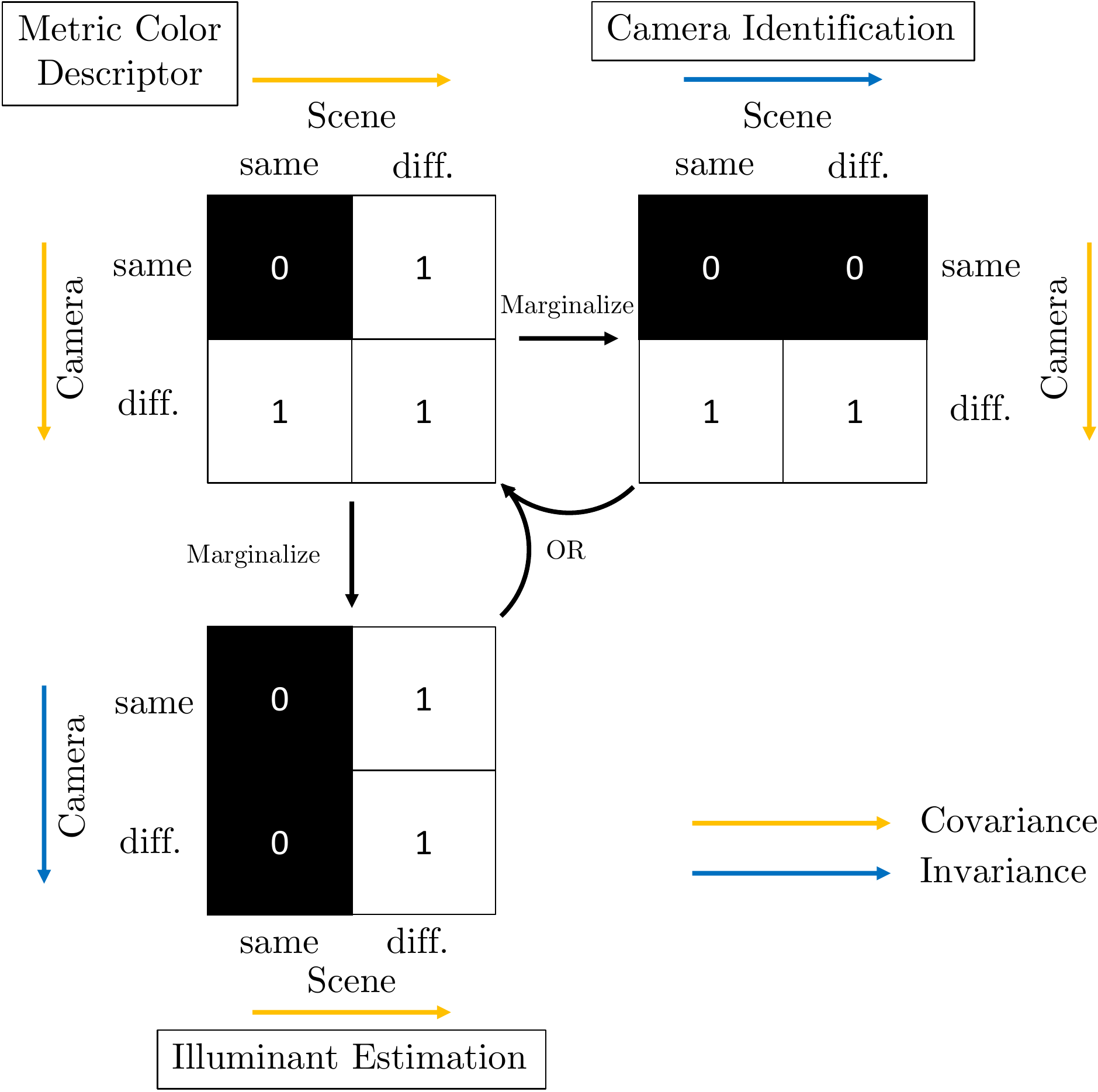}
	\caption{The proposed approach in relation to camera identification and illuminant estimation. ``0'', resp. ``1'', in the (pairwise boolean ground truth) matrices indicates that distances for the respective patch pairings are minimized, resp. maximized. The learned features inherit the covariances with respect to the camera pipeline and the scene. The tasks of camera identification and illuminant estimation can be regarded as marginalization of the proposed approach, while the proposed approach can be regarded as unification (disjunction of boolean ground truth matrices) of the two.}
	\label{fig:unification_camera_illuminant}
\end{figure}

\subsection{Relation to Camera Identification and Illuminant Color Estimation}
\label{sec:related_tasks}

The proposed metric learning approach is closely related to the tasks of camera
identification and illuminant estimation.
Camera identification requires features that only depend on the camera
(``covariance'' w.r.t.\ the camera), but not on the image content (``invariance''
w.r.t.\ the scene)~\cite{kirchner2015forensic}.
Conversely, illuminant estimation requires features that only depend on the
scene (``covariance'' w.r.t.\ the scene), but not on the camera (``invariance''
w.r.t.\ the camera)~\cite{cheng2014illuminant}.

Figure~\ref{fig:unification_camera_illuminant} illustrates that the proposed
metric color descriptor unites the covariances for both tasks.
This makes it possible to expose forgeries by discrepancies of either of the
two quantities.
At the same time, it is also possible to ``re-use'' the metric space for both
tasks individually: features for camera identification and illuminant
estimation can be obtained via marginalization of the space, by replacing one
covariance with an invariance. We experimentally demonstrate this property in
Sec.~\ref{sec:information_content_features}. There, we apply the learned
features to camera identification and also %to white-point estimation, which is closely related 
to illuminant estimation. %~\cite{DBLP:conf/cvpr/StantonHM19}.

%Another way to look at the proposed training approach is shown in Fig.~\ref{fig:unification_camera_illuminant}. For camera identification, it is desirable to have features independent of the image content~\cite{kirchner2015forensic} (``invariance'' w.r.t.\ scene), while only depending on the camera (``covariance'' w.r.t.\ camera). For illuminant estimation, on the other hand, it is desirable for the features to be independent of the camera~\cite{cheng2014illuminant} (``invariance'' w.r.t.\ camera), and to only depend on the illuminant distribution of a scene (``covariance'' w.r.t.\ scene). The proposed embeddings can be considered a unification of the disciminators for camera identification and illuminant estimation, as we jointly enforce the covariances desirable for both tasks. This allows us to expose forgeries by discrepancies of either of the two quantities.
%In the other direction, camera identification and illuminant estimation can be considered marginalizations of our approach, by replacing one covariance a time with an invariance. Indeed, in Sec.~\ref{sec:information_content_features}, we will show that our approach is related to these two tasks as the learned metric features contain information exploitable for camera identifiaction and white-point estimation, the latter of which is related to illuminant estimation~\cite{DBLP:conf/cvpr/StantonHM19}.

\subsection{Datasets and Training/Test/Validation Splits}
\label{sec:data_simulation}

The training requires images of identical scenes imaged with different camera
pipelines. To obtain such images, we post-process RAW images to final images
with various parameter combinations that affect the color processing. \edited{Each RAW image constitutes one scene.}

After removing broken and completely over- or underexposed images, we use a
total of $5997$ RAW images from the RAISE database~\cite{dang2015raise}, $4998$
RAW images from the MIT-Adobe FiveK
dataset~\cite{DBLP:conf/cvpr/BychkovskyPCD11}, $1632$ RAW images from the
dataset by Nam and Kim~\cite{DBLP:conf/iccv/NamK17}, and $645$ RAW images
crawled from \texttt{raw.pixls.us}.

The RAW-to-final-image conversion is performed with \texttt{rawpy}, a Python
wrapper for \texttt{LibRaw}.
The camera pipelines differ in the applied white balancing and color space
transformations.
More specifically, each image is white-balanced with each of the modes
\mbox{``autoWB''}, ``cameraWB'' and ``noWB'', combined with each of the four
color transformations ``raw'', ``sRGB'', ``Adobe'' and ``ProPhoto''.
These parameters yield a total of $\lvert\mathcal{C}\rvert = 12$ \edited{camera} pipelines per RAW image.
The resulting images differ in their color appearance, as can be seen for the
input images in Fig.~\ref{fig:training_approach}.  For some combinations of
scenes and acquisition devices, a subset of the pipelines can lead to almost
identical results, which is the reason for the use of the filter criterion in
Eqn.~(\ref{eq:heur_min_dissim}).

We use the preset training/validation/test split for the dataset by Nam and Kim~\cite{DBLP:conf/iccv/NamK17}. The remaining data is split by scenes with a ratio of $0.8$, $0.1$, $0.1$\edited{, such that the content of validation and test images has not been seen during training.}
This yields a total of $\lvert\mathcal{S}^{\mathrm{train}}\rvert=10494$, $\lvert\mathcal{S}^{\mathrm{val}}\rvert=1463$, and $\lvert\mathcal{S}^{\mathrm{test}}\rvert=1315$ scenes for training, validation and test, and thus $\lvert\mathcal{S}^{\mathrm{train}}\rvert \cdot \lvert\mathcal{C}\rvert = 125928$, $\lvert\mathcal{S}^{\mathrm{val}}\rvert \cdot \lvert\mathcal{C}\rvert = 17556$, and $\lvert\mathcal{S}^{\mathrm{test}}\rvert \cdot \lvert\mathcal{C}\rvert = 15780$ images in total. 

The training and validation process operates on image patches. Training and
validation patches without variation in
content, and patches that are severely under- or over-exposed are excluded to
ensure sufficient color variation.
More specifically, we exclude patches with $\min{\bm{p}_{i}}=\max{\bm{p}_{i}}$, and only allow patches with few overexposed pixels
\begin{equation}
\label{eq:heur_overexposed}
\sum_{k=0}^{2}\mathbbm{1}\left(\left(\sum_{m=0}^{M^{\prime}-1}\sum_{n=0}^{N^{\prime}-1} \mathbbm{1}(\bm{p}_{i}[m,n,k] > \rho_{\mathrm{hi}})\right) > \delta_{\mathrm{hi}}\right) \leq N_\mathrm{hi}
\end{equation}
and few underexposed pixels
\begin{equation}
\label{eq:heur_underexposed}
\sum_{k=0}^{2}\mathbbm{1}\left(\left(\sum_{m=0}^{M^{\prime}-1}\sum_{n=0}^{N^{\prime}-1} \mathbbm{1}(\bm{p}_{i}[m,n,k] < \rho_{\mathrm{lo}})\right) > \delta_{\mathrm{lo}}\right) \leq N_\mathrm{lo}
\end{equation}
with indicator function $\mathbbm{1}(\placeholder)$ and intensity thresholds
$\rho_{\mathrm{hi}}$, $\rho_{\mathrm{lo}}$, $\delta_{\mathrm{hi}}$,
$\delta_{\mathrm{lo}}$, $N_{\mathrm{hi}}$ and $N_{\mathrm{lo}}$.
These equations ensure that at most $N_{\mathrm{hi}}$ color channels exhibit
more than $\delta_{\mathrm{hi}}$ overexposed pixels with intensity larger than
$\rho_{\mathrm{hi}}$. Underexposed pixels are analogously treated with
$N_{\mathrm{lo}}$, $\delta_{\mathrm{lo}}$ and $\rho_{\mathrm{lo}}$.
Furthermore, when patches are paired with the same scene but different camera
pipelines, we ensure a minimum distance $\delta_{\mathrm{Lab}}$ of both patches in \emph{Lab}-color space:
\begin{equation}
\label{eq:heur_min_dissim}
	\begin{split}
	&\enspace\forall c_1\in\mathcal{C}_k^s, \enspace c_1\neq c_0:\\
	&\sum\limits_{m=0}^{M^\prime-1}\sum\limits_{n=0}^{N^\prime-1}	\frac{\sqrt{\sum\limits_{k=0}^{2}(\bm{\tilde{p}}_i^{c_0,s}[m,n,k] - \bm{\tilde{p}}_i^{c_1,s}[m,n,k])^2}}{M^\prime N^\prime} \geq \delta_{\mathrm{Lab}}\enspace,
	\end{split}
\end{equation}
where $\mathcal{C}_k^s$ is the set of all cameras selected for scene $s$ in batch $k$, see Sec.~\ref{sec:learning_metric_color_features}, and $\tilde{\bm{p}}_i$ denotes patch $\bm{p}_i$ in \emph{Lab}-color space.
This constraint prevents the attempt to learn differences between color
pipelines that do not manifest in differences of the color patch.

\subsection{Learning Metric Color Features}
\label{sec:learning_metric_color_features}

The proposed CNN is based on the ResNet-50
architecture~\cite{DBLP:conf/cvpr/HeZRS16}, with input patch size
$M^{\prime}\times N^{\prime} = 128\times 128$ and weights pretrained on
ImageNet~\cite{DBLP:conf/cvpr/DengDSLL009}.
The output layer is replaced with a
linear dense layer with $q=64$ units, where the weights are initialized with
Glorot Uniform Initialization~\cite{DBLP:journals/jmlr/GlorotB10}. 
\edited{The theoretical model is incorporated into the training of this network, which is described in the remainder of this Section.}

 %We employ cosine similarities \mbox{$r:\mathcal{M}\times\mathcal{M}\rightarrow [-1,1]$}, 
%\begin{equation}
%	\label{eq:cos_similarity}
%	r(\bm{y}_{i_1},\bm{y}_{i_2}) = \frac{\bm{y}_{i_1}^{\top}\bm{y}_{i_2}}{\left\lVert\bm{y}_{i_1}\right\rVert\cdot\left\lVert\bm{y}_{i_2}\right\rVert}\enspace,
%\end{equation}
%for the metric $d:\mathcal{M}\times\mathcal{M}\rightarrow [0,1]$ measuring the distance between two embeddings \mbox{$\bm{y}_{i_1}, \bm{y}_{i_2}\in\mathcal{M}$},
%\begin{equation}
%	d(\bm{y}_{i_1}, \bm{y}_{i_2}) = \frac{1}{2}\cdot\left(1 - s(\bm{y}_{i_1},\bm{y}_{i_2}) \right)\enspace,
%\end{equation}
%where the top indices $c$, $s$ have been dropped for ease of notation. We further apply $d_{i_1,i_2}$ as shorthand notation for $d(\bm{y}_{i_1}^{c_1,s_1}, \bm{y}_{i_2}^{c_2,s_2})$. Due to the normalization in Eqn.~(\ref{eq:cos_similarity}), this metric discards differences in length of the embedding vectors, and only measures their angular distance. 
%For parallel (i.e. most similar)
%embeddings, it follows $d_{i_1,i_2} = 0$, while $d_{i_1,i_2}=1$ indicates anti-parallelity, and $d_{i_1,i_2}=0.5$ holds for orthogonality.

 The parameters $\bm{\theta}$ of the CNN $f_{\bm{\theta}}$ are optimized to
satisfy the conditions in Eqns.~(\ref{eq:rel_dist_same_scene_diff_pipel}) and
(\ref{eq:rel_dist_diff_scene_diff_pipel}) to obtain the desired invariances and
covariances.
To this end, the CNN is presented training data in mini-batches \mbox{$\bm{P}_k\in\mathbb{R}^{N_{b}\times M^{\prime}\times N^{\prime}\times 3}$} indexed by $k$.
Each mini-batch is stack of $N_b$ patches $\bm{p}_i$.
The CNN maps a mini-batch $\bm{P}_k$ to the output \mbox{$\bm{Y}_k\in\mathbb{R}^{N_b\times q}$} of $N_b$ embeddings $\bm{y}_i$.
Given $\bm{Y}_k$, we compute the matrix \mbox{$\bm{D}_k\in[0,1]^{N_b\times N_b}$} of pairwise distances with \mbox{$\bm{D}_{k}[i_0,i_1] = d(\bm{y}_{i_0}, \bm{y}_{i_1})$} and indexing operator $[\placeholder]$.
We further define boolean ground truth matrices \mbox{$\bm{B}^{\star}_k\in\{0,1\}^{N_b\times N_b}$}, i.e., masks indicating similar ($\bm{B}_k^{\star}[i_0,i_1]=0$) and dissimilar ($\bm{B}_k^{\star}[i_0,i_1]=1$) pairings for indices $i_0$ and $i_1$.
The parameters $\bm{\theta}$ are iteratively updated after each mini-batch, to
continuously decrease the distances of similar embeddings while continuously
increasing the distances of dissimilar embeddings, as indicated by $\bm{B}_k^\star$.

%We use a set of scenes $\mathcal{S}$ captured in RAW format and developed to
%JPEG images. Details on data sources, dataset splits, and tools for RAW-to-JPEG
%conversion are presented in the next Subsection.
%Each RAW image is transformed to a final image with
%a set of camera pipelines $\mathcal{C}$, yielding
%$\lvert\mathcal{S}\rvert\cdot\lvert\mathcal{C}\rvert$ final images in total.
%The set of scenes $\mathcal{S}$ is split \emph{disjointly} into the three sets $\mathcal{S}^{\mathrm{train}}$, $\mathcal{S}^{\mathrm{val}}$ and $\mathcal{S}^{\mathrm{test}}$ for training, validation and testing, respectively.
%Thereby, the number of training images is $\lvert\mathcal{S}^{\mathrm{train}}\rvert\cdot\lvert\mathcal{C}\rvert$.
%{\color{red}{Benjamins:}}
During each training epoch, we split $\mathcal{S}^{\mathrm{train}}$ in disjoint subsets $\{\mathcal{S}_k^{\mathrm{train}}\}_k$ with  $\lvert\mathcal{S}_k^\mathrm{train}\rvert \coloneqq N^{\mathcal{S}}_b\leq\lvert\mathcal{S}^{\mathrm{train}}\rvert$.
One such subset is used per minibatch $k$.
For each $s\in\mathcal{S}_k^\mathrm{train}$, we select a random subset $\mathcal{C}_k^s\subset\mathcal{C}$ of different camera pipelines with $\lvert\mathcal{C}_k^s\rvert \coloneqq  N_b^{\mathcal{C}}<\lvert\mathcal{C}\rvert$.
Finally, from each of the resulting $N_b^{\mathcal{S}}\cdot N_b^{\mathcal{C}}$
images, we randomly extract $N_b^{\mathcal{P}}$ patches, which yields a total
of $N_b^{\mathcal{S}}\cdot N_b^{\mathcal{C}}\cdot N_b^{\mathcal{P}}\coloneqq
N_b$ patches per mini-batch. Note that the patch exclusion criteria from the
previous subsection (under-/overexposure, color variation) have been applied
before. After all scenes have been processed in an epoch, the order of the
scenes is randomly permuted before starting the next epoch to support the
training process.

\begin{figure}[t]
	\centering
	\includegraphics[width=\linewidth]{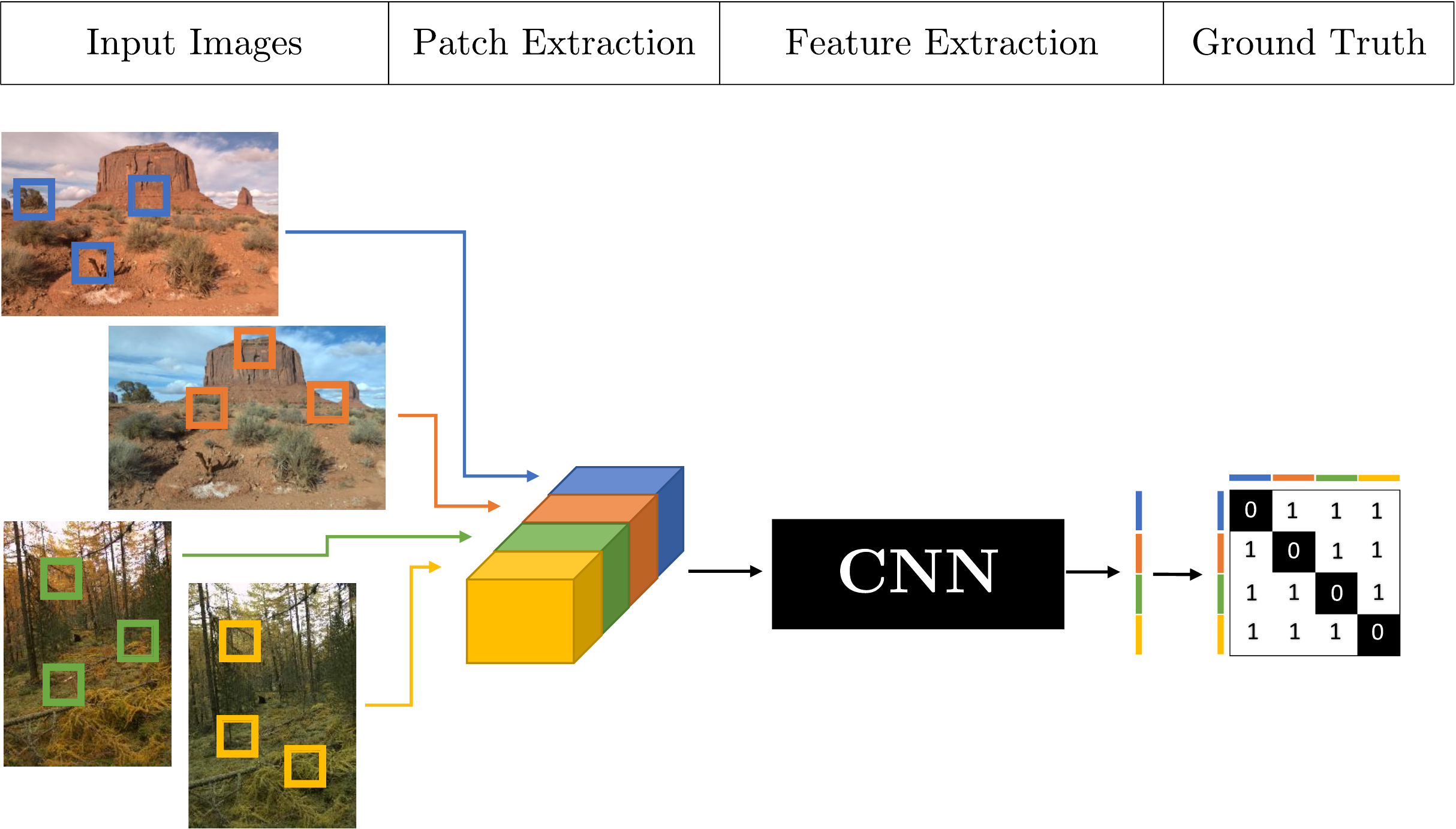}
	\caption{Illustration of the proposed training approach. Patches from the same scene \emph{and} same camera pipeline are similar (``0'' in boolean ground truth matrix), while patches from different camera pipelines \emph{or} different scenes are dissimilar (``1'').}
	\label{fig:training_approach}
\end{figure}

The elements of the boolean ground truth mask $\bm{B}_k^{\star}[i_0,i_1]$ %$(\bm{y}_{i_0}^{c_0,s_0},\bm{y}_{i_1}^{c_1,s_1})$ 
are set in agreement with the constraints for $\bm{y}_{i_0}^{c_0,s_0}$ and $\bm{y}_{i_1}^{c_1,s_1}$ in Sec.~\ref{sec:theo_background}: patch pairs from within the same image image ($s_0=s_1\land c_0=c_1$) are
considered similar, and obtain a logical $0$ in the ground truth mask. Patch pairs
are dissimilar if they stem from different scenes (\mbox{$s_0 \neq s_1$}) or
from the same scene but from different camera pipelines (\mbox{$s_0 = s_1 \land
c_0\neq c_1$}). These pairs obtain logical $1$, indicating dissimilarity.
The resulting boolean ground truth matrix $\bm{B}_k^{\star} =
\bm{B}^{\star}=\mathrm{const.}$ is a block matrix. On the diagonal are
\mbox{$N_b^{\mathcal{S}}\cdot N_b^{\mathcal{C}}$} block matrices of shape $N_b^\mathcal{P}\times N_b^\mathcal{P}$ that consist only of zeros. All remaining $N_{b}^{\mathcal{S}}\cdot
N_{b}^{\mathcal{C}}\cdot(N_{b}^{\mathcal{S}}\cdot N_{b}^{\mathcal{C}} - 1)$
off-diagonal blocks of the same shape consist only of ones.
Figure~\ref{fig:training_approach} provides an overview of the training
approach for $N_b^{\mathcal{S}} = 2$, $N_{b}^{\mathcal{C}} = 2$ and
$N_b^{\mathcal{P}} = 3$.

To measure and optimize the performance of the CNN, we employ the histogram loss $\mathcal{L}_{\bm{\theta}}^{\mathrm{hist}}$ by Ustinova and Lempitsky~\cite{DBLP:conf/nips/UstinovaL16}.
\edited{This loss function measures} the probability that a similar pair of embeddings has a
larger distance $d(\bm{y}_{i_0}^{c_0,s_0},\bm{y}_{i_1}^{c_1,s_1})$ than a
dissimilar pair.
\edited{In~\cite{DBLP:conf/nips/UstinovaL16}, this \edited{so-called} \emph{probability of the reverse} is expressed using the probability densities $p^{+}$, $p^{-}$ of the pairwise similarties of similar and dissimilar pairs, which are approximated using histograms, denoted $h_r^{+}$, $h_r^{-}$, respectively.} 
%is approximated using histogram estimates $h_r^{+}$, $h_r^{-}$ for the probability densities $p^{+}$, $p^{-}$ of the pairwise similarties of similar and dissimilar pairs.
Following Ustinova and Lempitsky, we estimate $h_r^{+}$ and $h_r^{-}$ at equidistant nodes with $t_0=-1$ and $t_{R-1}=1$, and we fix $R=26$.
Each similarity is herein attributed to the two neighboring nodes with weights
proportional to the distances between samples and nodes.
Note that we only consider distances $d(\bm{y}_{i_0}, \bm{y}_{i_1})$ for $i_0 >
i_1$ to avoid identical and duplicate pairs.
%, using a triangular interpolation kernel, i.e.\
%\begin{equation}
%	h_r^{+} = \frac{1}{\lvert\mathcal{I}_{+}^{\star}\rvert}\sum_{(i_1,i_2) \in \mathcal{I}_{+}^{\star}} \omega_{i_1,i_2,r}\enspace,
%\end{equation}
%with weights
%\begin{equation}
%	\omega_{i_1,i_2,r} = \begin{cases}
%	(s_{i_1,i_2} - t_{r-1})/T,\quad \mathrm{if}\, & s_{i_1,i_2}\in[t_{r-1},t_r]\enspace,\\
%	(t_{r+1} - s_{i_1,i_2})/T,\quad \mathrm{if}\, & s_{i_1,i_2}\in[t_{r},t_{r+1}]\enspace,\\
%	0 & \mathrm{otherwise}\enspace.
%	\end{cases}
%\end{equation}
%We denote by $\mathcal{I}_{+}^{\star}$ the set of pairs of indices $i_1,i_2$ of similar embeddings, i.e.\ pairs with $\bm{D}^{\star}[i_1,i_2]=0$, for which additionally \mbox{$i_1 < i_2$} holds, to only account for unique and distinct pairs. We further define $s_{i_1,i_2} \coloneqq s(\bm{y}_{i_1},\bm{y}_{i_2})$ and $T = 2/(R-1)$. $h_r^{-}$ is computed analogously.
With the cumulative density function $\Phi_r^{+} = \sum_{q=0}^{r}h_{q}^{+}$, the loss function is obtained as
\begin{equation}
	\mathcal{L}_{\bm{\theta}}^{\mathrm{hist}} = \sum_{r=0}^{R-1}h_r^{-}\Phi_{r}^{+}\enspace.
\end{equation}
The number $R$ of bins of the histograms is the only parameter of $\mathcal{L}_{\bm{\theta}}^{\mathrm{hist}}$; however, it has negligible effect on the outcome~\cite{DBLP:conf/nips/UstinovaL16}.

\begin{figure*}[t]
	\centering
	\includegraphics[width=\linewidth]{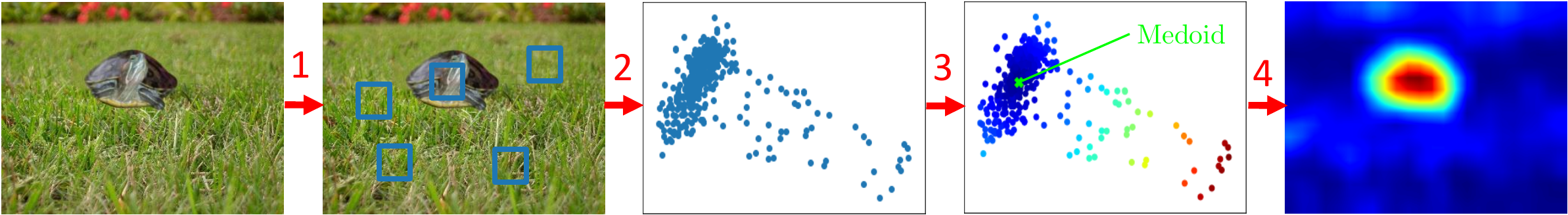}
	\caption{Application of the CNN embedding to manipulation localization. The distance of each image patch w.r.t.\ to the medoid of all patches is measured in the learned embedding space (distances color-coded; embeddings projected to $\mathbb{R}^2$ using Multidimensional Scaling~\cite{kruskal1964multidimensional} for visualization), and then aggregated back into the image space to create a heatmap of manipulation likelihoods.}
	\label{fig:manipulation_localization}
\end{figure*}

We further employ the so-called global orthogonal regularizer
$\mathcal{R}_{\bm{\theta}}^{\mathrm{orth}}$, proposed by Zhang~\emph{et al.}~\cite{DBLP:conf/iccv/ZhangYKC17},
to enforce dissimilar embeddings $\bm{y}_{i_0}^{c_0,s_0}$ and $\bm{y}_{i_1}^{c_1,s_1}$ with
$c_0\neq c_1 \lor s_0\neq s_1$ to spread out in the target domain $\mathcal{M}$.
This is achieved by regularizing the mean $M_1$ and second moment $M_2$ of the distribution of pairwise distances of dissimilar pairs towards the moments $M_1^{\mathcal{U}} = 0$ and $M_2^{\mathcal{U}} = q^{-1}$ of a uniform distribution on the unit sphere $\mathbb{S}^{q-1}$.
With $\mathcal{I}_{-}^{\star} = \{(i_0,i_1)\enspace|\enspace i_0>i_1 \land \bm{B}^{\star}[i_0,i_1] = 1\}$, the moments are computed as
\begin{equation}
	M_1 = \frac{1}{\lvert \mathcal{I}_{-}^{\star}\rvert}\sum_{(i_0,i_1)\in\mathcal{I}_{-}^{\star}} s(\bm{y}_{i_0}, \bm{y}_{i_1})\enspace,
\end{equation}
\begin{equation}
	M_2 = \frac{1}{\lvert \mathcal{I}_{-}^{\star}\rvert}\sum_{(i_0,i_1)\in\mathcal{I}_{-}^{\star}} s(\bm{y}_{i_0}, \bm{y}_{i_1})^2\enspace,
\end{equation}
\edited{such that the regularizer is obtained as}
\begin{equation}
\label{eq:regularizer}
\mathcal{R}_{\bm{\theta}}^{\mathrm{orth}} = M_1^2 + \max(0, M_2 - \frac{1}{q})\enspace.
\end{equation} 
In total, the CNN is trained by iteratively minimizing
\begin{equation}
\label{eq:loss_function}
	\mathcal{L}_{\bm{\theta}} = \mathcal{L}_{\bm{\theta}}^{\mathrm{hist}} + \eta\cdot\mathcal{R}_{\bm{\theta}}^{\mathrm{orth}}
\end{equation}
for some weighting factor $\eta$.

During training, we artificially degrade and randomly augment the image patches
to improve the generalization and robustness of the network.
Prior to filtering for under-/overexposure and color variation in
Eqn.~(\ref{eq:heur_overexposed}) to Eqn.~(\ref{eq:heur_min_dissim}), each image
is resized with constant aspect
ratio, such that the larger image dimension is $1536$ pixels. 
Each extracted patch is further augmented via random horizontal or vertical
flipping, random rotation and shearing, resizing, and JPEG compression.
Rotation and shearing is done with angles in the range of
$[-5^{\circ},5^{\circ}]$. Resizing is done with factors in the range of $[0.95,
1.05]$. A patch is JPEG compressed with a probability of $0.5$, with a random
quality level in the range $[50,100]$.  These resampling and recompression
steps remove much of the high-frequent statistical information, which forces
the CNN to learn more robust traces. 

The CNN's generalization is measured for identical validation mini-batches
throughout all epochs. For this, we use the validation set
$\mathcal{S}^{\mathrm{val}}$ filtered for under-/overexposure and color
variation as described in Subsec.~\ref{sec:data_simulation}, but without
any patch degradation or augmentation.

The number of similar and dissimilar patch pairs per batch are uneven.
To nevertheless measure the separability of the two distributions, the
validation performance in the experiments is measured with the area under the
ROC curve (ROC AUC) of the distributions of similar and dissimilar patches.
Based on validation loss, we select the optimal parameters
$\bm{\theta}^{\star}$ for manipulation analysis, which is presented in greater
detail in Sec.~\ref{sec:cnn_training}.

\subsection{Application to Forgery Detection and Localization}
\label{sec:application_to_manipulation_detection}

We create a heatmap from the trained CNN $f_{\bm{\theta}^{\star}}$ for
manipulation detection and localization. Here, \emph{detection} denotes a
binary yes/no statement whether an image is spliced. \emph{Localization}
denotes a segmentation of the spliced region within an image. The specific
processing steps are illustrated in Fig.~\ref{fig:manipulation_localization}
and described below.

A test image $\bm{I}^{\mathrm{test}}\in\mathbb{R}_{+}^{M\times N\times 3}$ is
first resized with constant aspect ratio such that its larger dimension is
$1536$ pixels. The following steps are to
\begin{enumerate}
	\item Extract patches $\bm{p}_{i}$ with stride 32,
	$i\in\{0,\dots,N^{\mathrm{test}}-1\}$. This yields $N^{\mathrm{test}}$
	patches. For typical image aspect ratios, $N^{\mathrm{test}}\approx 10^3$.
	\item Compute the CNN embedding $\bm{y}_i = f_{\bm{\theta}^{\star}}(\bm{p}_i)$ for each patch $\bm{p}_i$.
	\item Compute the medoid $\bm{\mu}$ of the embeddings,
	\begin{equation}
		\bm{\mu} = \argmin\limits_{\bm{y}\in\{\bm{y}_i\}_i}\sum_{i=0}^{N^{\mathrm{test}}-1}d(\bm{y}, \bm{y}_i)\enspace,
	\end{equation}
	to obtain a reference point for consistent imaging conditions. 
	Relate each embedding $\bm{y}_i$ to that reference $\bm{\mu}$
	by computing an inconsistency score $\gamma_i = d(\bm{\mu}, \bm{y}_i)\in
	[0,1]$.
	\item Project the scores $\gamma_i$ back to the image locations of the
	corresponding patch $\bm{p}_i$ to create a heatmap $\tilde{\mathcal{H}}$.
	In this map, larger values indicate a higher manipulation likelihood.
\end{enumerate}
Finally, the heatmap $\tilde{\mathcal{H}}$ is resized to the original test
image dimensions to obtain the heatmap $\mathcal{H}\in[0,1]^{M\times N}$.
Note that during this analysis, we apply the criteria in
Eqn.~(\ref{eq:heur_overexposed}) to Eqn.~(\ref{eq:heur_min_dissim}), and assign
$\gamma_i=0$ to all patches that were filtered out prior to the analysis.

For forgery \emph{detection}, we compute the average over the heatmap as an
image global manipulation score 
\begin{equation}
\label{eq:detection_score}
	\Gamma = \frac{1}{MN}\sum_{m=0}^{M-1}\sum_{n=0}^{N-1} \mathcal{H}[m,n]\enspace,
\end{equation}
with $0\leq\Gamma\leq1$.
%This equation assigns higher scores $\Gamma$ (and hence a higher manipulation likelihood for larger deviations a potentially manipulated area and the larger the deviations of the corresponding embeddings from the medoid, the higher the score $\Gamma$ and thus the assigned likelihood that an image is manipulated.

We also investigate an alternative variant of transforming the pairwise
distance matrix $\bm{D}^{\mathrm{test}}$ to patchwise scores $\gamma_i$ by
using the computationally more complex MeanShift
aggregation~\cite{DBLP:journals/pami/Cheng95}, which is also used
in~\cite{DBLP:conf/eccv/HuhLOE18}.

\section{Experimental Results}
\label{sec:experimental_results}

In this section, we first report the concrete training parameters and practical
training aspects. The subsequent evaluation consists of multiple parts. First,
we demonstrate the robustness of the learned embeddings to degraded image
qualities. Then, we compare our algorithm with state-of-the-art approaches for
manipulation localization and detection on various datasets and image
qualities.  Finally, we demonstrate the dual nature of the embeddings by
applying the metric space to camera identification and white-point estimation.

\subsection{Training and Evaluation of the CNN}
\label{sec:cnn_training}

\begin{figure*}[t]
	\centering
	\begin{subfigure}[t]{0.24\linewidth}
		\includegraphics[width=\linewidth]{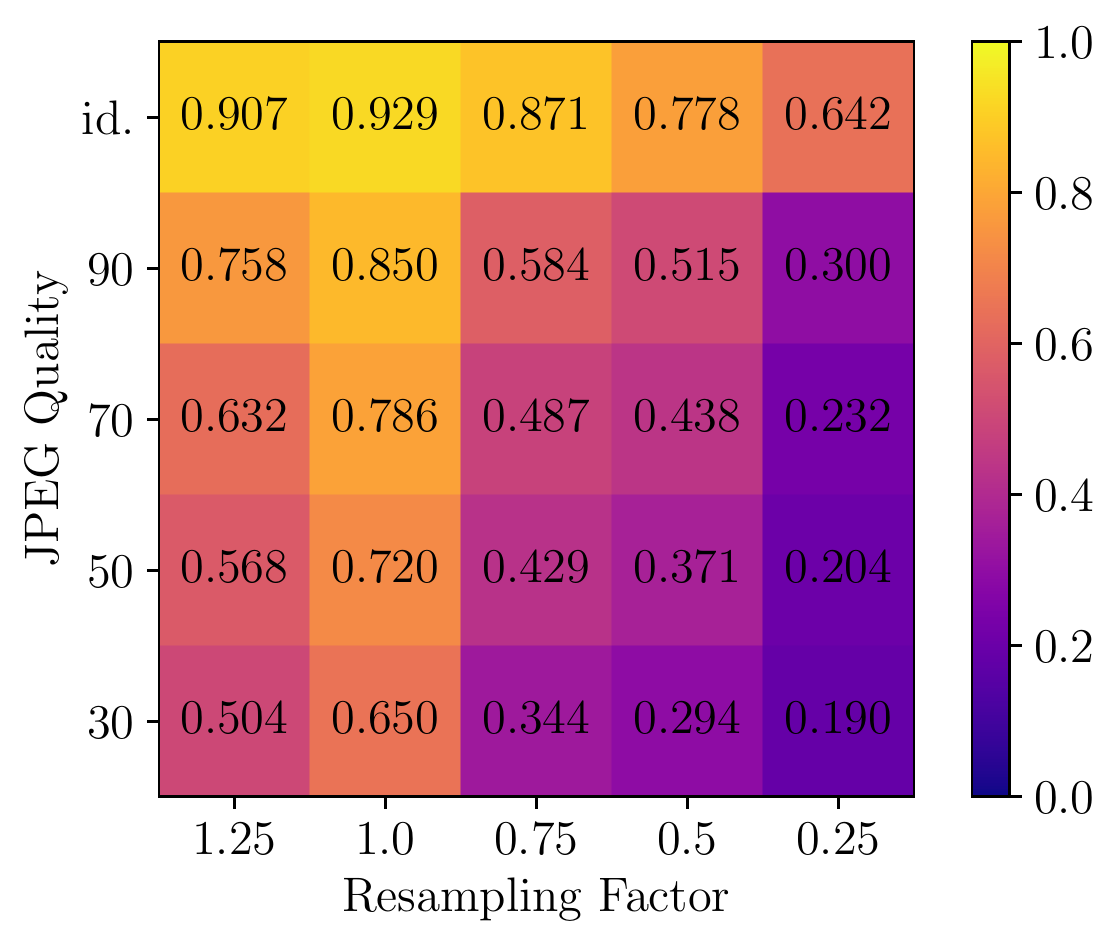}
		\caption{Different scenes}
		\label{fig:robustness_diff_scenes}
	\end{subfigure}
	\begin{subfigure}[t]{0.24\linewidth}
		\includegraphics[width=\linewidth]{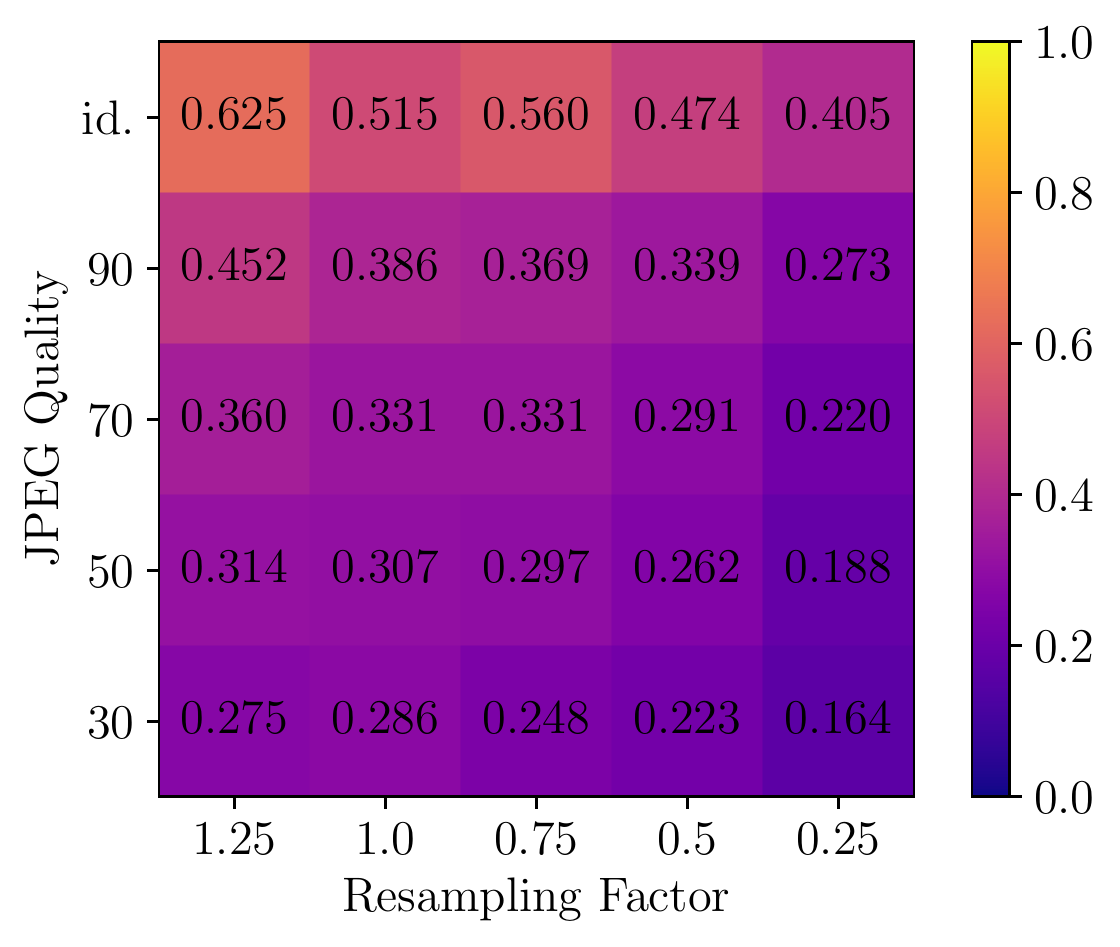}
		\caption{Same scenes, $\delta_{\mathrm{Lab}}^{\mathrm{min}}=0$}
		\label{fig:robustness_same_scenes_0}
	\end{subfigure}
	\begin{subfigure}[t]{0.24\linewidth}
		\includegraphics[width=\linewidth]{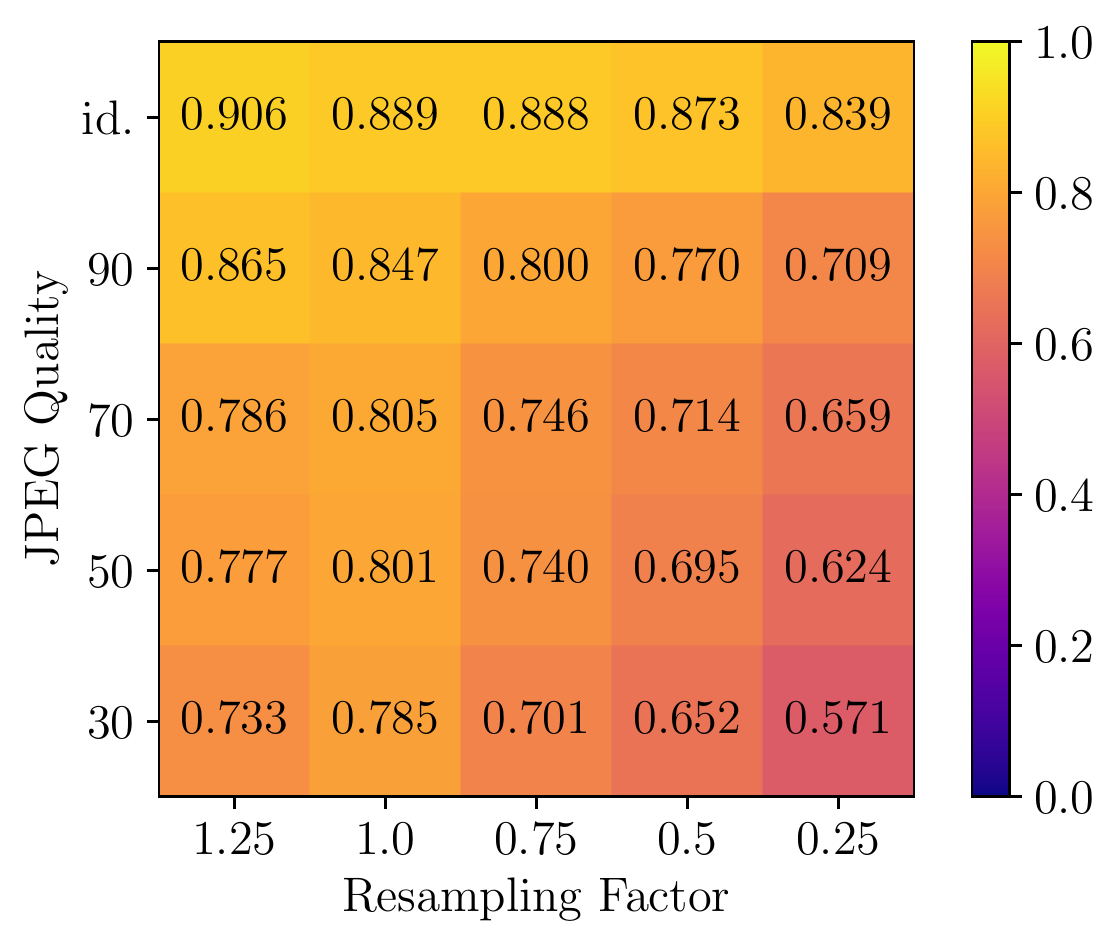}
		\caption{Same scenes, $\delta_{\mathrm{Lab}}^{\mathrm{min}}=20$\\ \centering\reedited{($14.7\%$ of patches)}}
		\label{fig:robustness_same_scenes_20}
	\end{subfigure}
	\begin{subfigure}[t]{0.24\linewidth}
		\includegraphics[width=\linewidth]{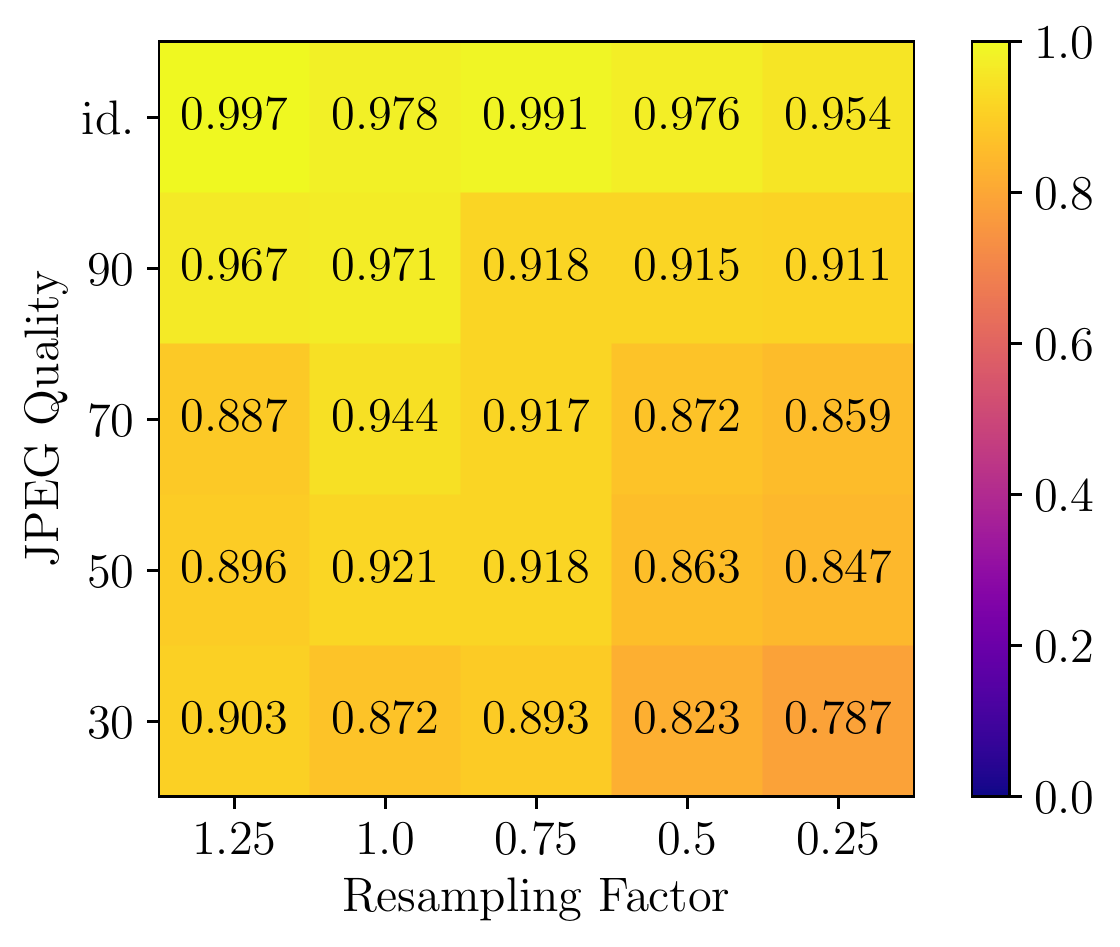}
		\caption{Same scenes, $\delta_{\mathrm{Lab}}^{\mathrm{min}}=50$\\\centering\reedited{($0.883\%$ of patches)}}
		\label{fig:robustness_same_scenes_50}
	\end{subfigure}
%	\caption{Separation of the empirical distributions of the pairwise embedding distances for patches from images of similar and dissimilar imaging conditions for various post-processing operations, measured with ROC AUC. a) $50$ pairs of images of different scenes. b)-d) $50$ image pairs where one scene each has been imaged with two different camera pipelines for increasing minimum \emph{Lab}-distances.}
	\caption{Separation of the empirical distributions of the pairwise embedding distances for patches from images of similar and dissimilar imaging conditions for various post-processing operations, measured \reedited{in terms of True Positive Rate at a fixed False Alarm Rate of $5\%$}. a)~$50$ pairs of images of different scenes. b)-d)~$50$ image pairs where one scene each has been imaged with two different camera pipelines for increasing minimum \emph{Lab}-distances.}
	\label{fig:robustness}
\end{figure*}

\edited{The hyperparameters are summarized in Tab.~\ref{tab:summary_parameters}. The patch size is set to $M^\prime\times{}N^\prime = 128\times{}128$, trading off resolution and number of patches for training.} The para\-meters for the filter criteria in Eqn.~(\ref{eq:heur_overexposed}) and
Eqn.~(\ref{eq:heur_underexposed}) are set to $\rho_{\mathrm{hi}} = 252$,
$\rho_{\mathrm{lo}} = 5$ for intensities $0\leq \bm{p}_i[m,n,k] \leq 255$, and
further $\delta_{\mathrm{hi}} = \delta_{\mathrm{lo}} = 0.3 M^{\prime}
N^{\prime}$, $N_{\mathrm{hi}} = 2$ and $N_{\mathrm{lo}}=3$. For Eqn.~(\ref{eq:heur_min_dissim}), we set $\delta_{\mathrm{Lab}} = 5$ with intensities in \emph{Lab}-color space as $0 \leq \tilde{\bm{p}}_i[m,n,k] \leq 255$. \reedited{For these parameters, Eqns.~(\ref{eq:heur_overexposed}) and (\ref{eq:heur_underexposed}) on average exclude $3.3\%$ of patches, that are either over- or underexposed. Equation~(\ref{eq:heur_min_dissim}) further excludes $32.8\%$ of patches whose counterparts from different color pipelines are very similar. The above parameter values are used throughout all training, validation and testing runs, except for the experiments depicted in Fig.~\ref{fig:robustness_same_scenes_0}, Fig.~\ref{fig:robustness_same_scenes_20} and Fig.~\ref{fig:robustness_same_scenes_50}, where we investigate the effect of different values for $\delta_\mathrm{Lab}$.}

\begin{table}
	\centering
	\caption{\edited{Summary of hyperparameters}}
	\label{tab:summary_parameters}
	\begin{tabular}{|c|c|c|c|}
		\hline
		\textbf{Category} & \textbf{Parameter} & \textbf{Value} & \textbf{Comment / Description}\\
		\hhline{|=|=|=|=|}
		Patch size & $M^\prime$, $N^\prime$ & $128$ & patch size: $M^\prime\times N^\prime$\\ \hline
		\multirow{4}{*}{\shortstack[0]{Filter criteria\\ (Eqns. \eqref{eq:heur_overexposed}--\eqref{eq:heur_min_dissim})}} & $\rho_{\mathrm{lo}}$ & $5$ & \multirow{2}{*}{$0\leq \bm{p}_i[m,n,k] \leq 255$} \\
		\cline{2-3}
		& $\rho_{\mathrm{hi}}$ & $252$ & \\
		\cline{2-4}
		& $\delta_{\mathrm{lo}}$, $\delta_{\mathrm{hi}}$ & $0.3 M^\prime N^\prime$ & relative to patch size \\
		\cline{2-4} 
		& $\delta_{\mathrm{Lab}}$ & $5$ & $0 \leq \tilde{\bm{p}}_i[m,n,k] \leq 255$ \\
		\hline
		\multirow{4}{*}{\shortstack[0]{Batch\\ configuration}} & $N_b^\mathcal{S}$ & $8$ & \#{}scenes per batch\\
		\cline{2-4}
		& $N_b^\mathcal{C}$ & $2$ & \#{}cameras per batch\\
		\cline{2-4}
		& $N_b^\mathcal{P}$ & $8$ & \#{}patches per image\\
		\cline{2-4}
		& $N_b$ & $128$ &  $N_b = N_b^\mathcal{S} N_b^\mathcal{C} N_b^\mathcal{P}$\\
		\hline	
		Loss function & $\eta$ & $0.5$ & Regularizer weight\\
		\hline
		\multirow{3}{*}{Optimizer} & $\alpha$ & $10^{-4}$ & Learning rate\\
		\cline{2-4}
		& $\beta_1$ & $0.9$ & \multirow{2}{*}{Moments}\\
		\cline{2-3}
		& $\beta_2$ & $0.999$ & \\
		\hline
	\end{tabular}
\end{table}

The tuning of all remaining hyperparameters, as well as the choice of the loss function, have been performed on the validation dataset.
The mini-batches for training and validation are created with $N_b^{\mathcal{S}} = 8$, $N_b^{\mathcal{C}} = 2$ and $N_b^{\mathcal{P}} = 8$, i.e., a batchsize of $N_b = 128$.
The weight in the loss function in Eqn.~(\ref{eq:loss_function}) is set to $\eta = 0.5$.
To optimize the CNN parameters $\bm{\theta}$, we use the Adam optimizer~\cite{kingma2014adam} with learning rate $\alpha=10^{-4}$ and moments $\beta_1 = 0.9$ and $\beta_2 = 0.999$.
Whenever the validation performance does not improve for at least 20 epochs, $\alpha$ is decreased by a factor of 10.
We apply early stopping once the validation loss is not decreasing anymore, to obtain the final parameters $\bm{\theta}^{\star}$.
This protocol lead to $149$ epochs for CNN training.

On a machine with 32 CPUs with 128\,GB RAM and an NVIDIA GeForce GTX 1080 GPU
with 12\,GB RAM, one epoch runs for about $40$ minutes in our implementation
with Python and Keras~\cite{chollet2015keras}. Hence, a full training run
requires about $100$ hours.
The final validation loss is $\mathcal{L}^{\mathrm{val}} = 0.0657$. 

\begin{figure}[t]
	\centering
	\begin{subfigure}[t]{0.4\linewidth}
		\includegraphics[width=\linewidth,trim={0 0 13.8cm 0},clip]{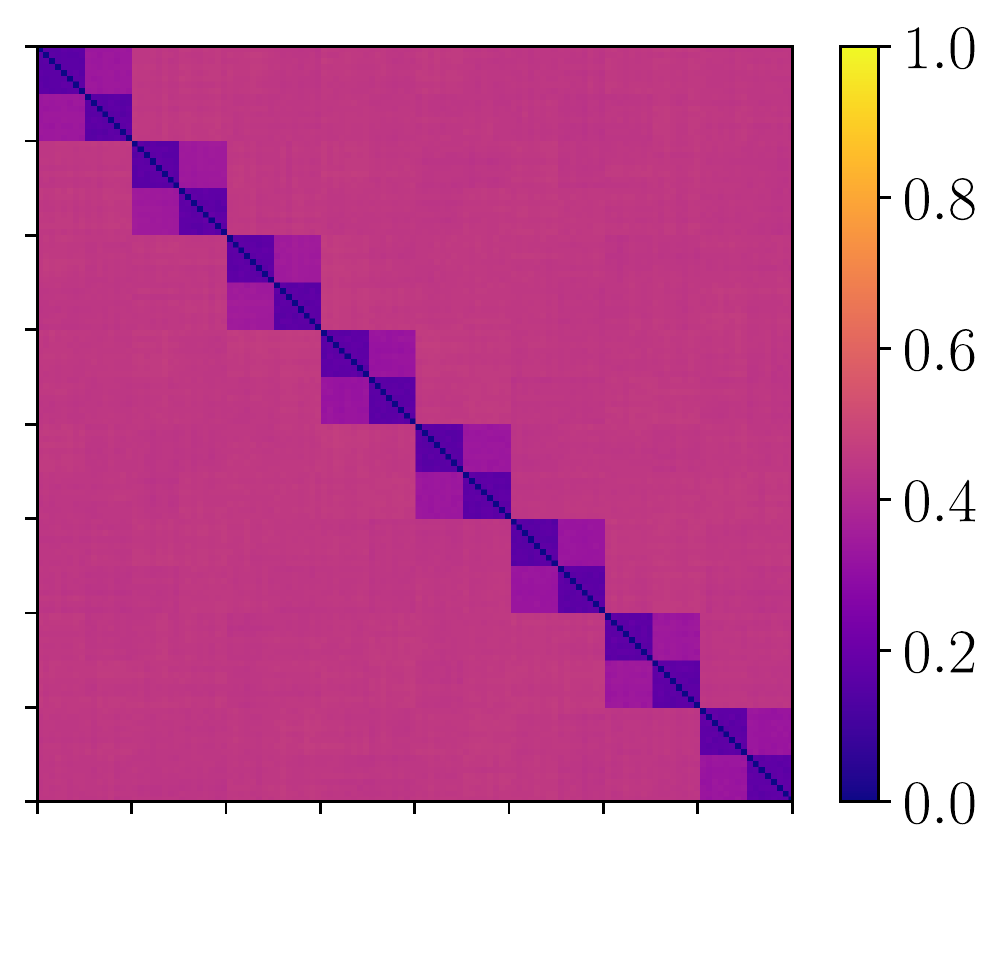}
		\caption{Pairwise distance matrix}
		\label{fig:val_dist_mat}
	\end{subfigure}
	\begin{subfigure}[t]{0.55\linewidth}
		\includegraphics[width=\linewidth,trim={10cm 0 0 0},clip]{figures/avg_hist_combined_pairwise_val_angular_dists.pdf}
		\caption{Distributions of pairwise distances}
		\label{fig:val_distributions}
	\end{subfigure}
	%\centering
	%\includegraphics[width=0.5\linewidth]{figures/hist_pairwise_val_angular_dists.pdf}
	\caption{Validation performance of the CNN. a) Matrix of pairwise distances, averaged over all validation mini-batches ($N_b=128$). b) Empirical distributions of pairwise distances for same (blue) and different (orange) imaging conditions.}
	\label{fig:validation_performance}
\end{figure}

%\begin{figure*}[t]
%	\centering
%	\begin{subfigure}[t]{0.24\linewidth}
%		\includegraphics[width=\linewidth]{figures/arbitraryScenes_rocauc_robustness.pdf}
%		\caption{Different scenes}
%		\label{fig:robustness_diff_scenes}
%	\end{subfigure}
%	\begin{subfigure}[t]{0.24\linewidth}
%		\includegraphics[width=\linewidth]{figures/sameScenes_rocauc_robustness_mindis0.pdf}
%		\caption{Same scenes, $\delta_{\mathrm{Lab}}^{\mathrm{min}}=0$}
%		\label{fig:robustness_same_scenes_0}
%	\end{subfigure}
%	\begin{subfigure}[t]{0.24\linewidth}
%		\includegraphics[width=\linewidth]{figures/sameScenes_rocauc_robustness_mindis20.pdf}
%		\caption{Same scenes, $\delta_{\mathrm{Lab}}^{\mathrm{min}}=20$\\ \centering\reedited{($14.7\%$ of patches)}}
%		\label{fig:robustness_same_scenes_20}
%	\end{subfigure}
%	\begin{subfigure}[t]{0.24\linewidth}
%		\includegraphics[width=\linewidth]{figures/sameScenes_rocauc_robustness_mindis50.pdf}
%		\caption{Same scenes, $\delta_{\mathrm{Lab}}^{\mathrm{min}}=50$\\\centering\reedited{($0.883\%$ of patches)}}
%		\label{fig:robustness_same_scenes_50}
%	\end{subfigure}
%	\caption{Separation of the empirical distributions of the pairwise embedding distances for patches from images of similar and dissimilar imaging conditions for various post-processing operations, measured with ROC AUC. a) $50$ pairs of images of different scenes. b)-d) $50$ image pairs where one scene each has been imaged with two different camera pipelines for increasing minimum \emph{Lab}-distances.}
%	\label{fig:robustness}
%\end{figure*}

\begin{figure*}
	\centering
	\begin{subfigure}[t]{0.32\linewidth}
		\includegraphics[width=\linewidth]{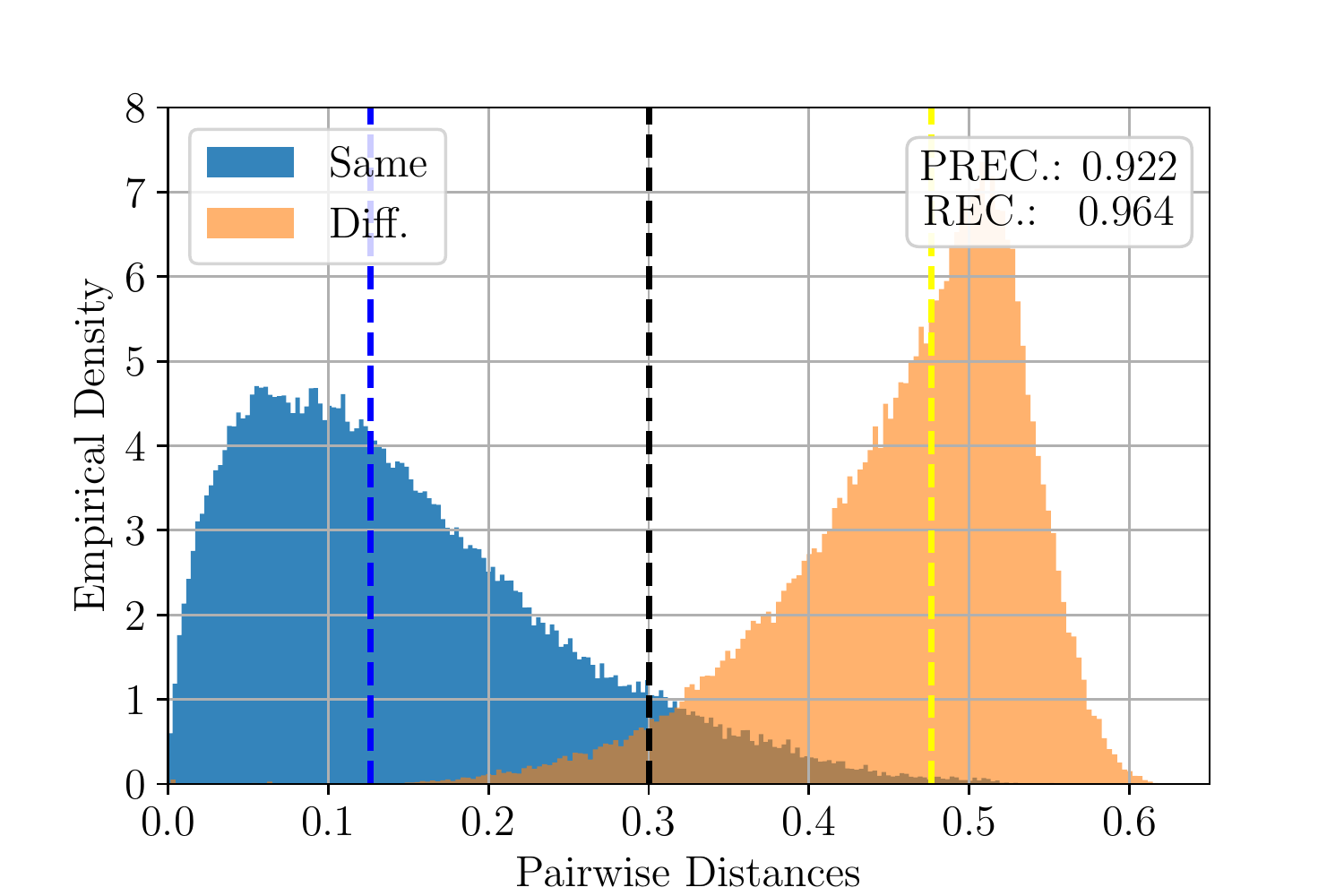}
		\caption{$f = 1.0$, $Q = \mathrm{id}.$}
		\label{fig:robustbess_hist_diff_scenes_NoneNone}
	\end{subfigure}
	\begin{subfigure}[t]{0.32\linewidth}
		\includegraphics[width=\linewidth]{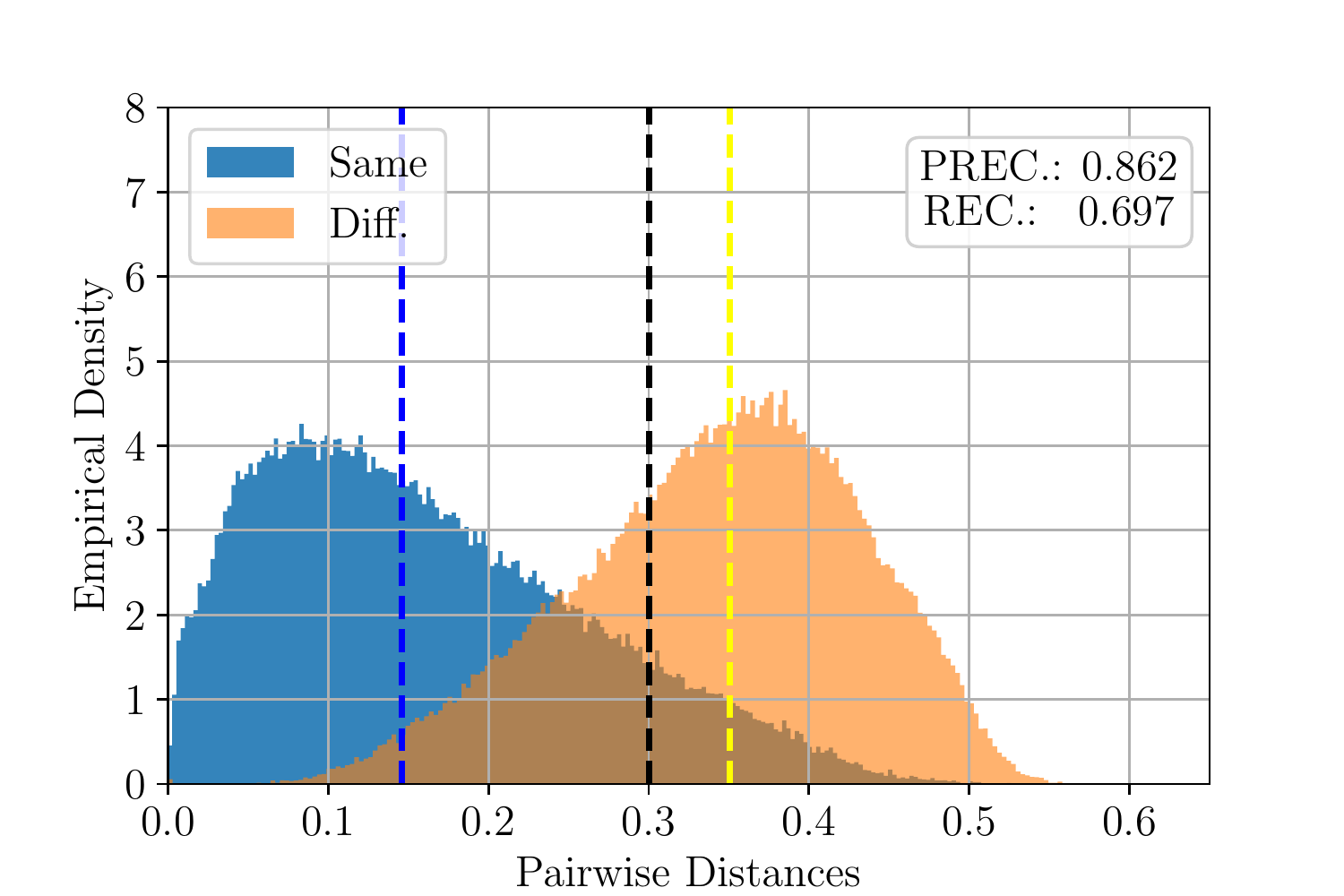}
		\caption{$f = 0.75$, $Q = 70$}
		\label{fig:robustbess_hist_diff_scenes_075_70}
	\end{subfigure}
	\begin{subfigure}[t]{0.32\linewidth}
		\includegraphics[width=\linewidth]{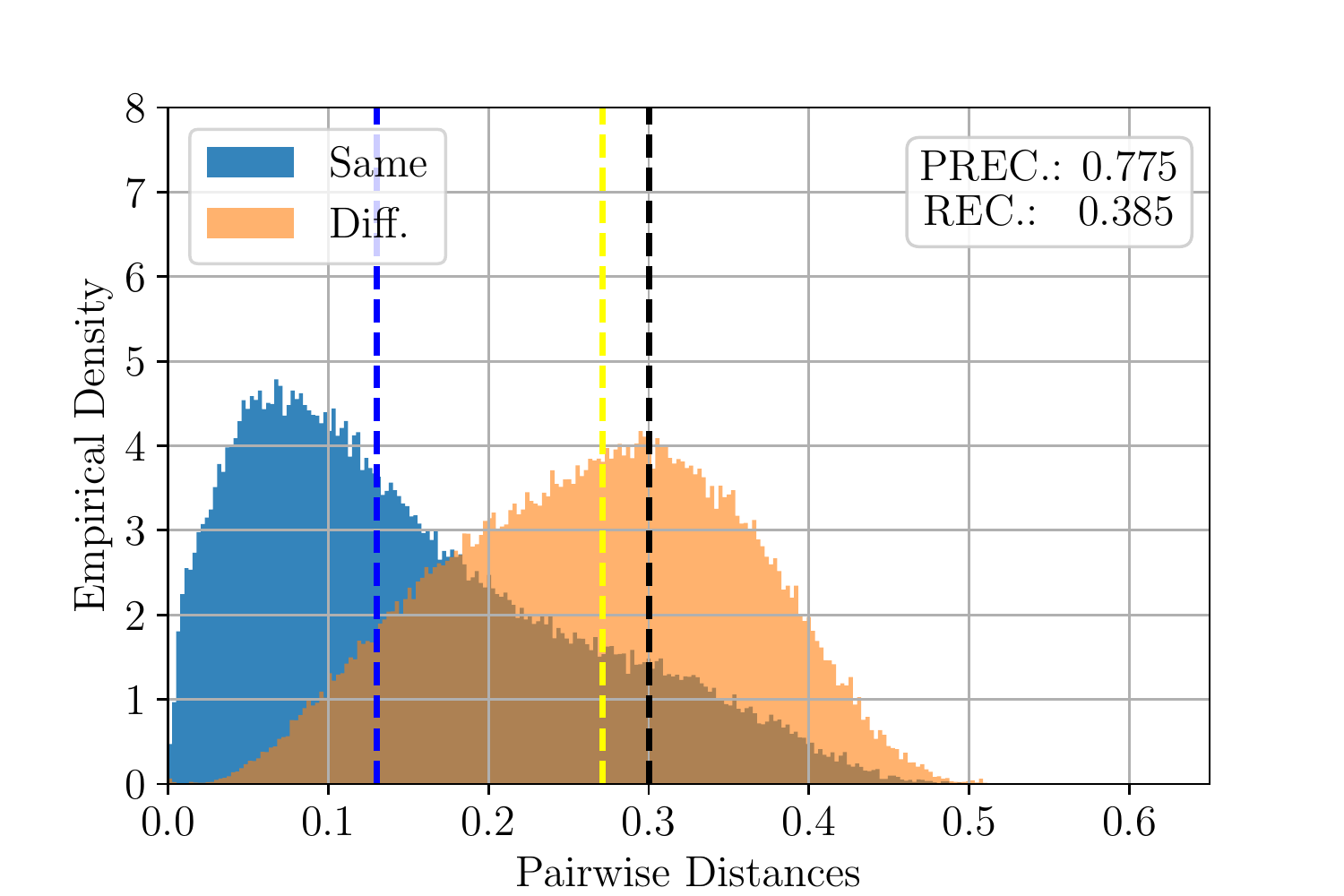}
		\caption{$f = 0.25$, $Q = 30$}
		\label{fig:robustbess_hist_diff_scenes_025_30}
	\end{subfigure}
	\caption{\edited{Empirical distributions of the pairwise embedding distances for similar and dissimilar imaging conditions from 50 pairs of different scenes with increasing post-processing strength from (a) to (c). Precision and recall are measured for a decision threshold fixed to $0.3$ (black dashed line). The colored dashed lines indicate the median of the corresponding distribution. Decreasing quality moves the distribution of distances for dissimilar pairs to smaller values, while the distribution for similar pairs is barely affected.}}
\label{fig:robustness_hist}
\end{figure*}

Figure~\ref{fig:validation_performance} visualizes the validation performance.
Figure~\ref{fig:val_dist_mat} shows the pairwise distance matrix averaged over all validation mini-batches.
The average distances for pairs of embeddings from identical imaging conditions (\edited{blocks of size $N_b^{\mathcal{P}}\times N_b^{\mathcal{P}} = 8\times 8$} on the diagonal) are consistently lower than the distances for different imaging conditions (off-diagonal blocks).
%This indicates that the distances of the embeddings represent the dissimilarities of the imaging conditions.
Pairs of different scenes exhibit larger distances than pairs that only differ in the camera pipeline.
Hence, different scenes are easier separable than different camera pipelines for identical scenes.
Figure~\ref{fig:val_distributions} shows the empirical distributions of pairwise distances for same and different imaging conditions, aggregated over all validation mini-batches.
The distribution for similar pairs concentrates on small distances around $0.1$, whereas the distribution for dissimilar pairs peaks at about $0.5$.
This corresponds to orthogonal embeddings, as enforced by the regularizer in Eqn.~(\ref{eq:regularizer}).
The validation ROC AUC is $0.967$, indicating a good separation of the distributions.

\subsection{Robustness of the Learned Embeddings}

This experiment shows the robustness against resizing and JPEG recompression, which are common post-processing operations on images from social networks.
We randomly select $100$ test scenes from $\mathcal{S}^{\mathrm{test}}$, and per scene a random camera pipeline to obtain $100$ test images.

Post-processing is applied for each combination of resizing factors $f\in\{1.25,
1.0, 0.75, 0.5, 0.25\}$ and JPEG quality levels $Q\in\{\mathrm{id.}, 90, 70,
50, 30\}$, where $f=1.0$ implies no resizing, $\mathrm{id.}$
(``idempotent'') implies no JPEG recompression. Then, each image is resized such that the larger dimension is $1536$ pixels, and embeddings are calculated
for $50$ random pairs of images and $50$ random
patches per image.
%The evaluation follows the algorithm in
%Sec.~\ref{sec:application_to_manipulation_detection} 
We then compute distributions of within-image pairwise distances and between-image
pairwise distances to represent similar and dissimilar imaging conditions.

\reedited{We measure the separation of the distributions in terms of True Positive Rate ($\mathrm{TPR}$) at a fixed False Alarm Rate ($\mathrm{FAR}$) of $5\%$, which we denote $\mathrm{TPR}_{5\%}$}. The results for embeddings on different scenes are shown in Fig.~\ref{fig:robustness_diff_scenes}. For images with no post-processing ($f=1.0$, $Q = \mathrm{id.}$), the distributions separate well with a \reedited{$\mathrm{TPR}_{5\%}$ of $0.929$}. 
For resampling factors down to $0.5$ and JPEG quality down to $50$, the \reedited{$\mathrm{TPR}_{5\%}$ is still $0.371$} or higher.
Extreme downsampling and compression with $f=0.25$ and $Q=30$ still yields an
\reedited{$\mathrm{TPR}_{5\%}$ of $0.190$}, which is remarkably high given these strong degradations.

The more challenging case of identical scene but different camera pipelines is
reported in Fig.~\ref{fig:robustness_same_scenes_0},
Fig.~\ref{fig:robustness_same_scenes_20}, and
Fig.~\ref{fig:robustness_same_scenes_50}. The three evaluation protocols only differ in
the minimum color difference of the evaluated image pairs, compare Eqn.~(\ref{eq:heur_min_dissim}), the remaining evaluation protocol is unchanged.
From left to right, patches are compared with a minimum \emph{Lab} color difference that is arbitrarily small ($\delta_{\mathrm{Lab}}^{\mathrm{min}}=0$), at least
$\delta_{\mathrm{Lab}}^{\mathrm{min}} = 20$ \reedited{($14.2\%$ of test patches)}, and at least
$\delta_{\mathrm{Lab}}^{\mathrm{min}} = 50$ \reedited{($0.883\%$ of test patches)}. 
%We consider unrestricted pairings (Fig.~\ref{fig:robustness_same_scenes_0}), and pairings where the image pairs have \emph{Lab} color distances (confer Eqn.~(\ref{eq:heur_min_dissim})) of at least $\delta_{\mathrm{Lab}}^{\mathrm{min}} = 20$ (Fig.~\ref{fig:robustness_same_scenes_20}) and $\delta_{\mathrm{Lab}}^{\mathrm{min}} = 50$ (Fig.~\ref{fig:robustness_same_scenes_50}) .

For arbitrary similar color distributions $\delta_{\mathrm{Lab}}^{\mathrm{min}}=0$, \reedited{$\mathrm{TPR}_{5\%}$ ranges between $0.625$ and $0.262$} for resampling factors down to $f=0.5$ and JPEG compression down to $Q=50$. Extreme degradations
with $f = 0.25$ and $Q=30$ still achieve a \reedited{$\mathrm{TPR}_{5\%}$ of $0.164$}. It is encouraging that this worst-case configuration still maintains good separability significantly above guessing chance\reedited{, which is equal to the fixed FPR of $5\%$}.
The easier case of a minimum color difference
$\delta_{\mathrm{Lab}}^{\mathrm{min}}=20$ between the pipelines exhibits \reedited{a $\mathrm{TPR}_{5\%}$ above $0.57$} throughout all post-processing parameters. Even larger
color differences $\delta_{\mathrm{Lab}}^{\mathrm{min}}=50$ can be almost
perfectly separated, with \reedited{$\mathrm{TPR}_{5\%}$ above $0.78$} across all degradations.
These results show that larger differences in camera pipelines lead to a better
separability. %Overall, the learned embeddings are remarkably robust to even 
%strong resampling and JPEG compression. 
Overall, the learned embeddings are remarkably robust to even strong resampling and JPEG compression. 

%===================================================================================

\edited{We now consider the behavior of the pairwise distance distributions for pairs from different images (compare Fig.~\ref{fig:robustness_diff_scenes}) for increasing post-processing strengths with respect to a fixed threshold $\vartheta$. A patch pair is classified as dissimilar (``positive''), if their distance in the learned metric space is larger than $\vartheta$, and as similar (``negative'') otherwise. The quality of the split can be measured with precision ($\mathrm{PREC}$) and recall ($\mathrm{REC}$),
%	Correctly classified pairs make up true positives ($\mathrm{TP}$) or true negatives ($\mathrm{TN}$), respectively, while incorrectly classified pairs lead to false positives ($\mathrm{FP}$) and false negatives ($\mathrm{FN}$), respectively. Then, 
\begin{equation}
	\mathrm{PREC} = \frac{\mathrm{TP}}{\mathrm{TP} + \mathrm{FP}}\enspace,
\end{equation}
\begin{equation}
	\mathrm{REC} = \frac{\mathrm{TP}}{\mathrm{TP} + \mathrm{FN}}\enspace,
\end{equation}
calculated from the true positives $\mathrm{TP}$, true negatives $\mathrm{TN}$, false positives $\mathrm{FP}$ and false negatives $\mathrm{FN}$. Based on the empirical distance distributions on the validation data, see Fig.~\ref{fig:val_distributions}, we fix $\vartheta=0.3$.

In case of no post-processing (Fig.~\ref{fig:robustbess_hist_diff_scenes_NoneNone}), the distributions are split well by $\vartheta=0.3$ with $\mathrm{PREC}$ $0.922$ and $\mathrm{REC}$ $0.964$. For $f=0.75$ and $Q=70$ (Fig.~\ref{fig:robustbess_hist_diff_scenes_075_70}), $\mathrm{PREC}$ is still at $0.862$, while REC is $0.697$. For extreme downsampling and compression with $f=0.25$ and $Q=30$, $\mathrm{PREC}$ still amounts to $0.775$, while REC is reduced to $0.385$. 

%This behavior can be attributed to a shift of the distance distribution for dissimilar imaging conditions to lower values with increasing post-processing, while the distribution for similar imaging conditions remains quite stable, as can be seen from the median of the respective distributions in Fig.~\ref{fig:robustness_hist}. 
This behavior can be attributed to a shift of the distribution of positives to lower values with increasing post-processing, while the distribution negatives remains quite stable, as can be seen from the median of the respective distributions in Fig.~\ref{fig:robustness_hist}.
As a consequence, more positives are classified negative, while the number of false positives remains low. This experiment shows that while the distributions can still be separated well for strong post-processing (compare Fig.~\ref{fig:robustness}), for optimal splitting the threshold needs to be adapted depending on the strength of the operations. However, note that splicing localization only requires relative pairwise distances. Hence, determining a globally optimal threshold is only necessary for splicing detection, not localization.
}

\begin{figure*}[h]
	\centering
	\begin{subfigure}[t]{\linewidth}
		\includegraphics[width=\textwidth,trim={0 14.5cm 0 0},clip]{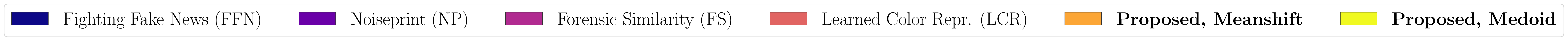}
	\end{subfigure}\\
	\vspace{0.2cm}
%	\begin{subfigure}[t]{0.19\textwidth}
%		\includegraphics[width=\textwidth]{figures/SplicingLocalization_Columbia_F1.pdf}
%		%		\caption{Columbia\cite{DBLP:conf/icmcs/HsuC06}}
%	\end{subfigure}
%	\begin{subfigure}[t]{0.19\textwidth}
%		\includegraphics[width=\textwidth]{figures/SplicingLocalization_DSO-1_F1.pdf}
%		%		\caption{DSO-1\cite{DBLP:journals/tifs/CarvalhoRAPR13}}
%	\end{subfigure}
%	\begin{subfigure}[t]{0.19\textwidth}
%		\includegraphics[width=\textwidth]{figures/SplicingLocalization_InTheWild_F1.pdf}
%		%		\caption{In-The-Wild\cite{DBLP:conf/eccv/HuhLOE18}}
%	\end{subfigure}
%	\begin{subfigure}[t]{0.19\textwidth}
%		\includegraphics[width=\textwidth]{figures/SplicingLocalization_SyntheticSplices_F1.pdf}
%		%		\caption{\newDatasetName}
%	\end{subfigure}
%	\begin{subfigure}[t]{0.198\textwidth}
%		\includegraphics[width=\textwidth]{figures/SplicingLocalization_AlignedScenesDIDB_F1.pdf}
%		%		\caption{Aligned Scenes\cite{DBLP:conf/icassp/HadwigerBPR19}}
%	\end{subfigure}\\
%	\begin{subfigure}[t]{0.19\textwidth}
%		\includegraphics[width=\textwidth]{figures/SplicingLocalization_Columbia_MCC.pdf}
%		\caption{Columbia\cite{DBLP:conf/icmcs/HsuC06}}
%		\label{fig:loc_columbia}		
%		%		\caption{Columbia\cite{DBLP:conf/icmcs/HsuC06}}
%	\end{subfigure}
	\begin{subfigure}[t]{0.24\textwidth}
		\includegraphics[width=\textwidth]{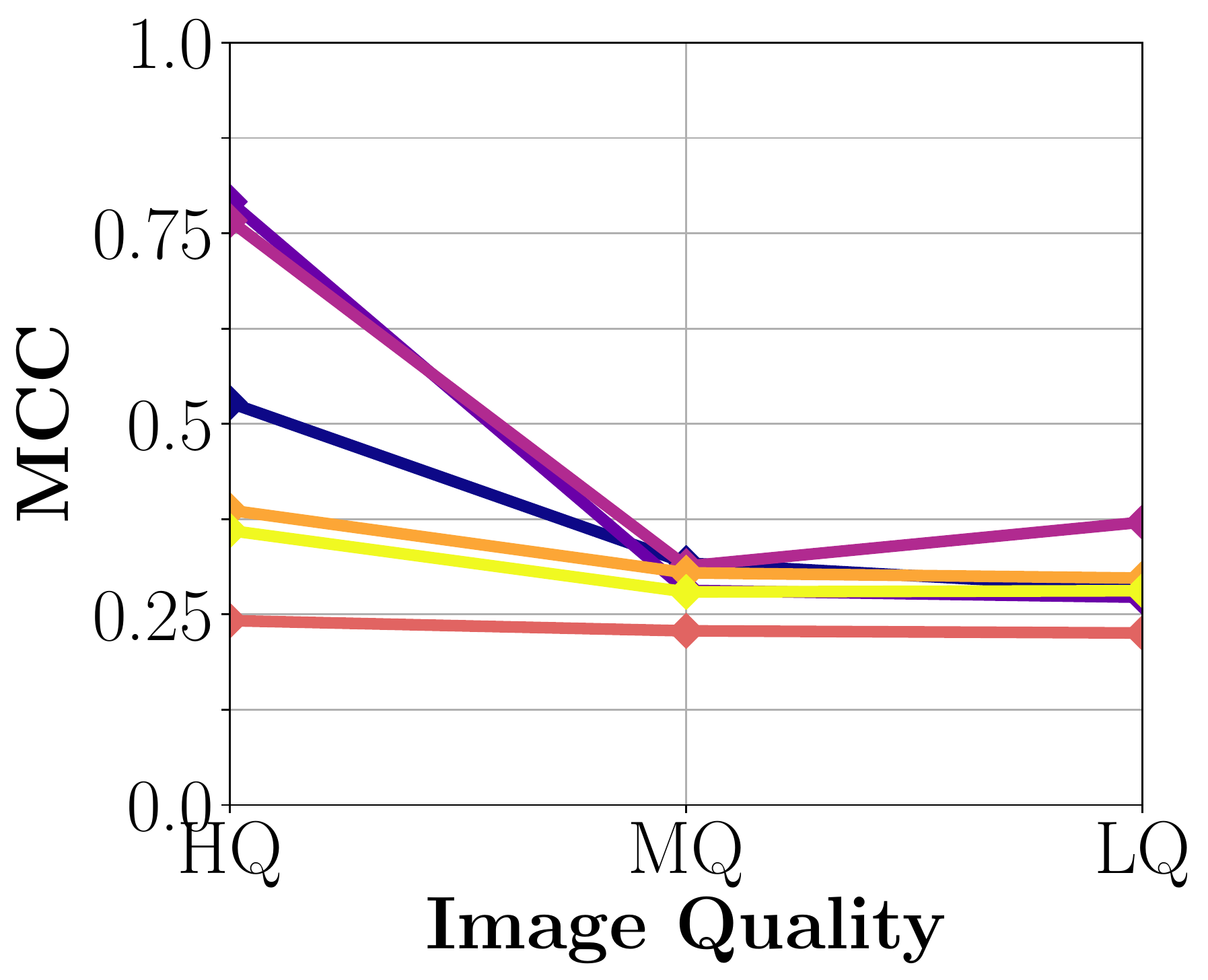}
		%		\caption{DSO-1\cite{DBLP:journals/tifs/CarvalhoRAPR13}}
		\caption{DSO-1\cite{DBLP:journals/tifs/CarvalhoRAPR13}}
		\label{fig:loc_dso1}		
	\end{subfigure}
	\begin{subfigure}[t]{0.24\textwidth}
		\includegraphics[width=\textwidth]{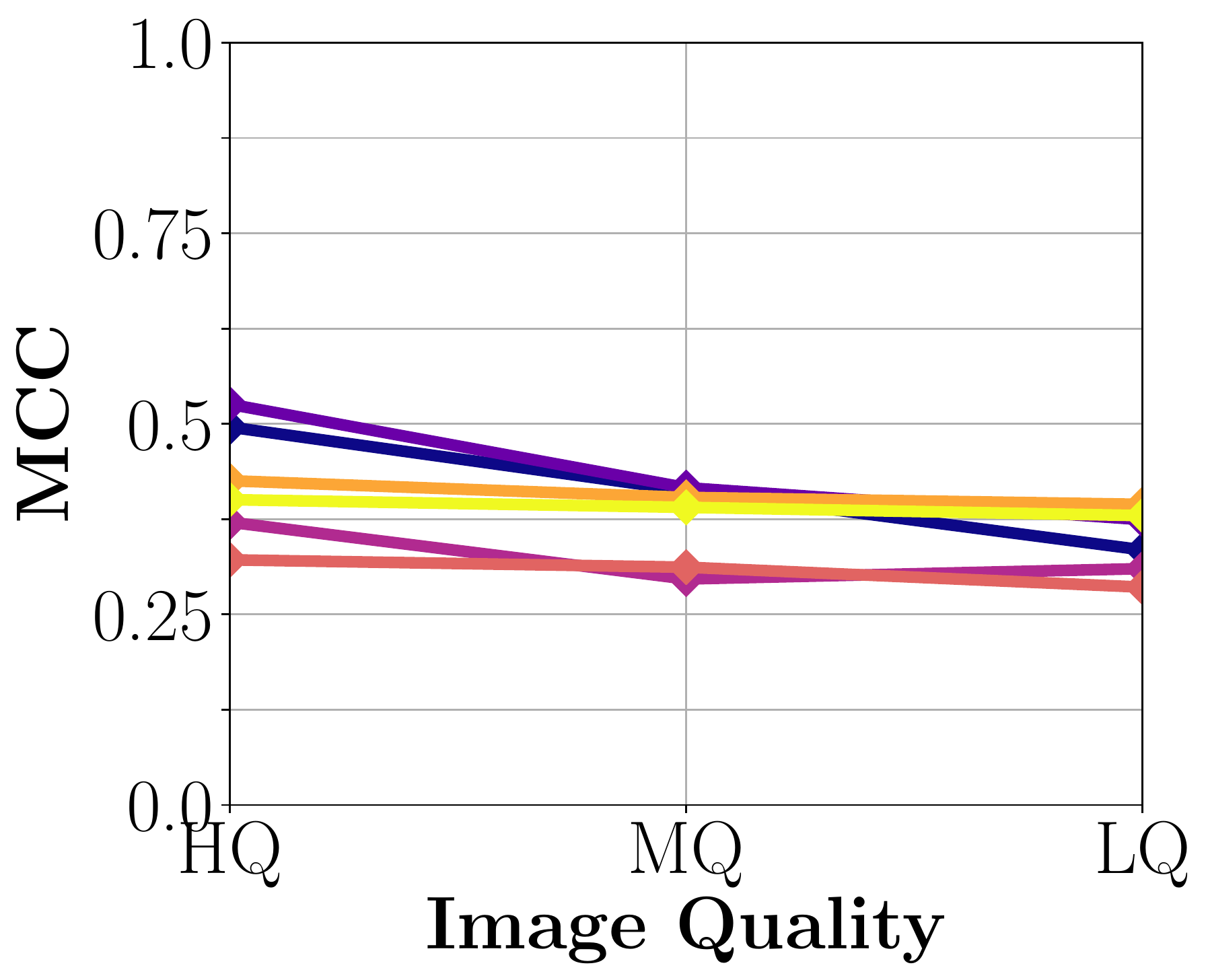}
		\caption{In-The-Wild\cite{DBLP:conf/eccv/HuhLOE18}}
		\label{fig:loc_inthewild}		
		%		\caption{In-The-Wild\cite{DBLP:conf/eccv/HuhLOE18}}
	\end{subfigure}
	\begin{subfigure}[t]{0.24\textwidth}
		\includegraphics[width=\textwidth]{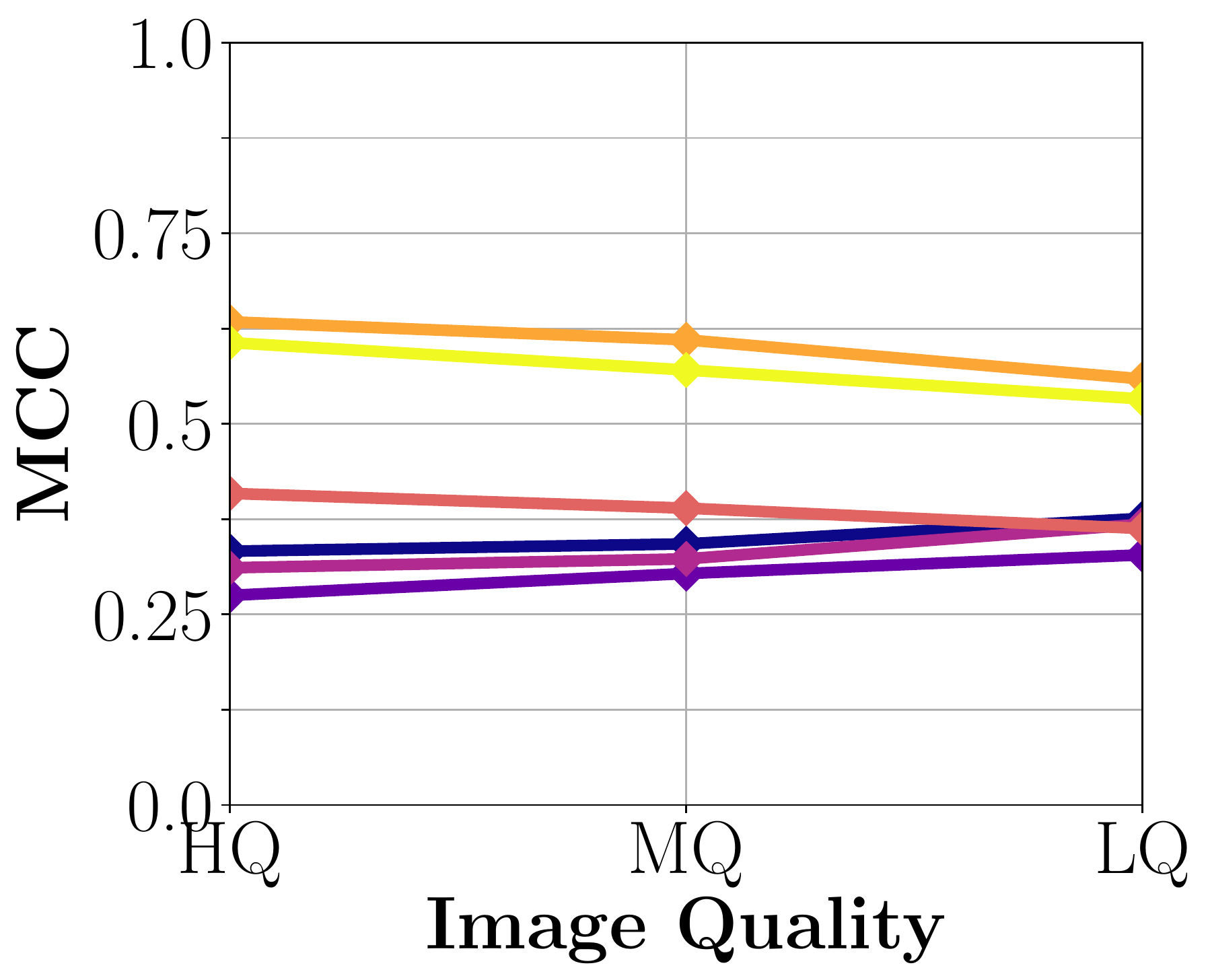}
		\caption{\newDatasetName}
		\label{fig:loc_synthspl}		
		%		\caption{\newDatasetName}
	\end{subfigure}
	\begin{subfigure}[t]{0.24\textwidth}
		\includegraphics[width=\textwidth]{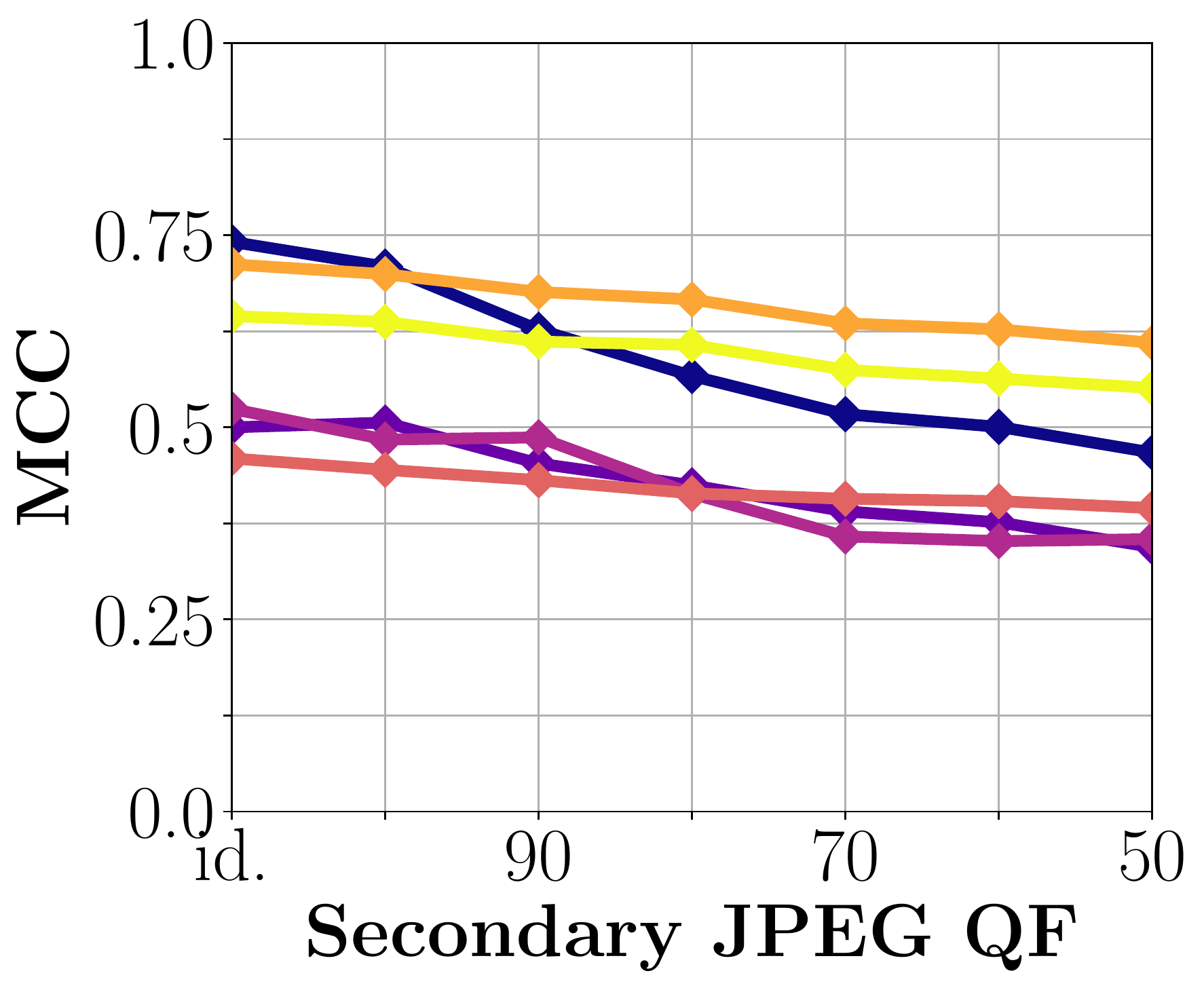}
		\caption{Aligned Scenes\cite{DBLP:conf/icassp/HadwigerBPR19}}
		\label{fig:loc_alignedscenes}
		
		%		\caption{Aligned Scenes\cite{DBLP:conf/icassp/HadwigerBPR19}}
	\end{subfigure}\\
%	\begin{subfigure}[t]{0.19\textwidth}
%		\includegraphics[width=\textwidth]{figures/SplicingLocalization_Columbia_mAP.pdf}
%		\caption{Columbia\cite{DBLP:conf/icmcs/HsuC06}}
%		\label{fig:loc_columbia}
%	\end{subfigure}
%	\begin{subfigure}[t]{0.19\textwidth}
%		\includegraphics[width=\textwidth]{figures/SplicingLocalization_DSO-1_mAP.pdf}
%		\caption{DSO-1\cite{DBLP:journals/tifs/CarvalhoRAPR13}}
%		\label{fig:loc_dso1}
%	\end{subfigure}
%	\begin{subfigure}[t]{0.19\textwidth}
%		\includegraphics[width=\textwidth]{figures/SplicingLocalization_InTheWild_mAP.pdf}
%		\caption{In-The-Wild\cite{DBLP:conf/eccv/HuhLOE18}}
%		\label{fig:loc_inthewild}
%	\end{subfigure}
%	\begin{subfigure}[t]{0.19\textwidth}
%		\includegraphics[width=\textwidth]{figures/SplicingLocalization_SyntheticSplices_mAP.pdf}
%		\caption{\newDatasetName}
%		\label{fig:loc_synthspl}
%	\end{subfigure}
%	\begin{subfigure}[t]{0.198\textwidth}
%		\includegraphics[width=\textwidth]{figures/SplicingLocalization_AlignedScenesDIDB_mAP.pdf}
%		\caption{Aligned Scenes\cite{DBLP:conf/icassp/HadwigerBPR19}}
%		\label{fig:loc_alignedscenes}
%	\end{subfigure}
	\caption{Performance for splicing \emph{localization} using the %$F_1$-score (first row), 
		Matthews Correlation Coefficient %(second row) and Mean Average Precision (third row) 
		on various datasets. %All three metrics exhibit similar results. 
		The proposed method (orange/yellow) compares particularly well on very low-quality images (see text for details).}
	\label{fig:results_localization}
\end{figure*}

\subsection{Comparison for Splicing Detection and Localization}
\label{sec:splicing_det_loc}

This Section consists of five parts. First, we present the benchmark datasets
and the algorithms for comparison. Then, we evaluate the localization
performance by presenting quantitative and qualitative results, which is the
primary focus of this method. We finally report detection performances in
comparison to related work.

\begin{table}[t]
	\centering
	\caption{Overview of the benchmark datasets. The second column details the number of pristine, resp. manipulated images \emph{per} quality level, i.e., the total number of test images per dataset is obtained as product with the number of quality levels.}
	\label{tab:datasets}
	\begin{tabular}[h]{|c|c|c|c|}
		\hline
		\textbf{Name} & \textbf{\#Prist. / \#Manip.} & \textbf{\#Qualities} & \textbf{Tasks}\\ \hhline{|=|=|=|=|}
		In-The-Wild\cite{DBLP:conf/eccv/HuhLOE18} & $0$ / $201$ & 3 & Loc.\\ \hline
		DSO-1\cite{DBLP:journals/tifs/CarvalhoRAPR13} & $100$ / $100$ & 3 & Loc. / Det.\\ \hline
		% Columbia\cite{DBLP:conf/icmcs/HsuC06} & $183$ / $180$ & 3 & Loc. / Det.\\ \hline
		Aligned Scenes\cite{DBLP:conf/icassp/HadwigerBPR19} & $166$ / $166$ & 11 & Loc. / Det.\\ \hline
		\newDatasetName & 200 / 200 & 3 & Loc. / Det.\\ \hline
	\end{tabular}
\end{table}

\paragraph{Benchmark Datasets} We use a total of four datasets listed in Tab.~\ref{tab:datasets}.
Three datasets have been previously published, namely %``Columbia''~\cite{DBLP:conf/icmcs/HsuC06},
``DSO-1''~\cite{DBLP:journals/tifs/CarvalhoRAPR13}, ``Aligned Scenes'' \cite{DBLP:conf/icassp/HadwigerBPR19} and `In-The-Wild''\cite{DBLP:conf/eccv/HuhLOE18}.
%``Columbia'', 
``DSO-1'', and ``Aligned Scenes'' support evaluation of
manipulation localization and detection.  ``In-the-Wild'' contains no pristine
images, and can hence only assess localization performance.  %``Columbia'' is an early benchmark consisting of relatively simple splices. 
``DSO-1'' contains splices of people.  ``Aligned Scenes'' consists of splices where rectangular image blocks are replaced by an image block with the same scene content, but recorded with a different camera.  ``In-The-Wild'' contains a collection of online images with splicing and other manipulations, and manually annotated ground truth.

We further create a fourth dataset called ``SplicedColorPipeline'' to evaluate
detection and localization for the specific case of different camera pipelines.
The dataset consists of $200$ pristine images, consisting of randomly selected
scenes from $\mathcal{S}^{\mathrm{test}}$, each developed with a randomly
selected camera pipeline.
It also consists of $200$ manipulations, where one region of the scene is
replaced by identical scene content, but developed with a randomly selected
different camera pipeline.
The region is a randomly selected superpixel with minimum size of $5\cdot 10^4$
pixels (i.e., about three patch sizes), obtained with the segmentation
algorithm by Felzenszwalb and
Huttenlocher~\cite{DBLP:journals/ijcv/FelzenszwalbH04} with scale parameter
$10$ and $\sigma=0.5$.
We replace a region such that the average \emph{Lab}-distance between the original and inserted region is at least $5$ to simulate local differences in within-camera color processing.

The images of all datasets are post-processed to simulate quality degradations
of images distributed over the internet. The
``Aligned Scenes'' dataset is first downsampled and then recompressed with JPEG
qualities from $100$ down to $10$ in steps of
$10$~\cite{DBLP:conf/icassp/HadwigerBPR19}.
The remaining datasets are prepared in three variants:
The high-quality (HQ) variant contains each image as-is.
For the medium-quality (MQ) variant, each image is resized to a larger
dimension of $1200$ pixels, and JPEG compressed with quality $75$.
For the low-quality (LQ) variant,  each image is resized to a larger
dimension of $800$ pixels, and JPEG compressed with quality $50$.
%
%high-quality here, each image is
%resized {\color{red}{Factor?}}, and added to the dataset with single JPEG
%compression of quality {\color{red}{quality?}}, and also with second JPEG
%compressions with JPEG qualities from $100$ down to $10$ in steps of $10$.
%That is, besides evaluating on the images ``as is'' (high quality, HQ), we further test on medium quality (MQ) and low quality (LQ) versions resized to $1200$, resp. $800$ pixels in the larger dimension and JPEG compressed with quality $75$, resp. $50$.
%This allows us to probe the algorithm for conditions similar to the ones met for images from social networks. 

\paragraph{Algorithms for Comparison} The proposed method is evaluated for medoid-based scores (MED), and MeanShift-based aggregation (MSA), see Sec.~\ref{sec:application_to_manipulation_detection}.
We benchmark our algorithm with other works not requiring manipulated training images.
In particular, we compare against ``Noiseprint'', (\noisep)~\cite{DBLP:journals/tifs/CozzolinoV20}, ``Fighting Fake News'' (\ffn)~\cite{DBLP:conf/eccv/HuhLOE18}, ``Forensic Similarity'' (\forsim)~\cite{DBLP:journals/tifs/MayerS20} and ``Learned Color Representations'' (\lcr)~\cite{DBLP:conf/icassp/HadwigerBPR19}. 

\edited{NP learns an image-to-image mapping, and enforces output distances based on the agreement of the cameras that recorded the images. FFN uses a Siamese CNN and derives labels from image metadata. FS is based on a Siamese network that classifies the camera and post-processing consistency of image patches. LCR learns a color descriptor in a supervised fashion, and checks the consistency of this cue throughout an image.}

For the former two algorithms, we take the available implementations from their official repositories. For \forsim, we use the available trained CNN in the variant with input patch sizes $128\times 128$. After filtering out patches with inadequate entropy~\cite{DBLP:journals/tifs/MayerS20} from a test image, we aggregate pairwise predicted similarities of that CNN to a heatmap using MeanShift as in~\cite{DBLP:conf/eccv/HuhLOE18}.
For \lcr, we use our own implementation, and aggregate pairwise similarity scores in the same way using MeanShift to obtain heatmaps. 

\paragraph{Splicing Localization} Localization performance is measured with averaged scores over all images per dataset.
We use %the $F_1$-score, 
Matthews Correlation Coefficient ($\mathrm{MCC}$), %and Average Precision ($\mathrm{AP}$), 
defined as
%\begin{equation}
%F_1 = 2\cdot\frac{\mathrm{PREC}\cdot\mathrm{REC}}{\mathrm{PREC}+\mathrm{REC}}\enspace, 
%\end{equation}
\begin{equation}
	\mathrm{MCC} = \frac{\mathrm{TP}\cdot \mathrm{TN} - \mathrm{FP}\cdot\mathrm{FN}}{\sqrt{(\mathrm{TP} + \mathrm{FP})(\mathrm{TP} + \mathrm{FN})(\mathrm{TN} + \mathrm{FP})(\mathrm{TN} + \mathrm{FN})}}\enspace,
\end{equation}
%\begin{equation}
%	\mathrm{AP} = \sum_{j} (\mathrm{REC}_j-\mathrm{REC}_{j-1})\cdot\mathrm{PREC}_{j}\enspace.
%\end{equation}
%Here, PREC denotes precision, and REC denotes recall. AP integrates over the precision/recall curve, which is indicated by index $j$.
%refers to indicates the Precision (PREC) and Recall (REC) value for the
%range of thresholds {\color{red}{unscharf}}, and
%\begin{equation}
%\mathrm{PREC} = \frac{TP}{TP + FP}\enspace,
%\end{equation}
%\begin{equation}
%\mathrm{REC} = \frac{TP}{TP + FN}\enspace,
%\end{equation}
%are calculated from the true positives $\mathrm{TP}$, true negatives
%$\mathrm{TN}$, false positives $\mathrm{FP}$ and false negatives $\mathrm{FN}$.
% 
For %the $F_1$-score and $\mathrm{MCC}$, we use for 
each image, we use the decision threshold that
maximizes the MCC score~\cite{DBLP:conf/eccv/HuhLOE18, DBLP:journals/tifs/CozzolinoV20}.
Further, as the algorithms cannot distinguish whether foreground or background
are inserted, we evaluate heatmap $\mathcal{H}$ and its inverse $1 -
\mathcal{H}$, and use the better result~\cite{DBLP:conf/eccv/HuhLOE18,
DBLP:journals/tifs/CozzolinoV20}. \edited{Our method is tailored to discrepancies in color imaging conditions and low quality images. However, for better comparison we also show results on general splices and high quality images}. The results are shown in Fig.~\ref{fig:results_localization}.%, with one row per metric and one column per dataset.  Qualitatively, the results for all three metrics are highly similar. We hence discuss the results for $\mathrm{MCC}$. 

%On the Columbia dataset (Fig.~\ref{fig:loc_columbia}), the proposed algorithm performs well (MSA HQ: $0.723$)\footnote{The values in brackets denote the $\mathrm{MCC}$ score.} and is robust, outperforming the other algorithms for LQ (MSA LQ: $0.678$).
%The performance of the other methods except of \lcr{} decreases quickly with decreasing quality.

On the DSO-1 dataset (Fig.~\ref{fig:loc_dso1}), the best results are obtained for the purely statistics-based approaches \noisep{} (HQ: $0.791$) and \forsim{} (HQ: $0.767$). The proposed method with MSA achieves a performance of $0.387$ for HQ images. The proposed method is challenged by the fact that both the background and the spliced part of the image are in many cases recorded with the same camera and with flash light \edited{as dominant light source}~\cite{DBLP:journals/tifs/CarvalhoRAPR13},  which violates our assumption \edited{that the color formation differs}. Surprisingly, all methods quickly deteriorate with decreasing image quality, which may be due to degradation-induced removal of tell-tale manipulation artifacts.

On the In-The-Wild dataset (Fig.~\ref{fig:loc_inthewild}), \noisep{} performs best for HQ images ($0.524$), followed by \ffn{} ($0.496$) and the proposed approach with MSA ($0.425$). The performance of \noisep{} and \ffn{} decreases with increasing degradations (\noisep{} LQ: $0.364$, \ffn{} LQ: $0.325$). The proposed method remains robust (MSA MQ: $0.403$, LQ MSA: $0.394$), and performs best on low-quality images.

On the \newDatasetName{} dataset (Fig.~\ref{fig:loc_synthspl}), the proposed
method (MSA) performs best throughout all qualities  with $\mathrm{MCC}$
$0.634$ and $0.559$ for MQ and LQ, respectively.
With a large margin, the second best performance is obtained by \lcr{} (HQ: $0.409$, MQ: $0.363$).

On the Aligned Scenes dataset (Fig.~\ref{fig:loc_alignedscenes}), the proposed
method achieves the best results for JPEG qualities of $90$ ($0.676$) and
below.

\edited{The proposed method outperforms some state-of-the-art approaches for HQ images. However, it particularly excells in its robustness against image
degradations, where other methods rapidly lose performance. The proposed method performs best on low-quality images, }which are notoriously difficult to analyze for statistical approaches. The \newDatasetName{} and
Aligned Scenes datasets specifically benchmark differences in the color
formation. The proposed methods is particularly well-suited to detect these
traces, for which it outperforms \edited{all} remaining methods also for high-quality images.

\begin{figure}[t]
	\centering
	\begin{tabular}{>{\centering}p{0.16\linewidth}@{}>{\centering}p{0.21\linewidth}@{}>{\centering}p{0.20\linewidth}@{}>{\centering}p{0.22\linewidth}@{}>{\centering\arraybackslash}p{0.14\linewidth}}
		& &  \multicolumn{3}{c}{\small{Results on}} \\\cline{3-5}
		\small{Image\phantom{m}} & \small{Ground truth} & \small{HQ} & \small{MQ} & \small{LQ} \\
	\end{tabular}
	\includegraphics[width=0.19\linewidth,height=0.28\linewidth]{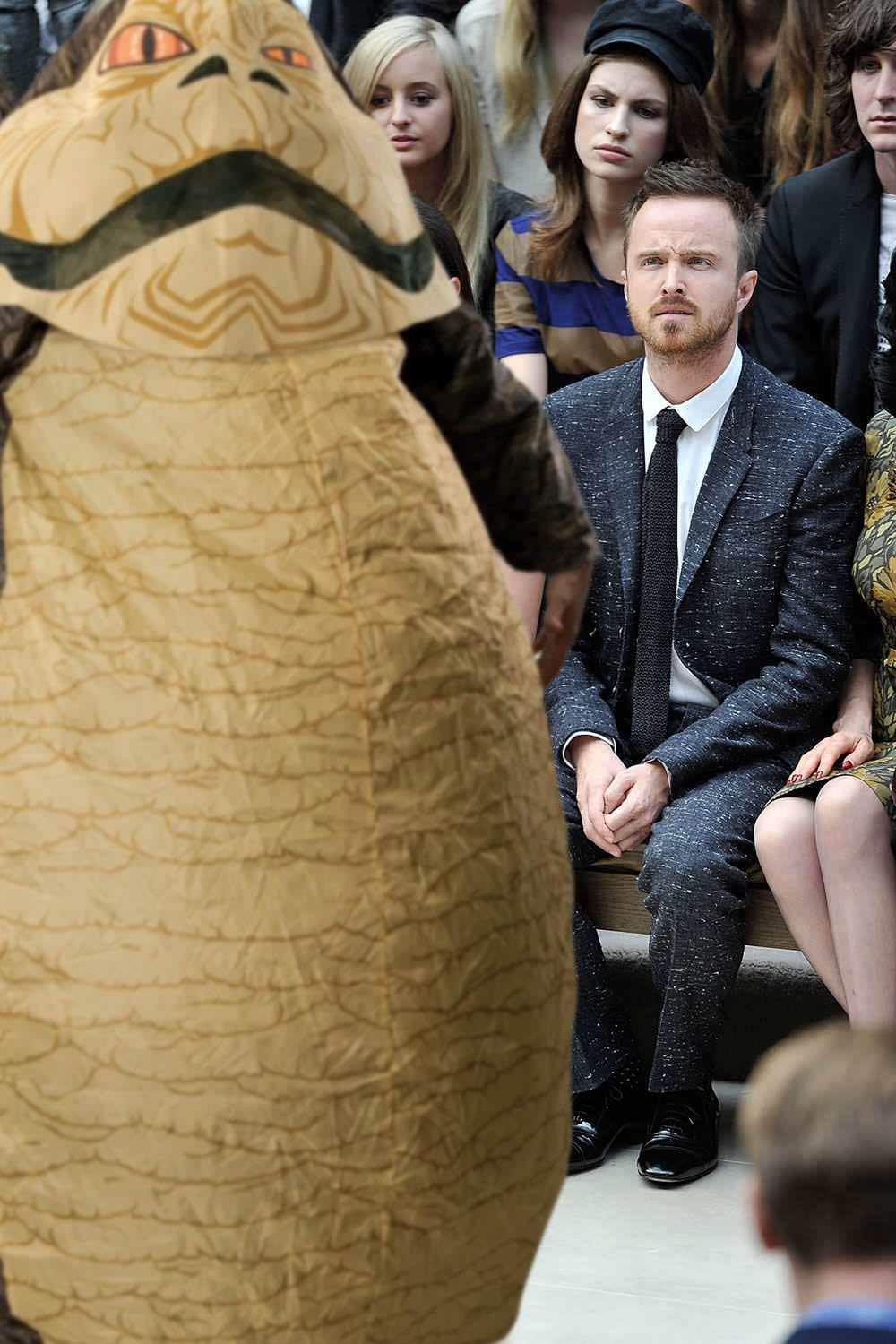}
	\includegraphics[width=0.19\linewidth,height=0.28\linewidth]{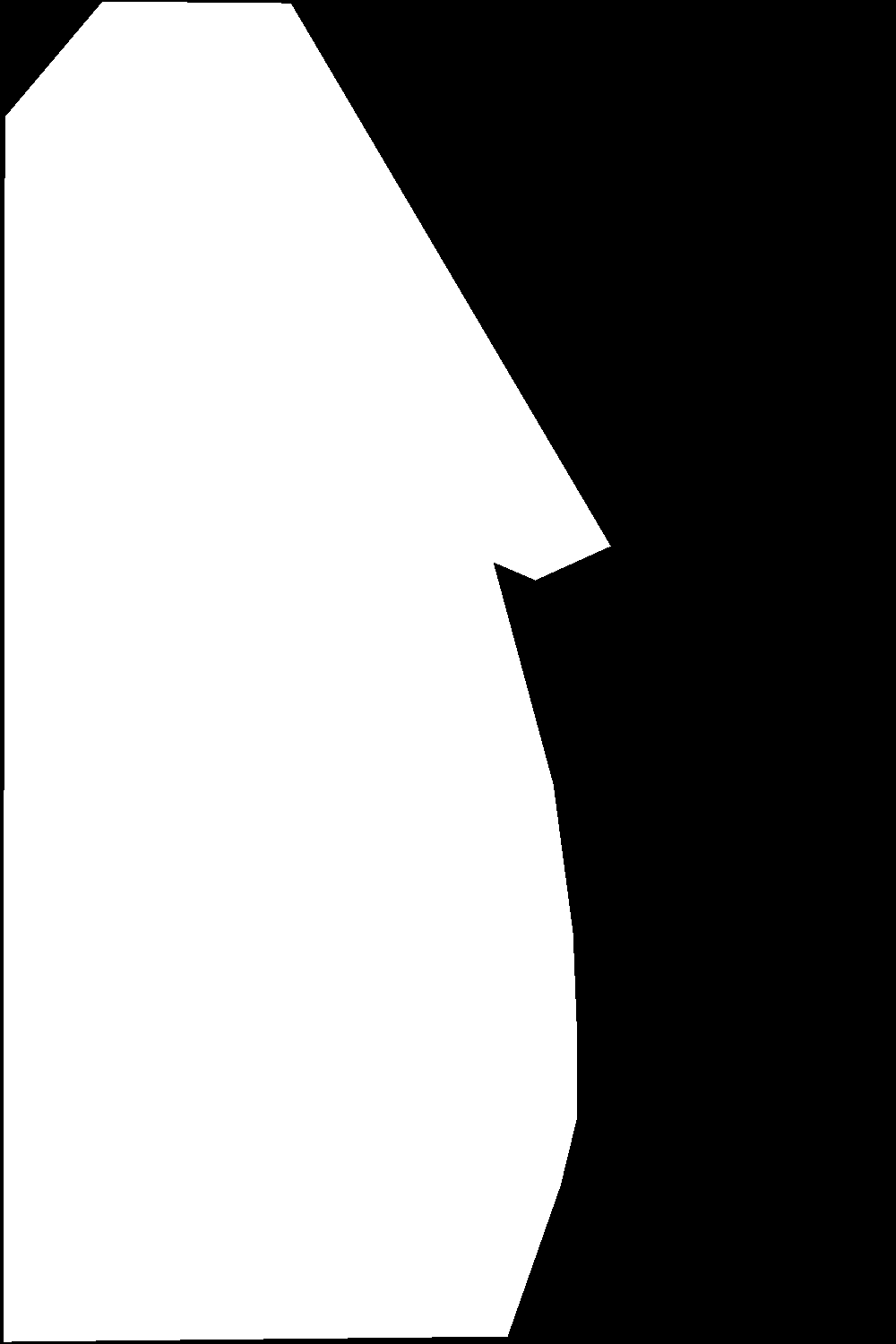}	
	\includegraphics[width=0.19\linewidth,height=0.28\linewidth]{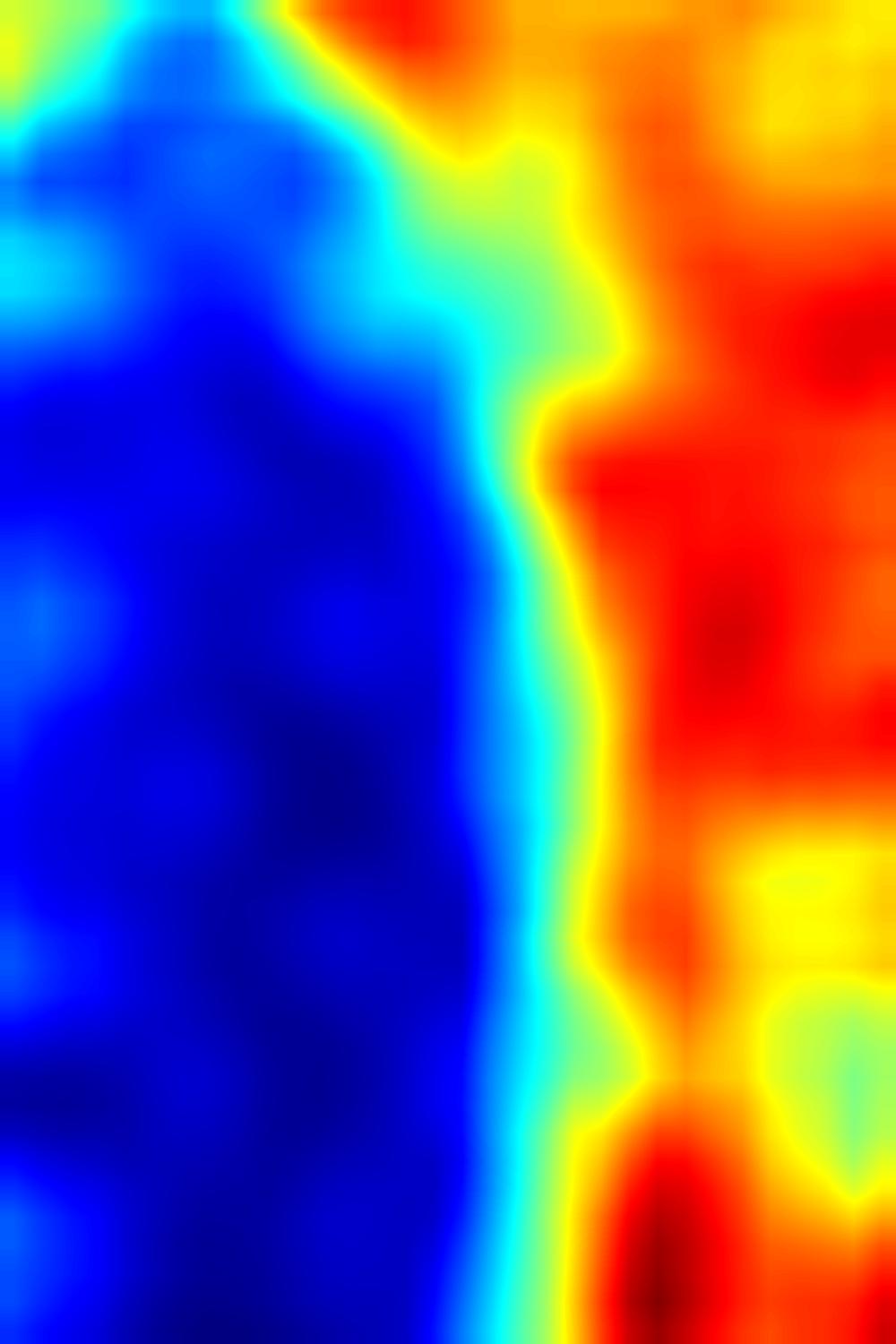}
	\includegraphics[width=0.19\linewidth,height=0.28\linewidth]{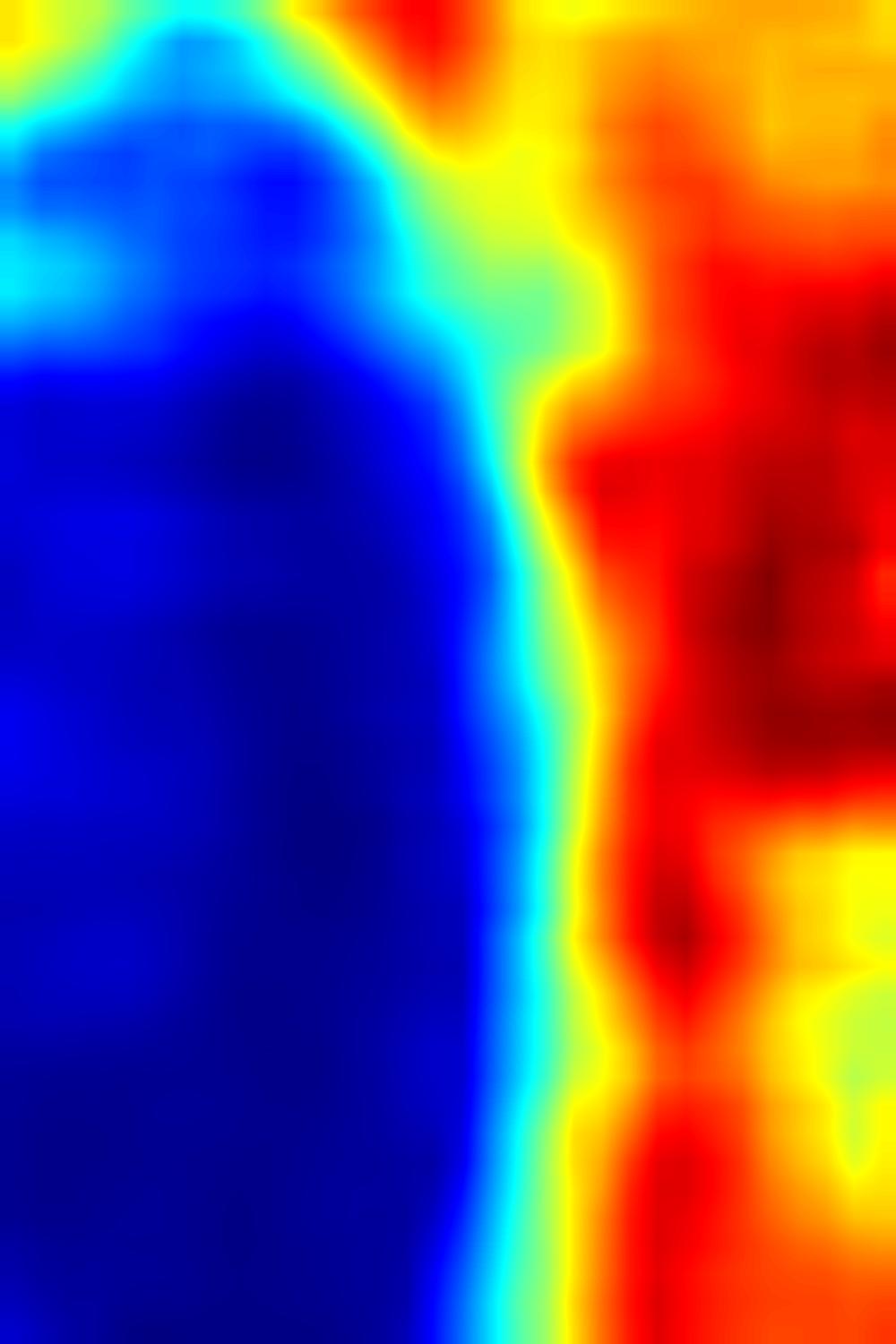}
	\includegraphics[width=0.19\linewidth,height=0.28\linewidth]{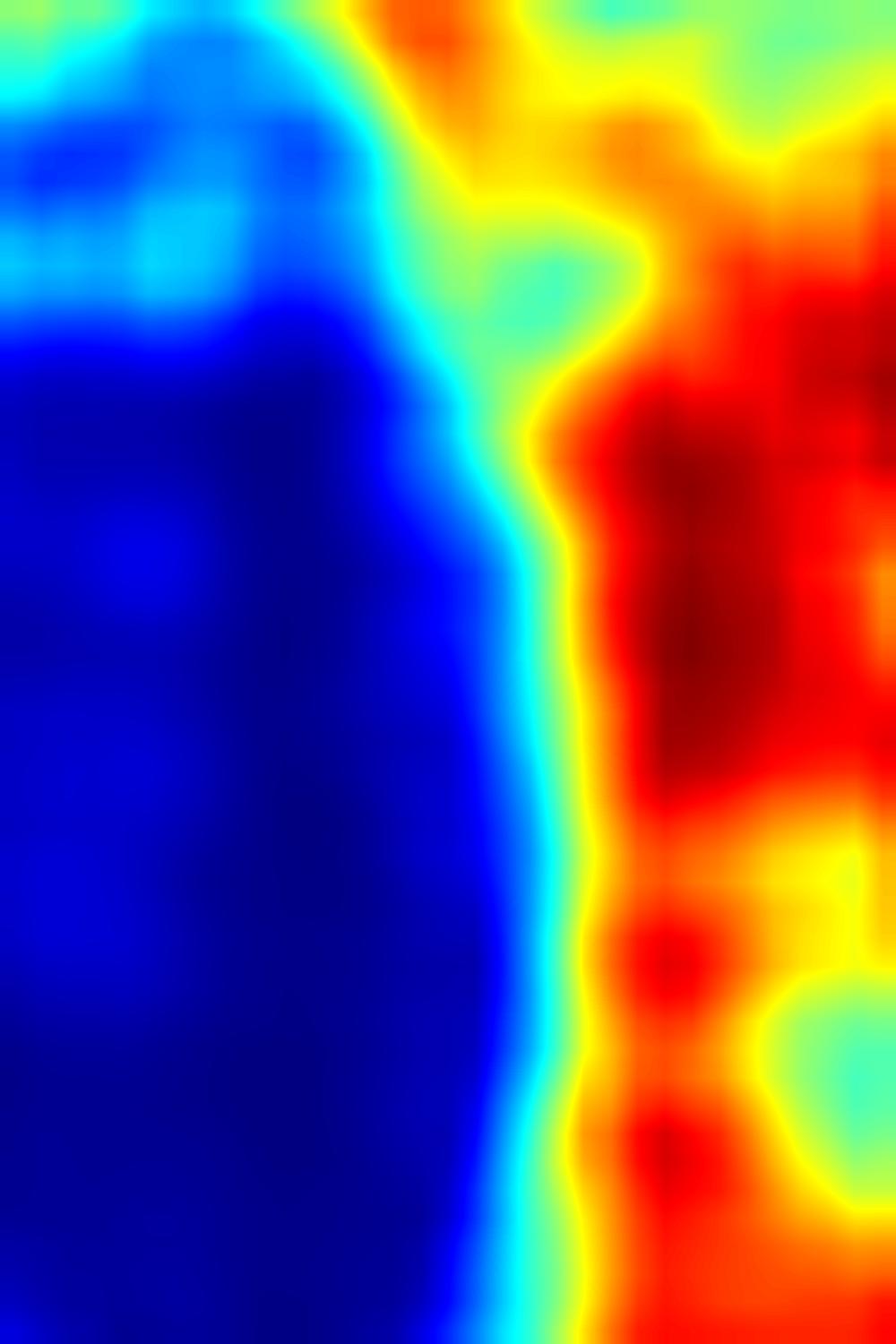}\\
	\vspace{0.1cm}
	\includegraphics[width=0.19\linewidth,height=0.28\linewidth]{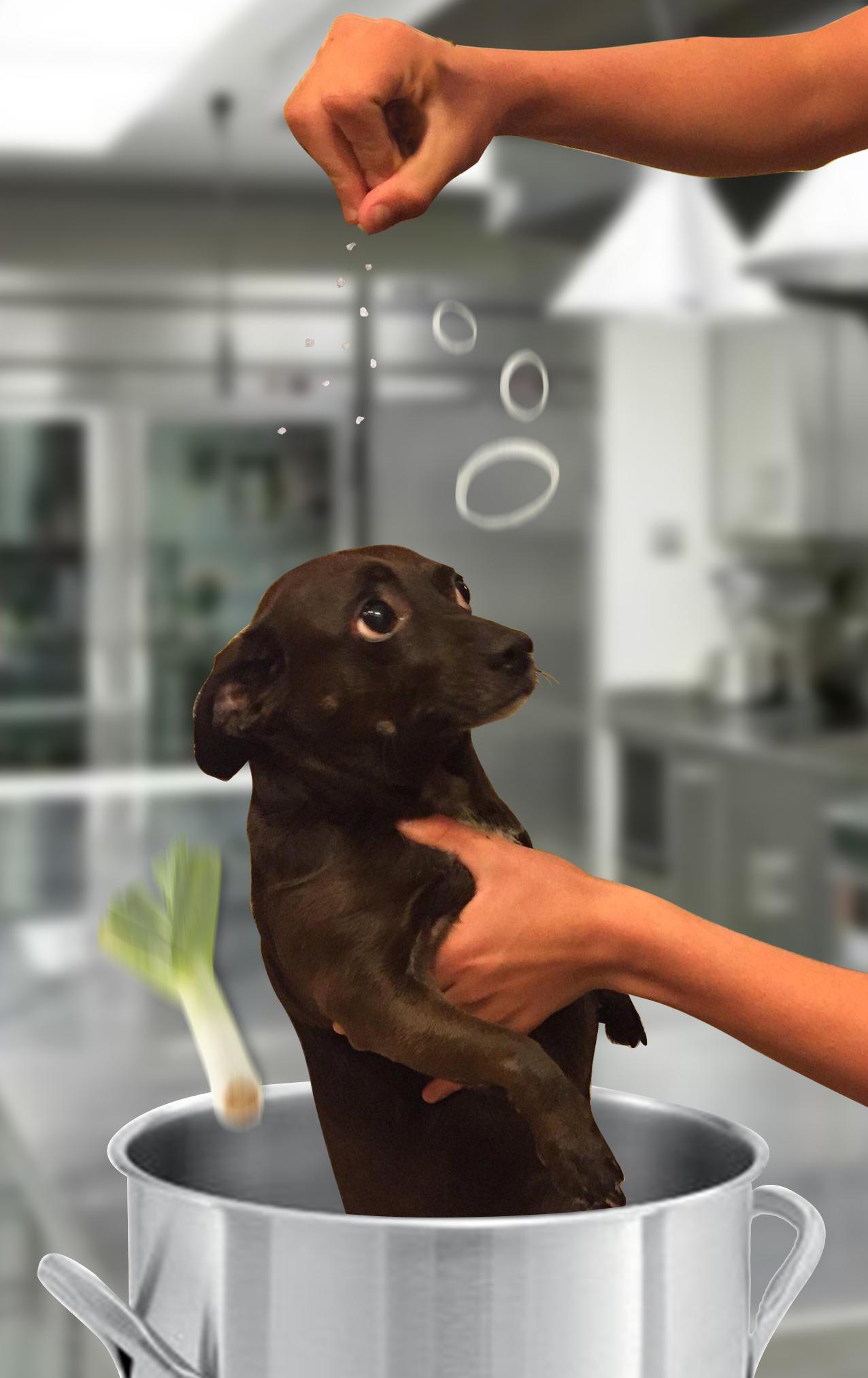}
	\includegraphics[width=0.19\linewidth,height=0.28\linewidth]{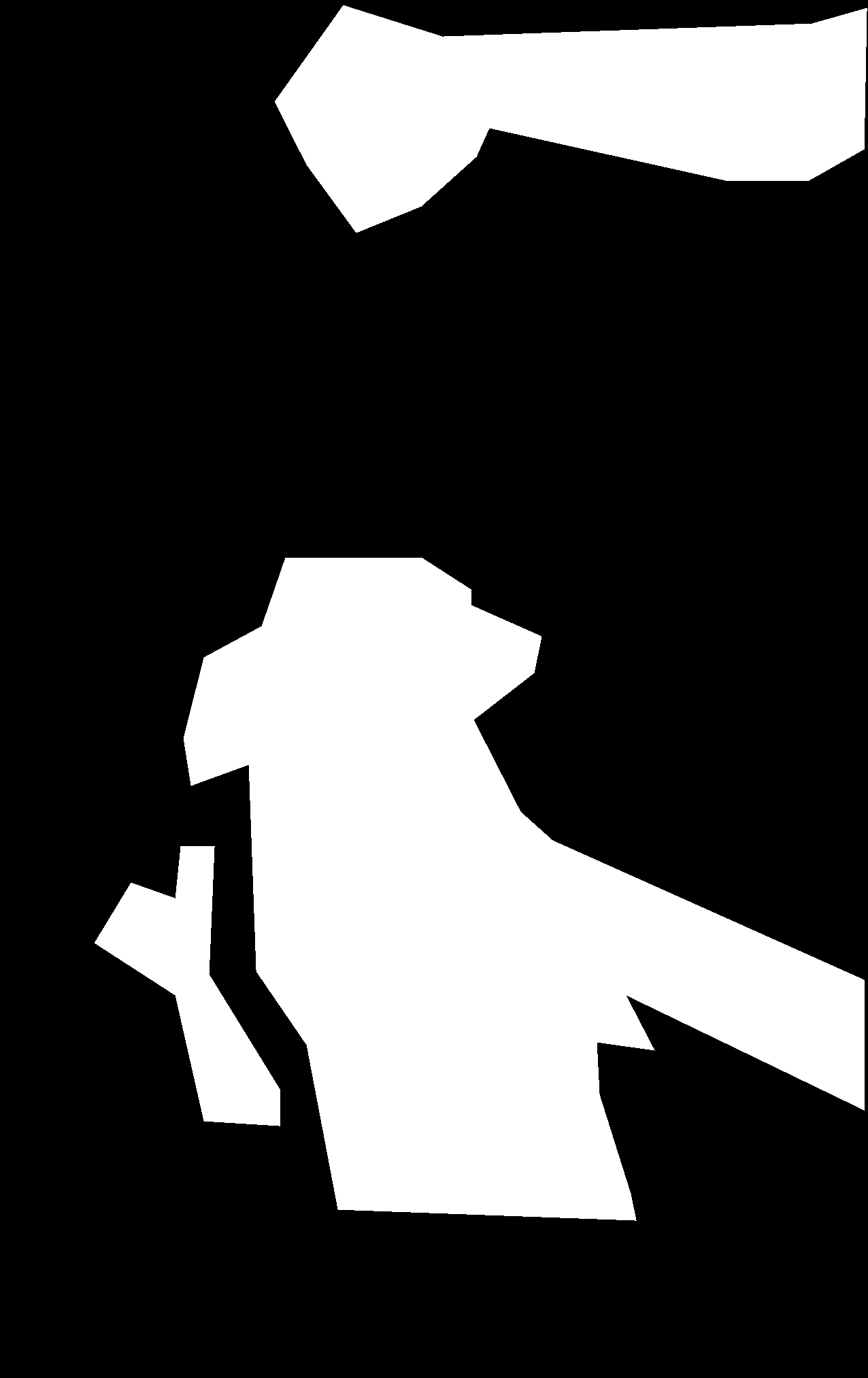}
	\includegraphics[width=0.19\linewidth,height=0.28\linewidth]{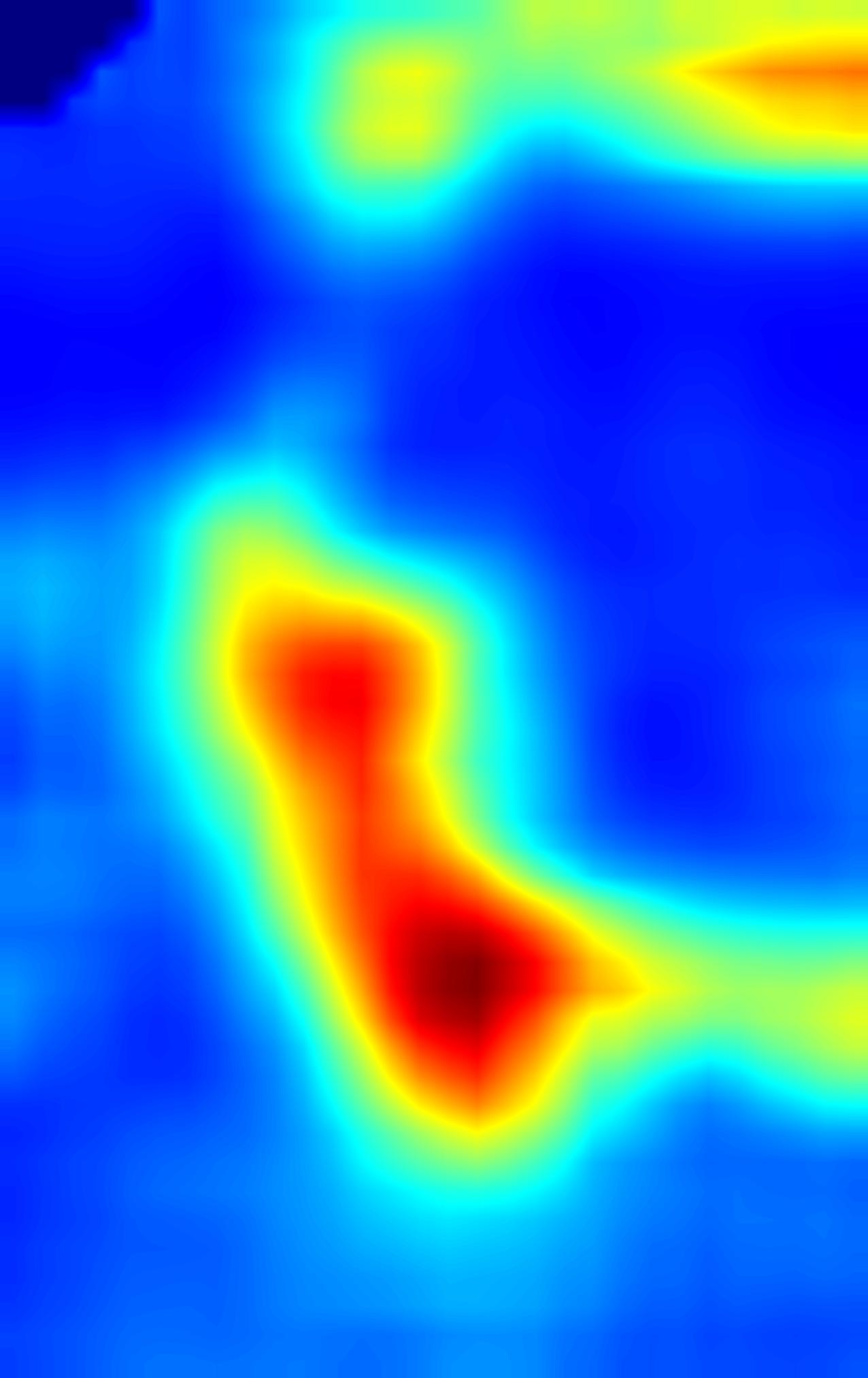}
	\includegraphics[width=0.19\linewidth,height=0.28\linewidth]{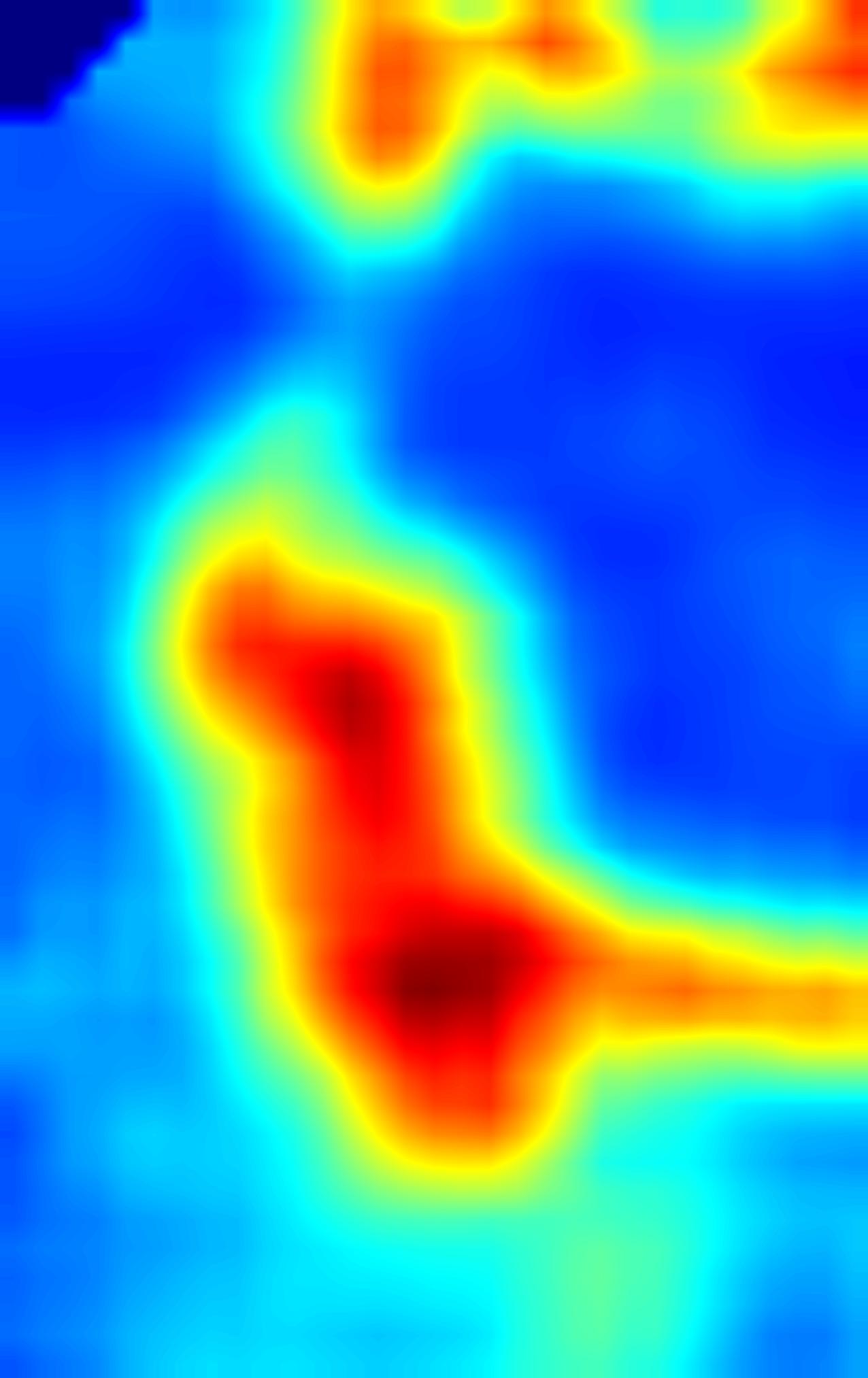}
	\includegraphics[width=0.19\linewidth,height=0.28\linewidth]{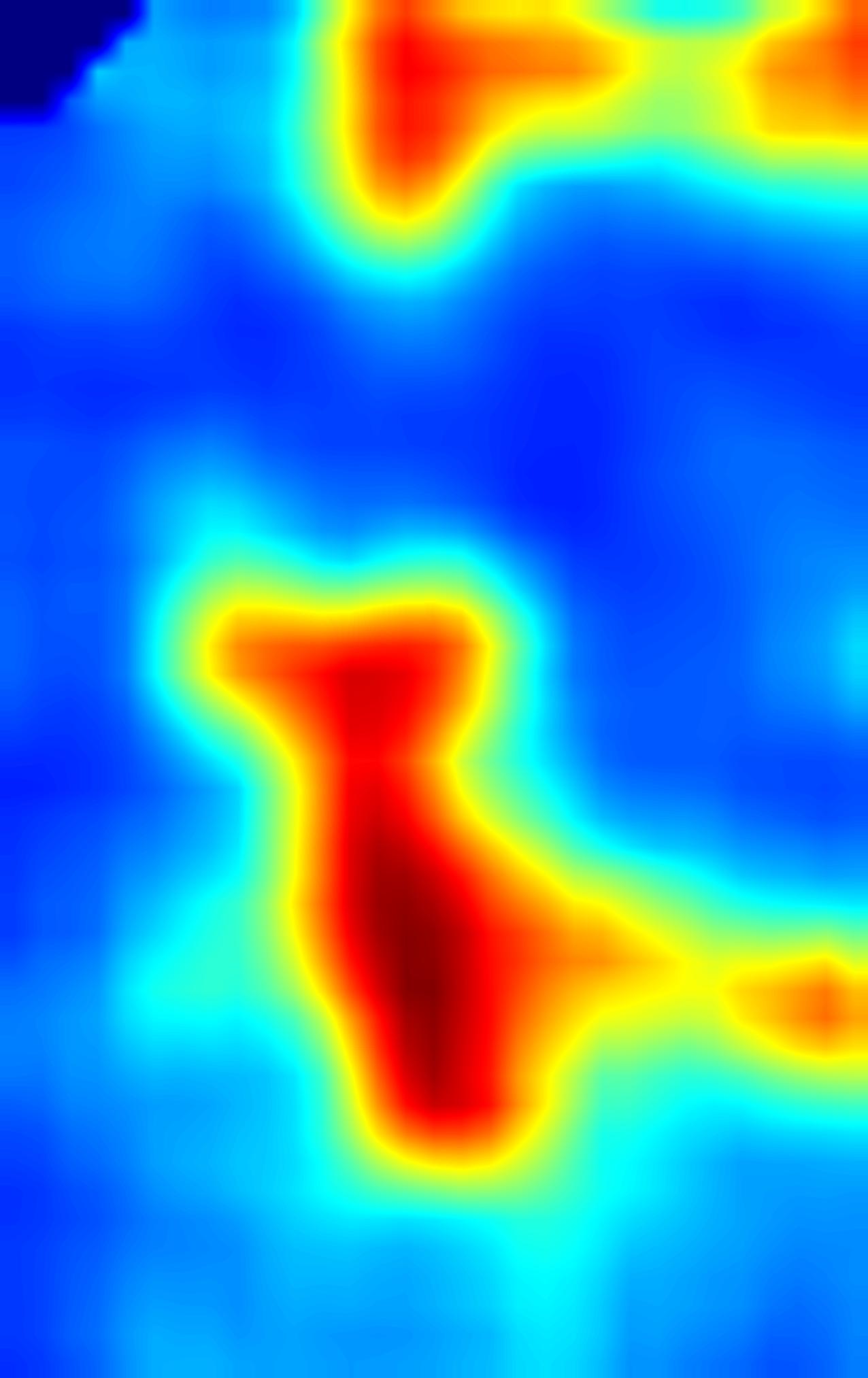}\\
	\vspace{0.1cm}
	\includegraphics[width=0.19\linewidth]{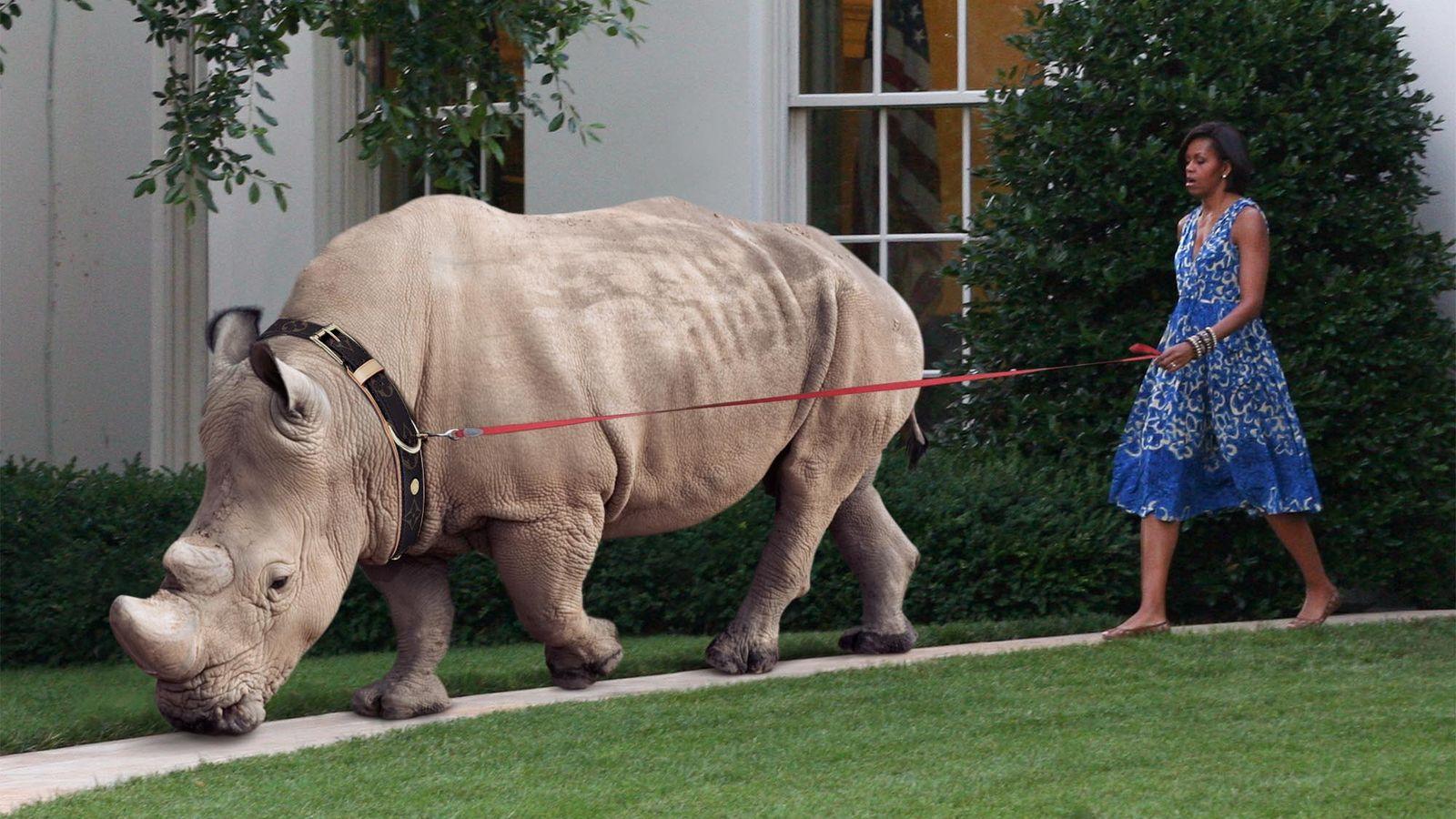}
	\includegraphics[width=0.19\linewidth]{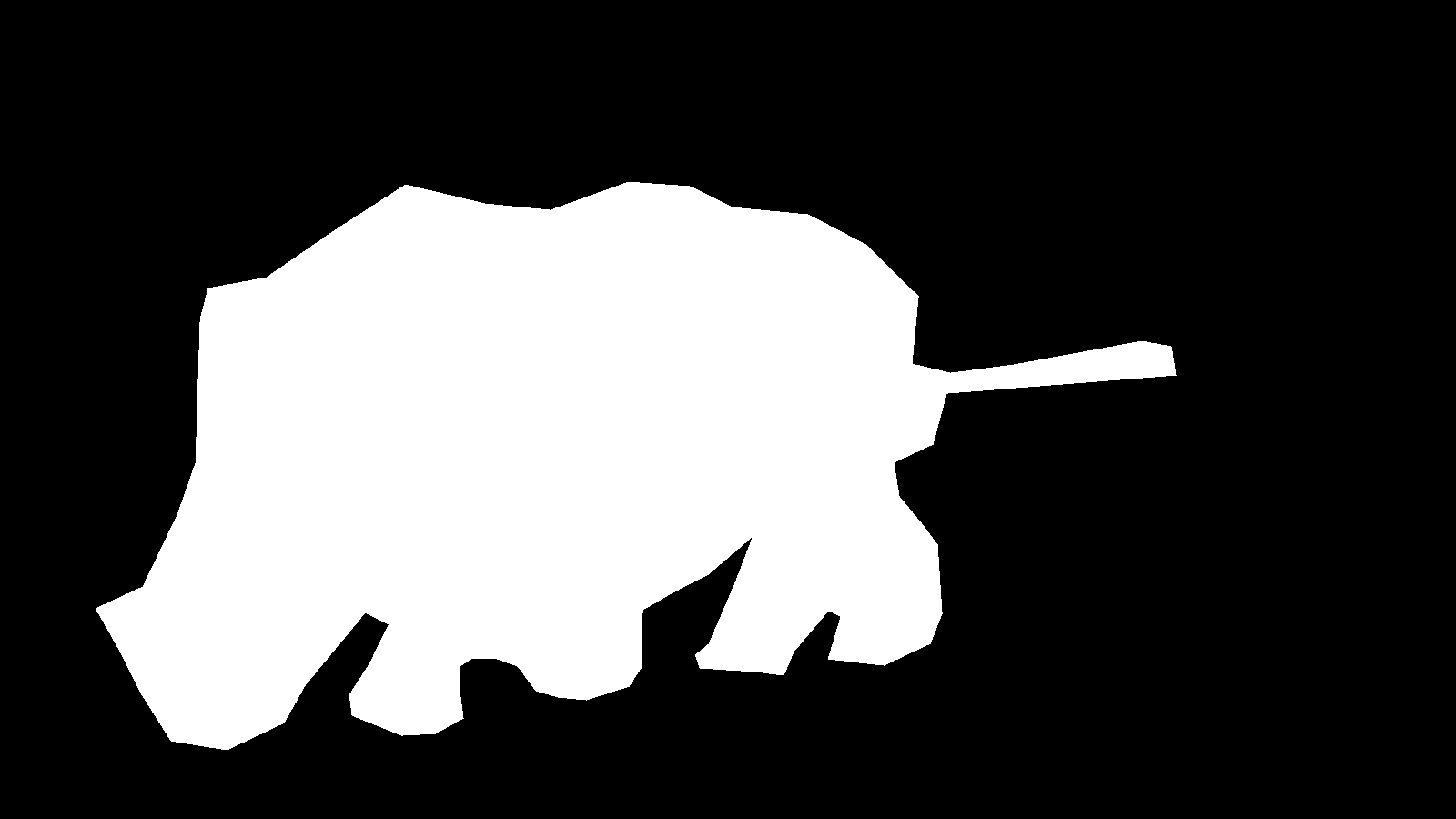}	
	\includegraphics[width=0.19\linewidth]{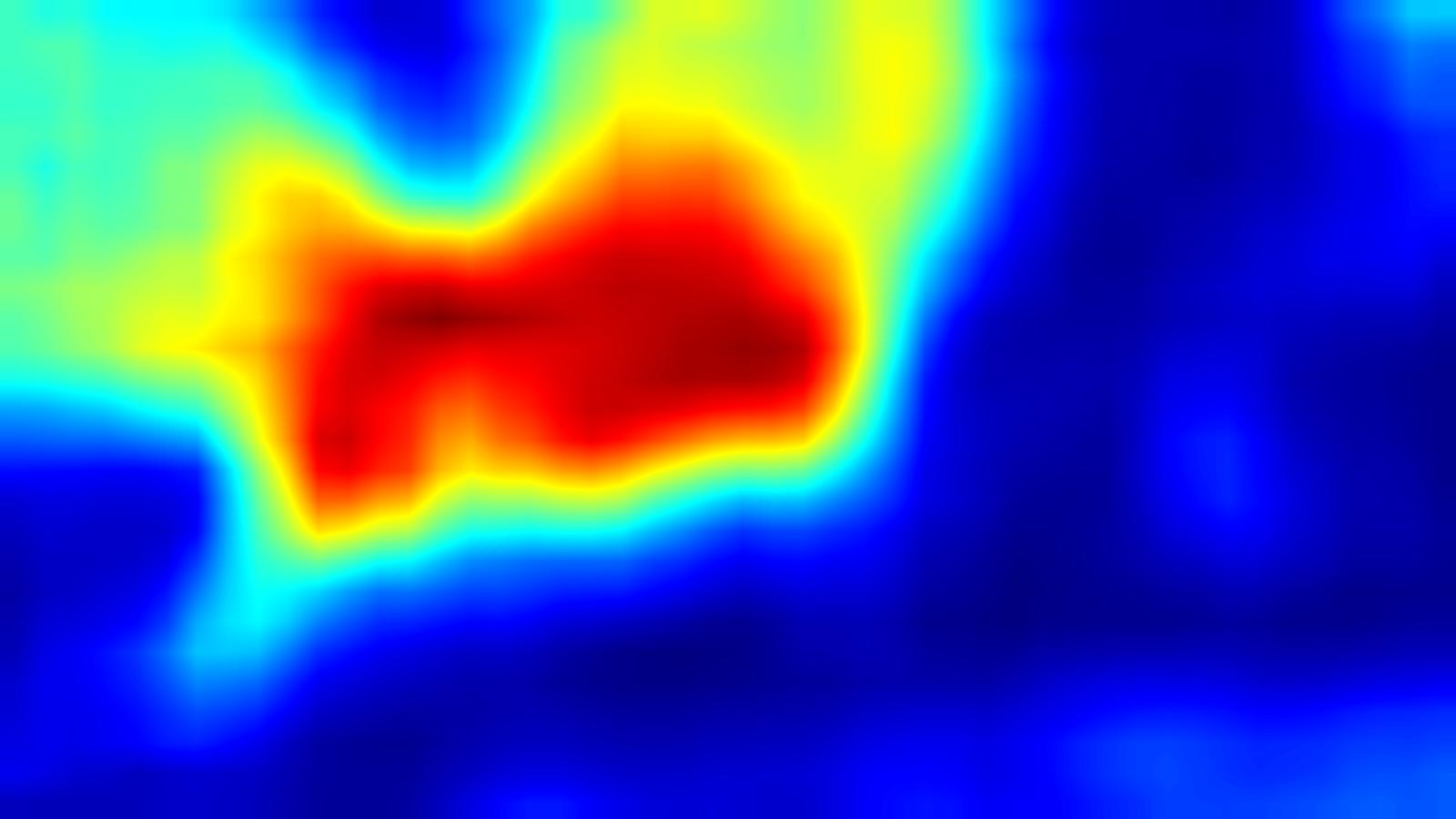}
	\includegraphics[width=0.19\linewidth]{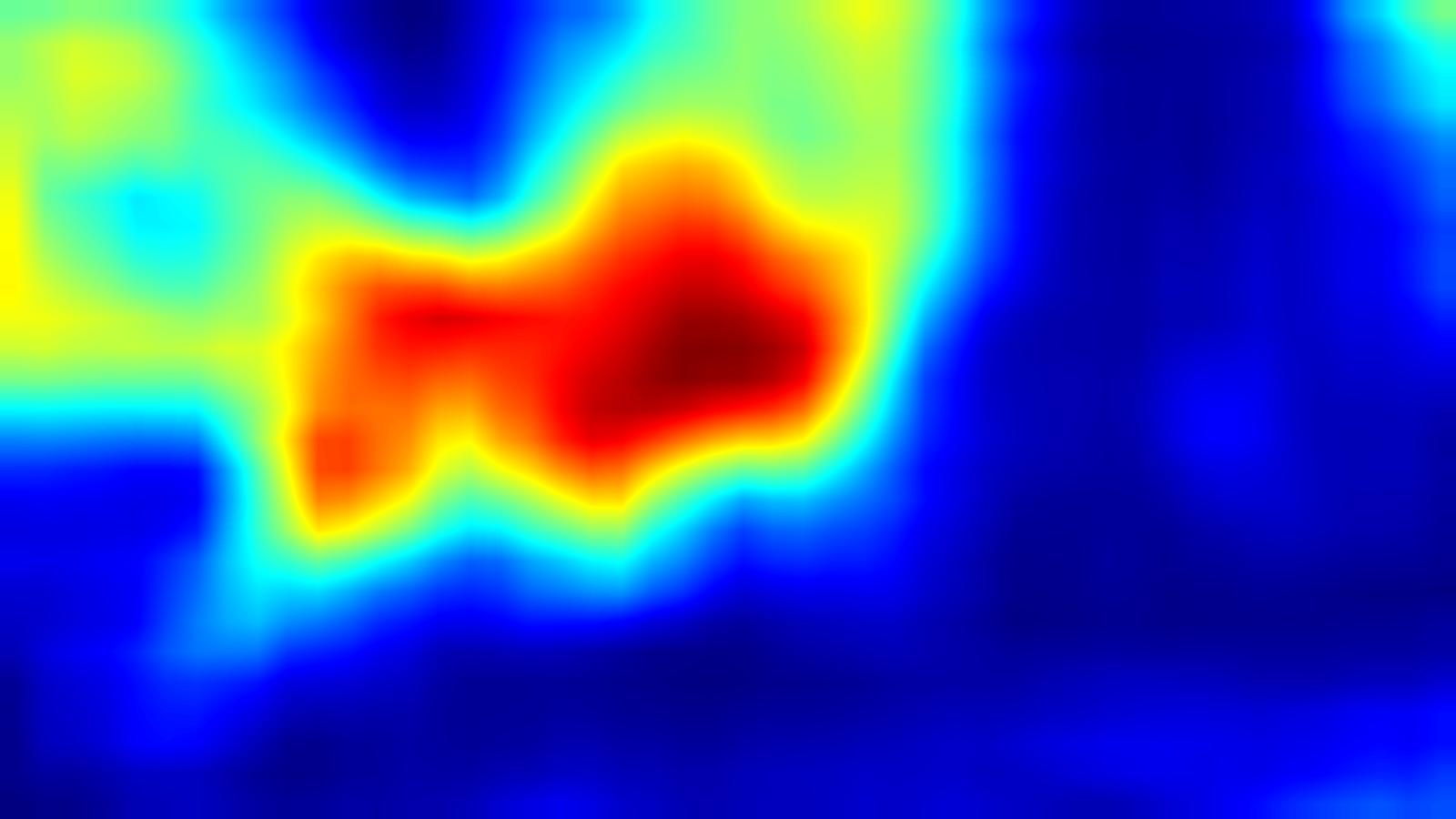}
	\includegraphics[width=0.19\linewidth]{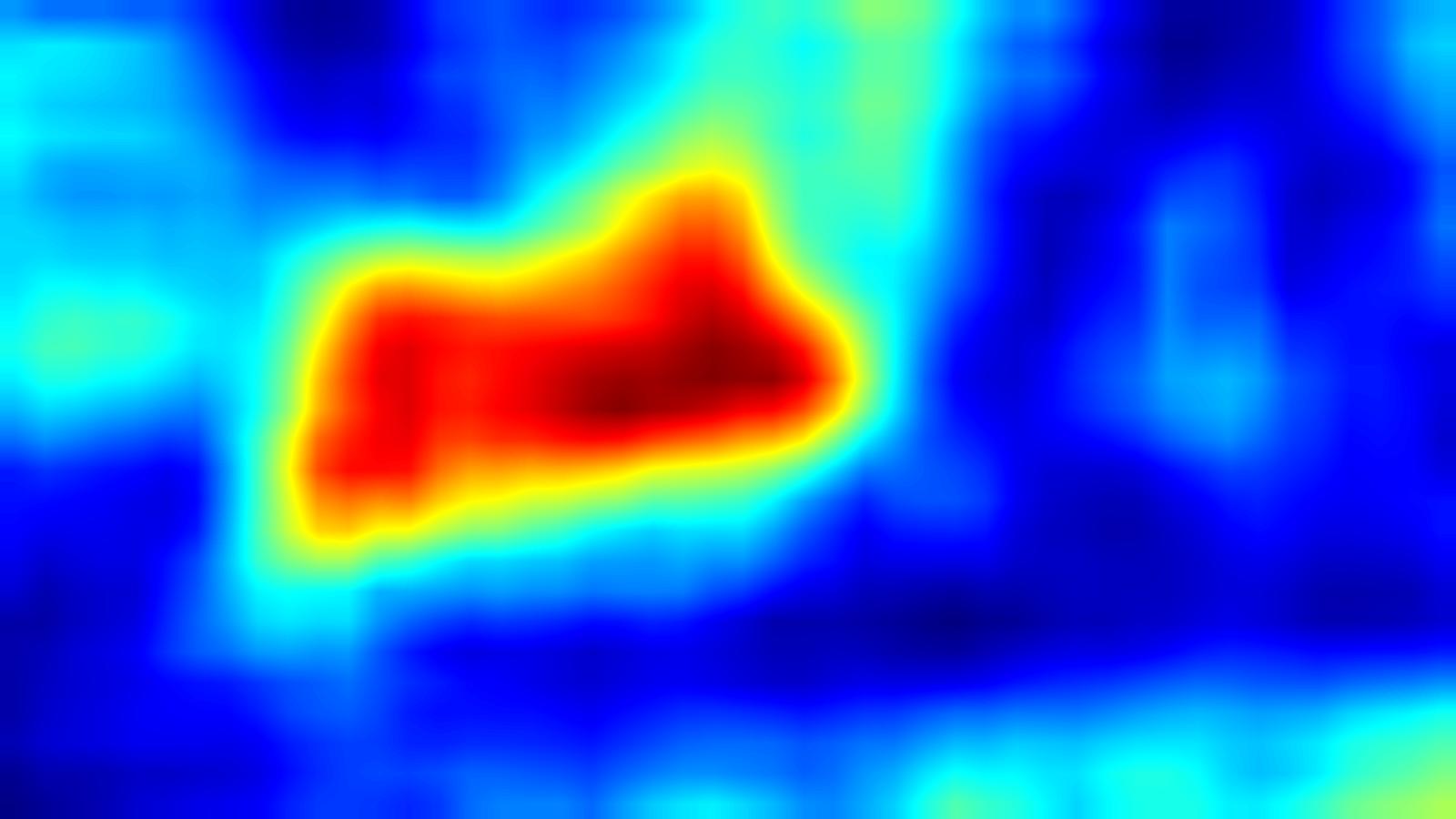}\\
	\vspace{0.1cm}
	\includegraphics[width=0.19\linewidth]{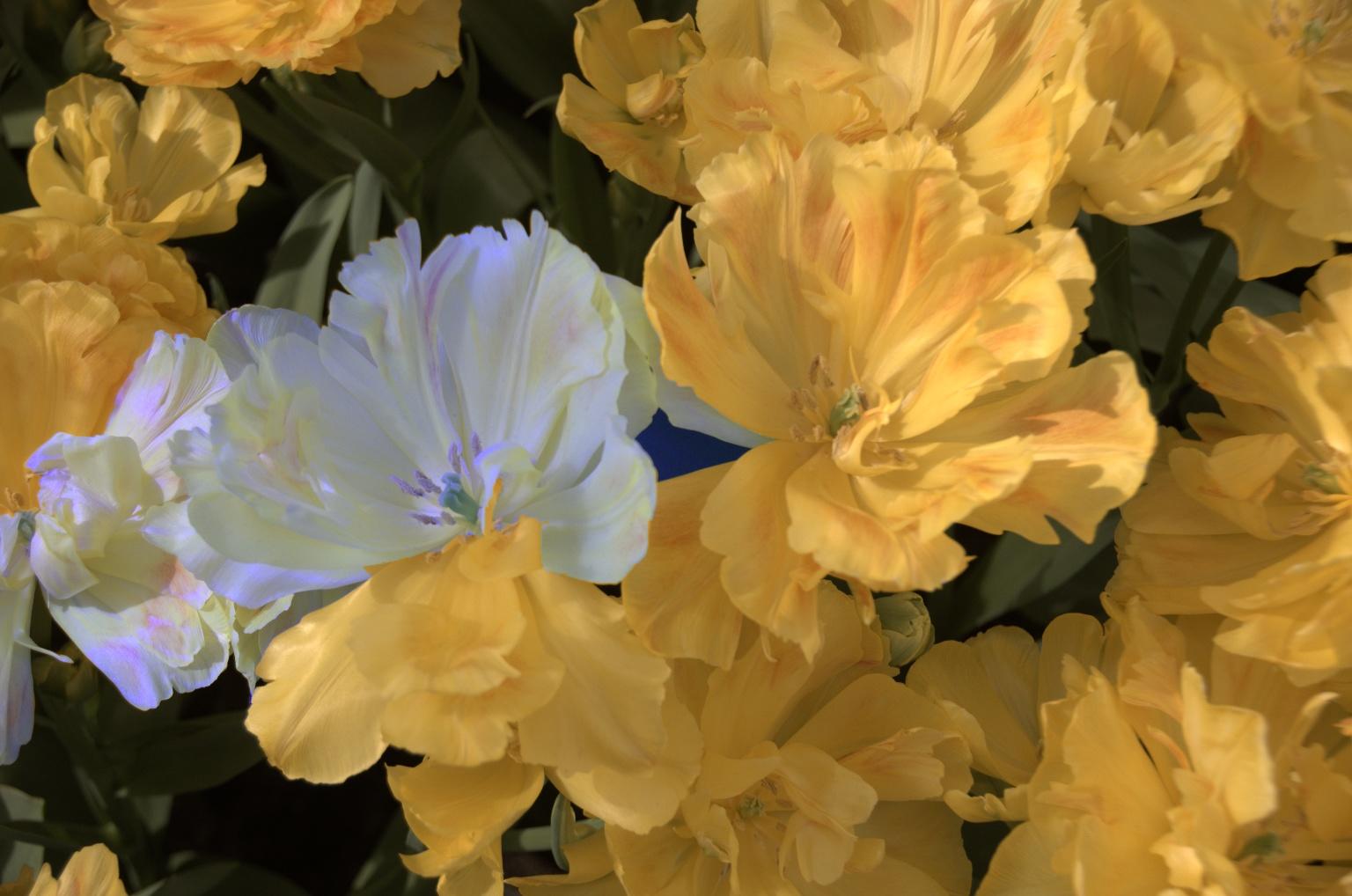}
	\includegraphics[width=0.19\linewidth]{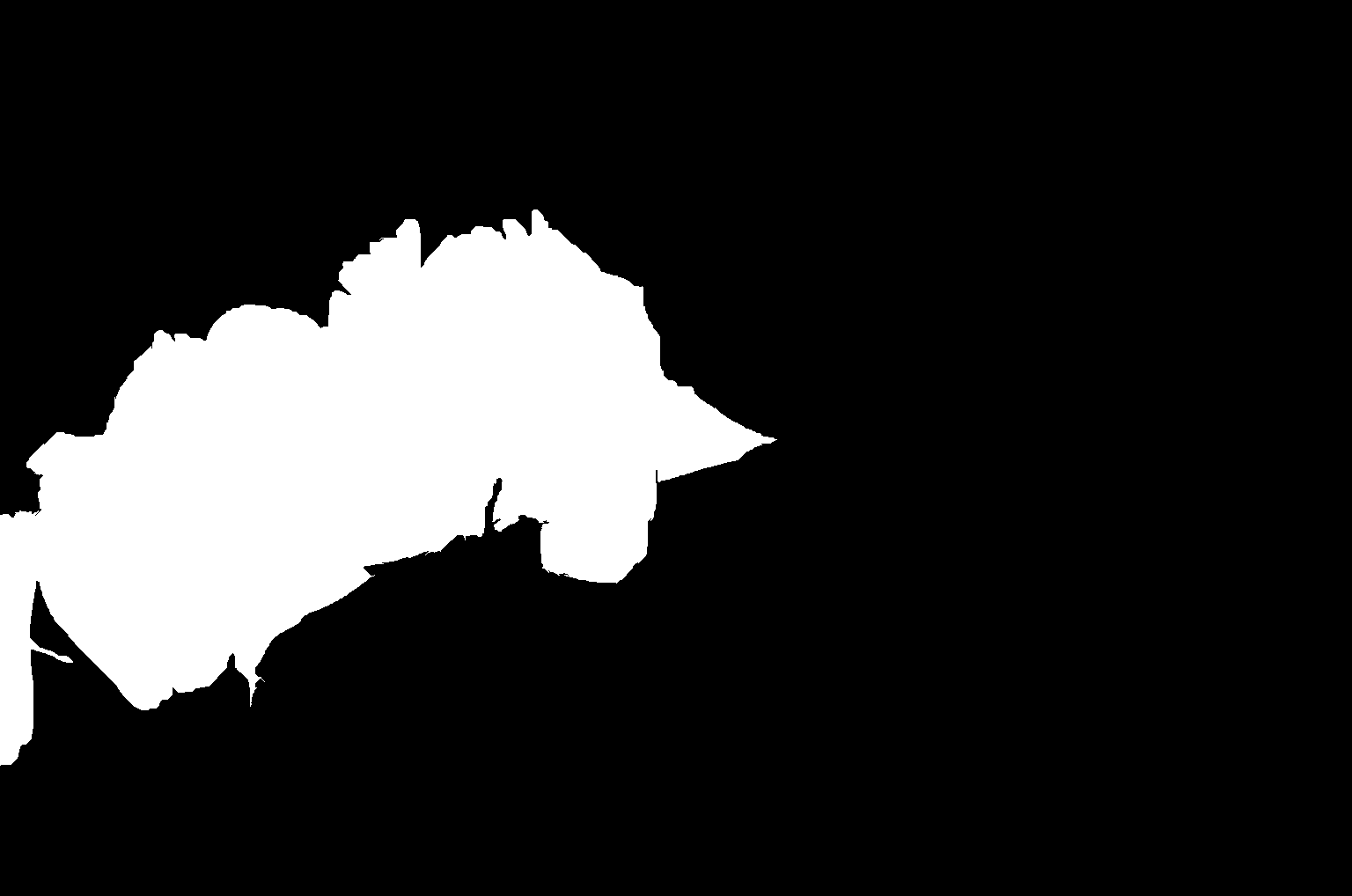}
	\includegraphics[width=0.19\linewidth]{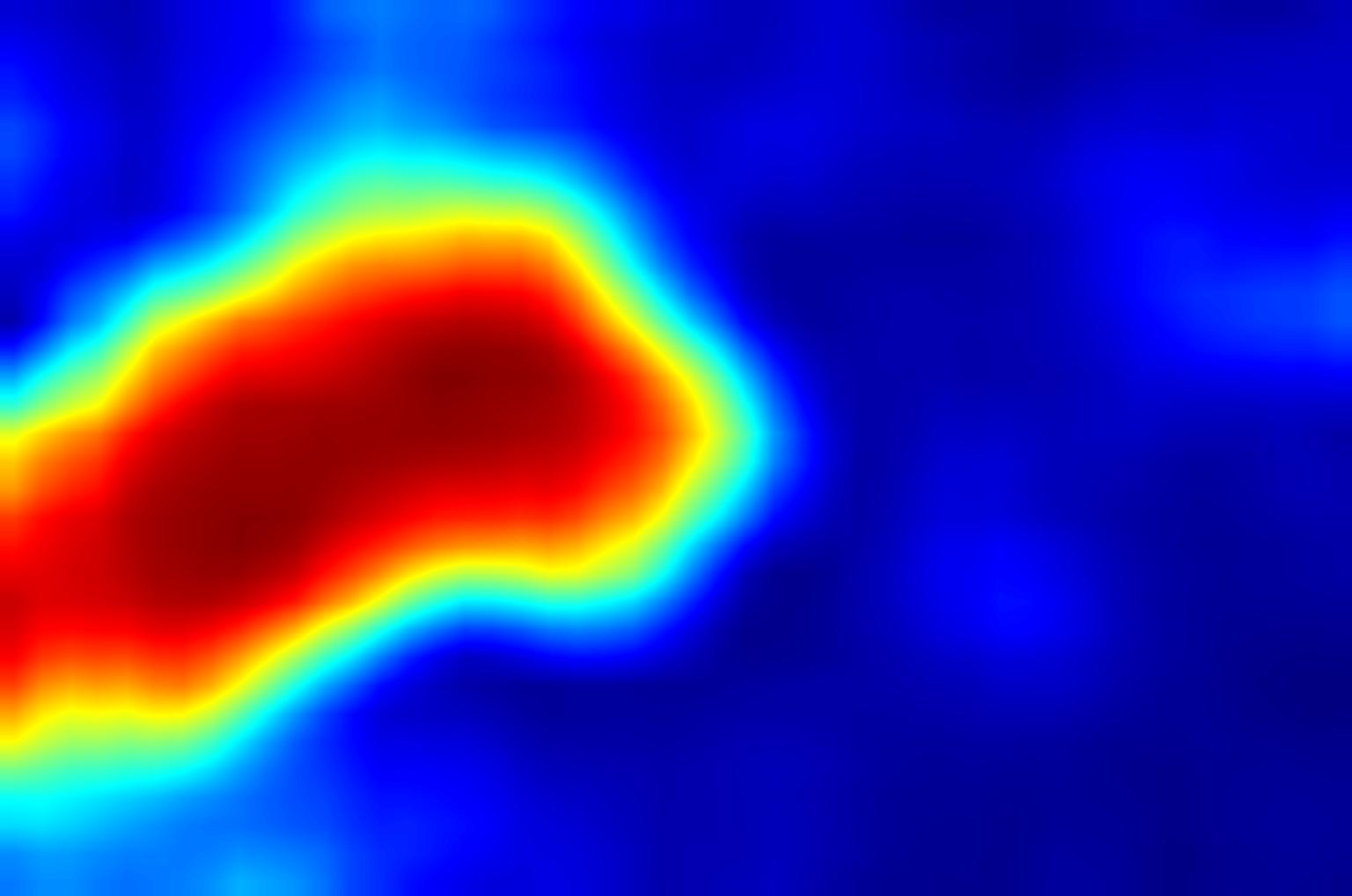}
	\includegraphics[width=0.19\linewidth]{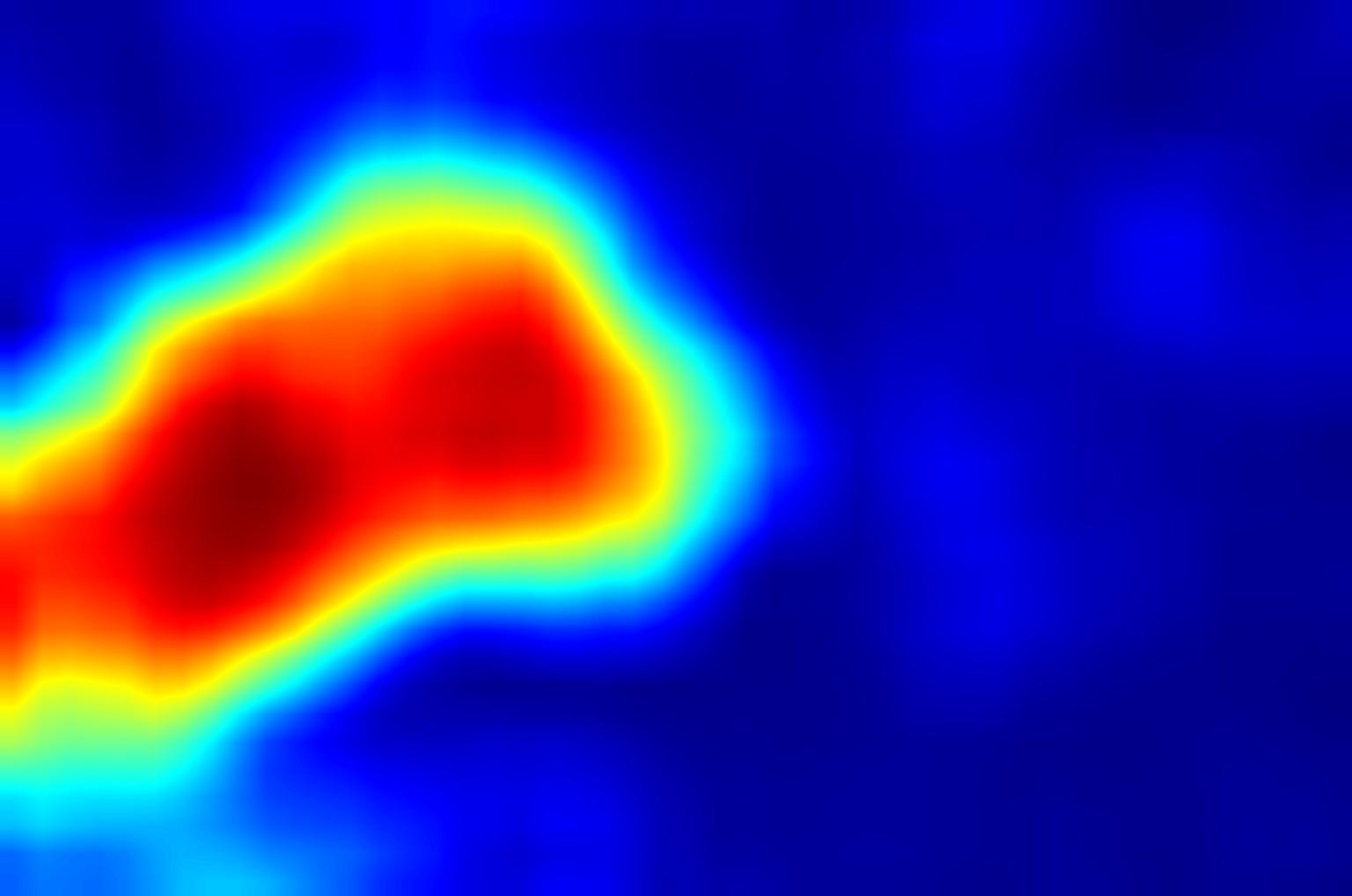}
	\includegraphics[width=0.19\linewidth]{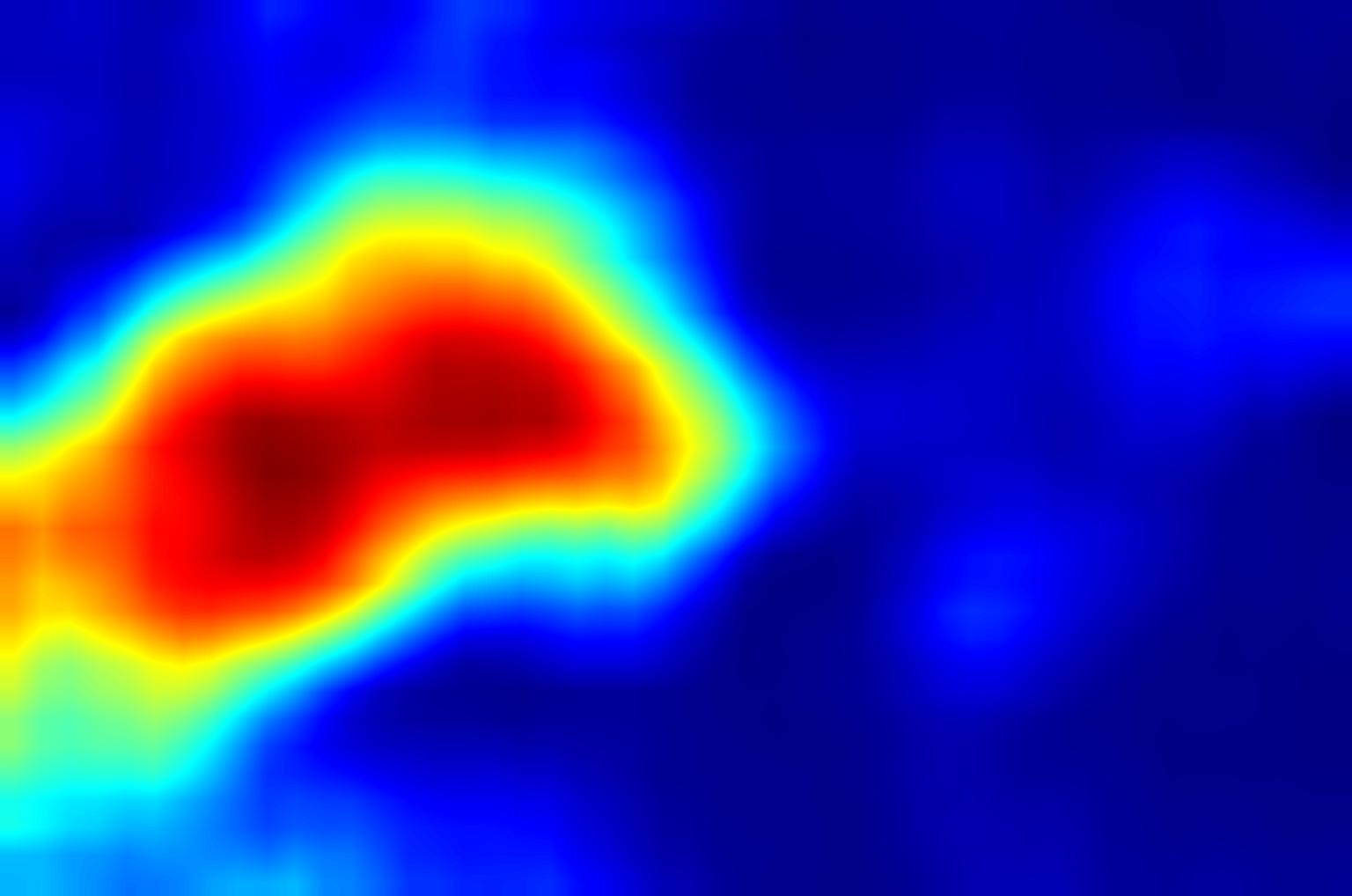}\\
	\vspace{0.1cm}
	\includegraphics[width=0.19\linewidth]{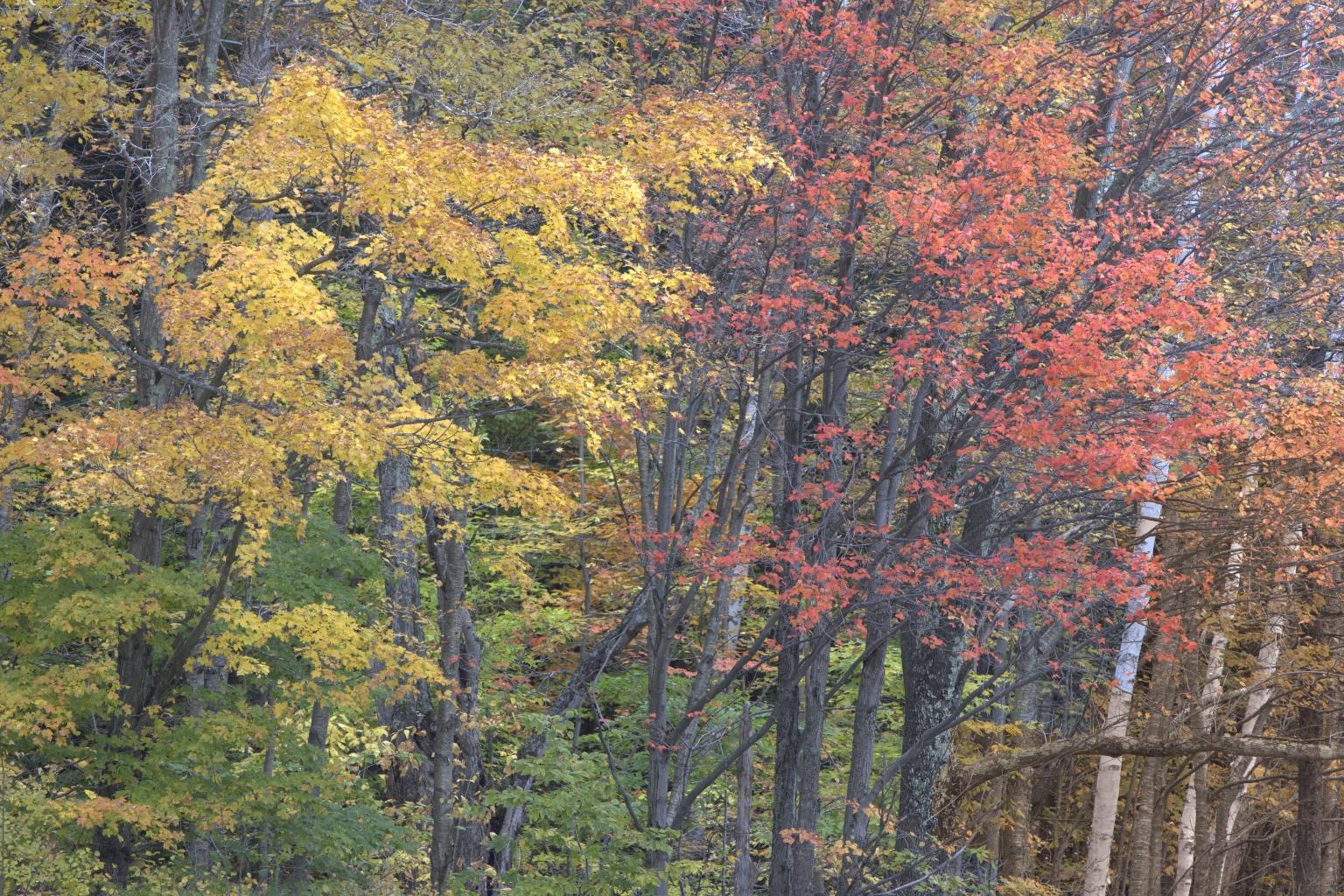}
	\includegraphics[width=0.19\linewidth]{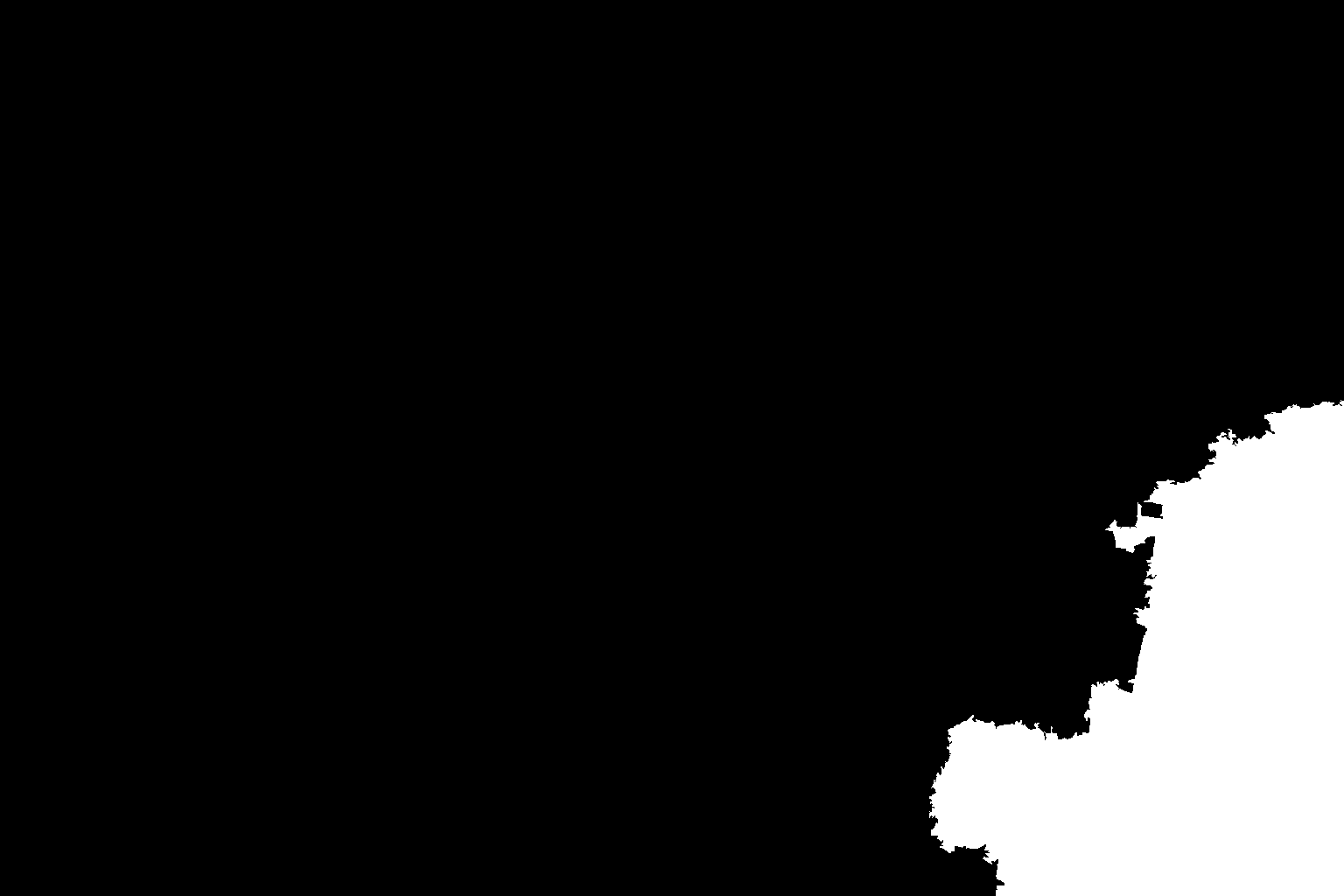}
	\includegraphics[width=0.19\linewidth]{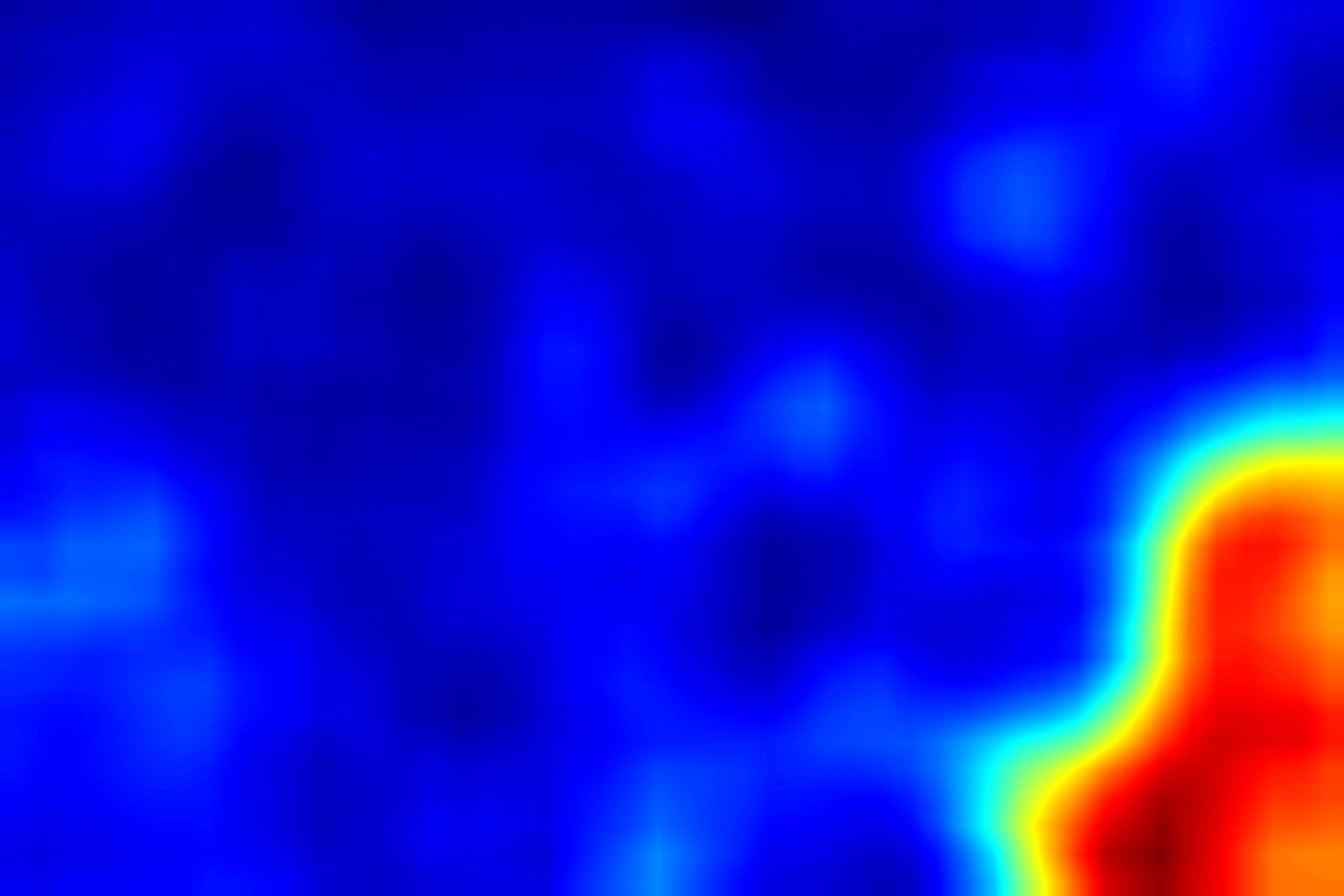}
	\includegraphics[width=0.19\linewidth]{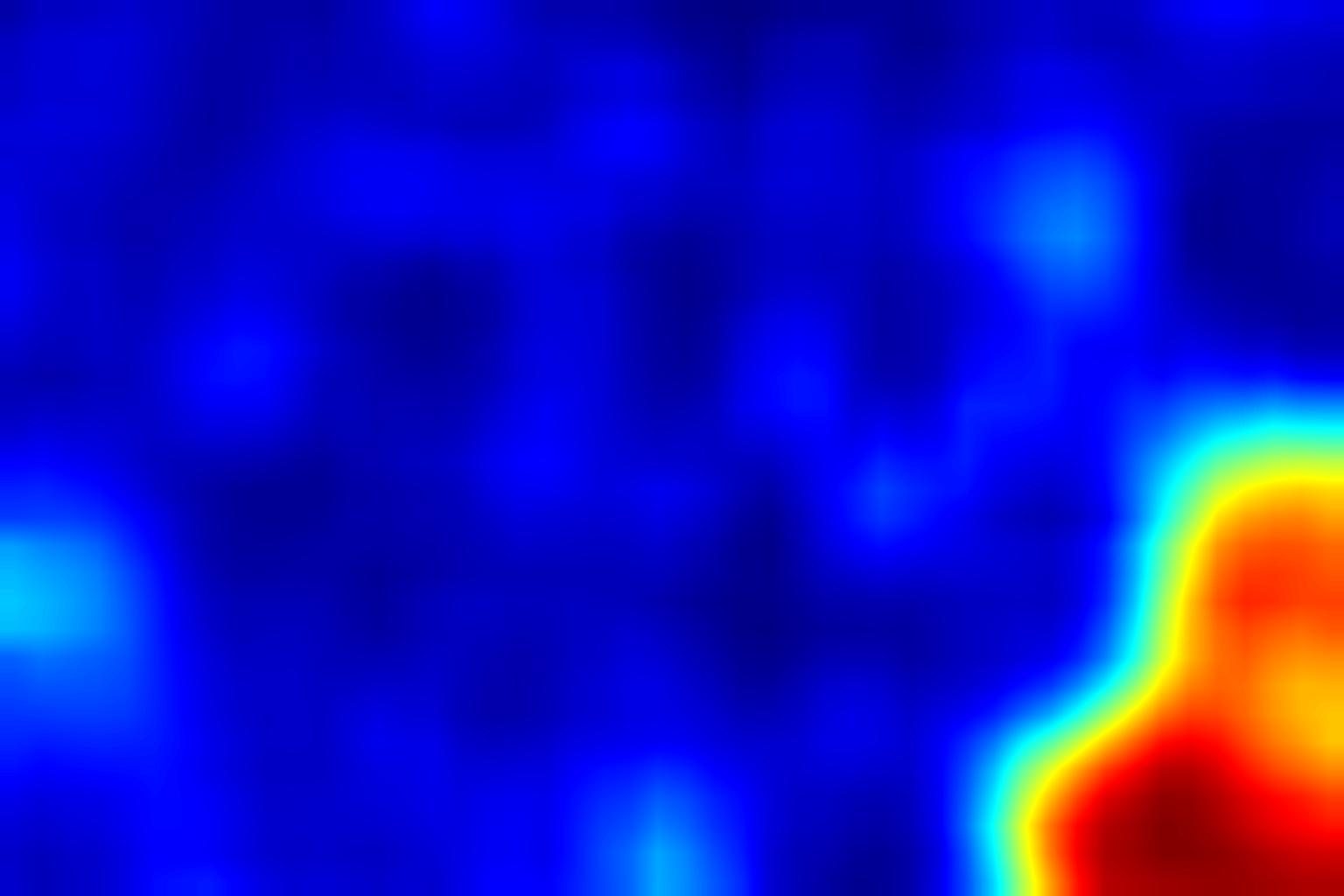}
	\includegraphics[width=0.19\linewidth]{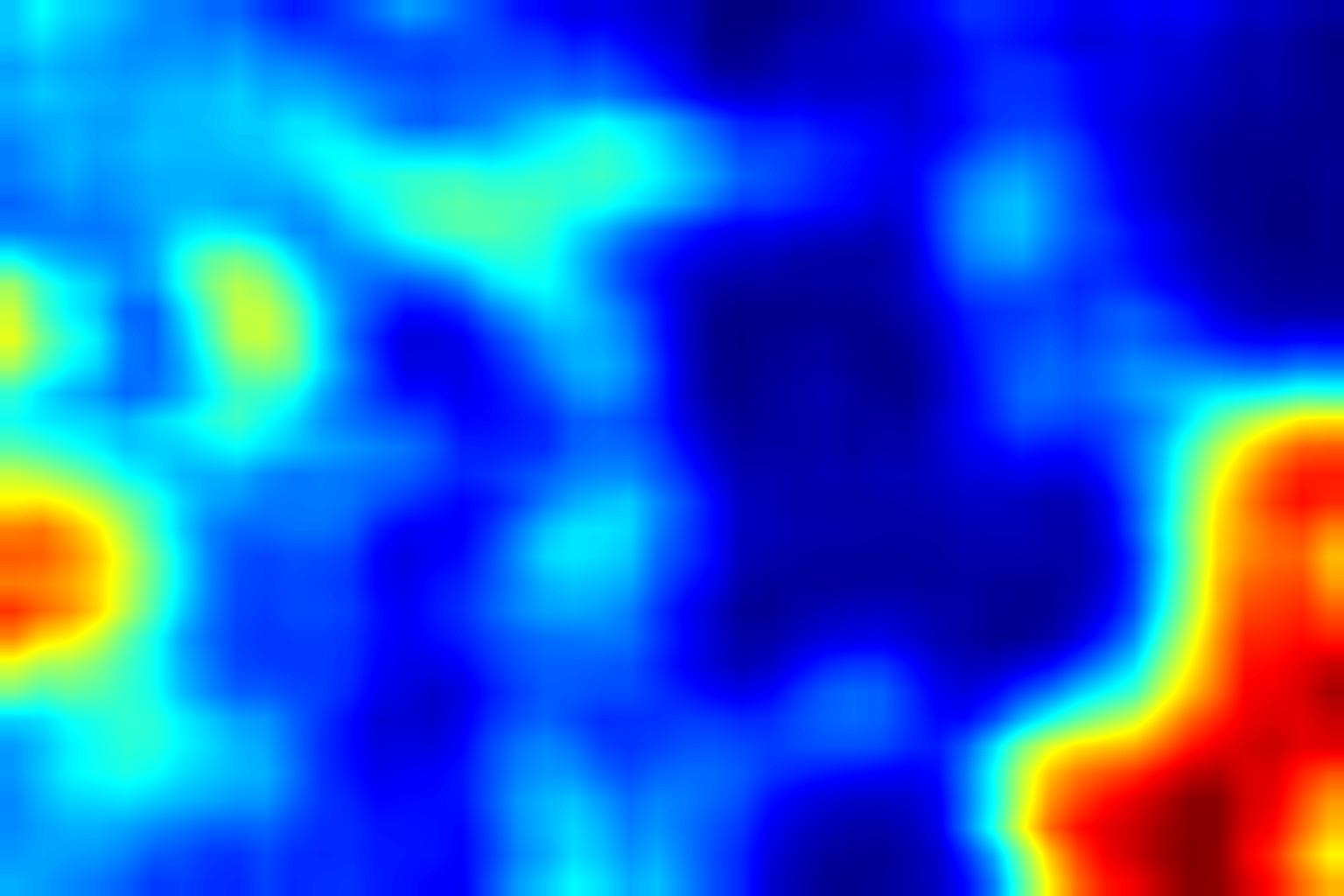}\\
	\vspace{0.1cm}
%	\includegraphics[width=0.19\linewidth]{figures/canonxt_kodakdcs330_sub_02_resorig_qualunc.jpg}
%	\includegraphics[width=0.19\linewidth]{figures/canonxt_kodakdcs330_sub_02_mask_inv.png}
%	\includegraphics[width=0.19\linewidth]{figures/canonxt_kodakdcs330_sub_02_resorig_qualunc_hm_ms5.jpg}
%	\includegraphics[width=0.19\linewidth]{figures/canonxt_kodakdcs330_sub_02_res1200_qual75_hm_ms5.jpg}
%	\includegraphics[width=0.19\linewidth]{figures/canonxt_kodakdcs330_sub_02_res800_qual50_hm_ms5.jpg}\\
%	\vspace{0.1cm}
	\includegraphics[width=0.19\linewidth]{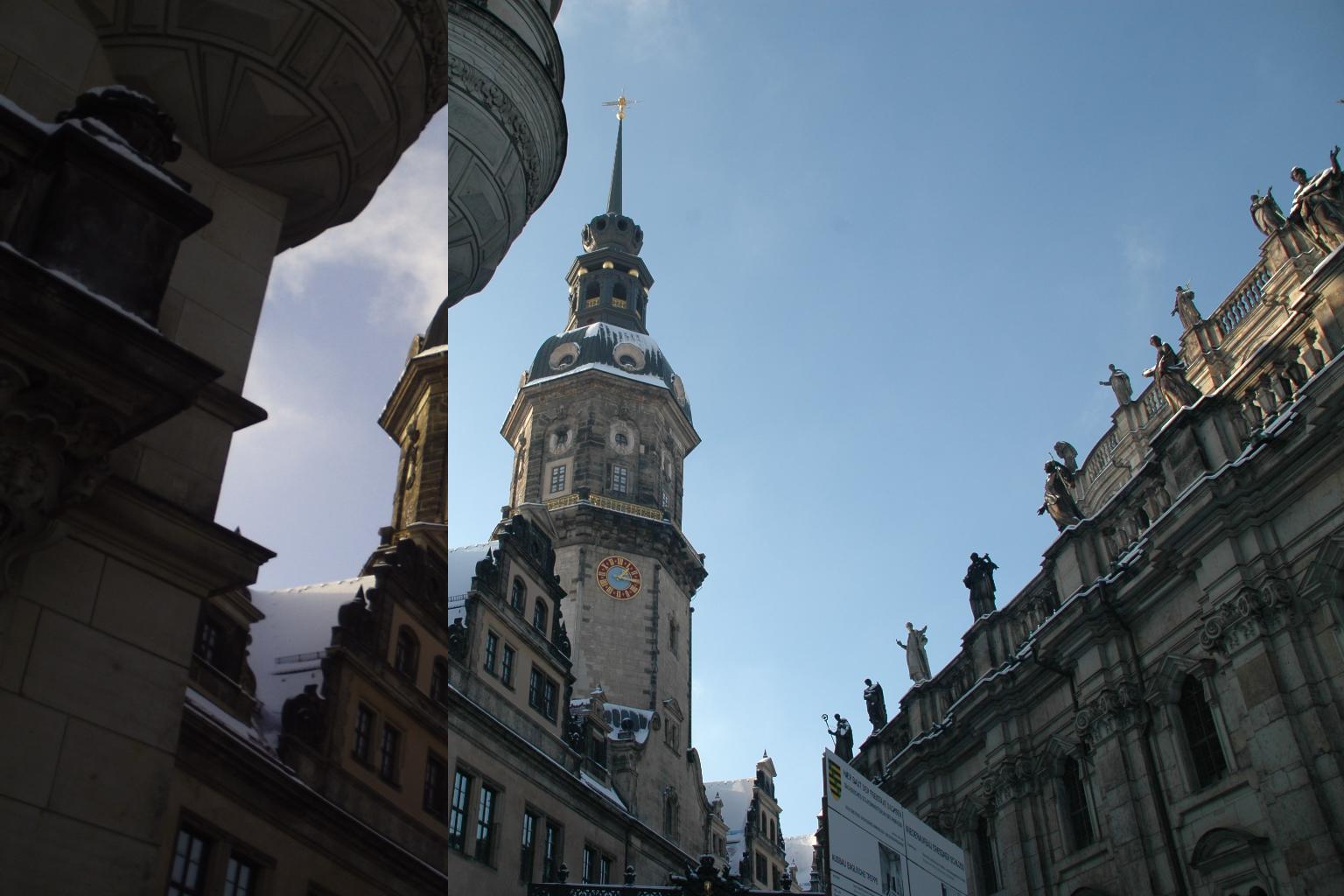}
	\includegraphics[width=0.19\linewidth]{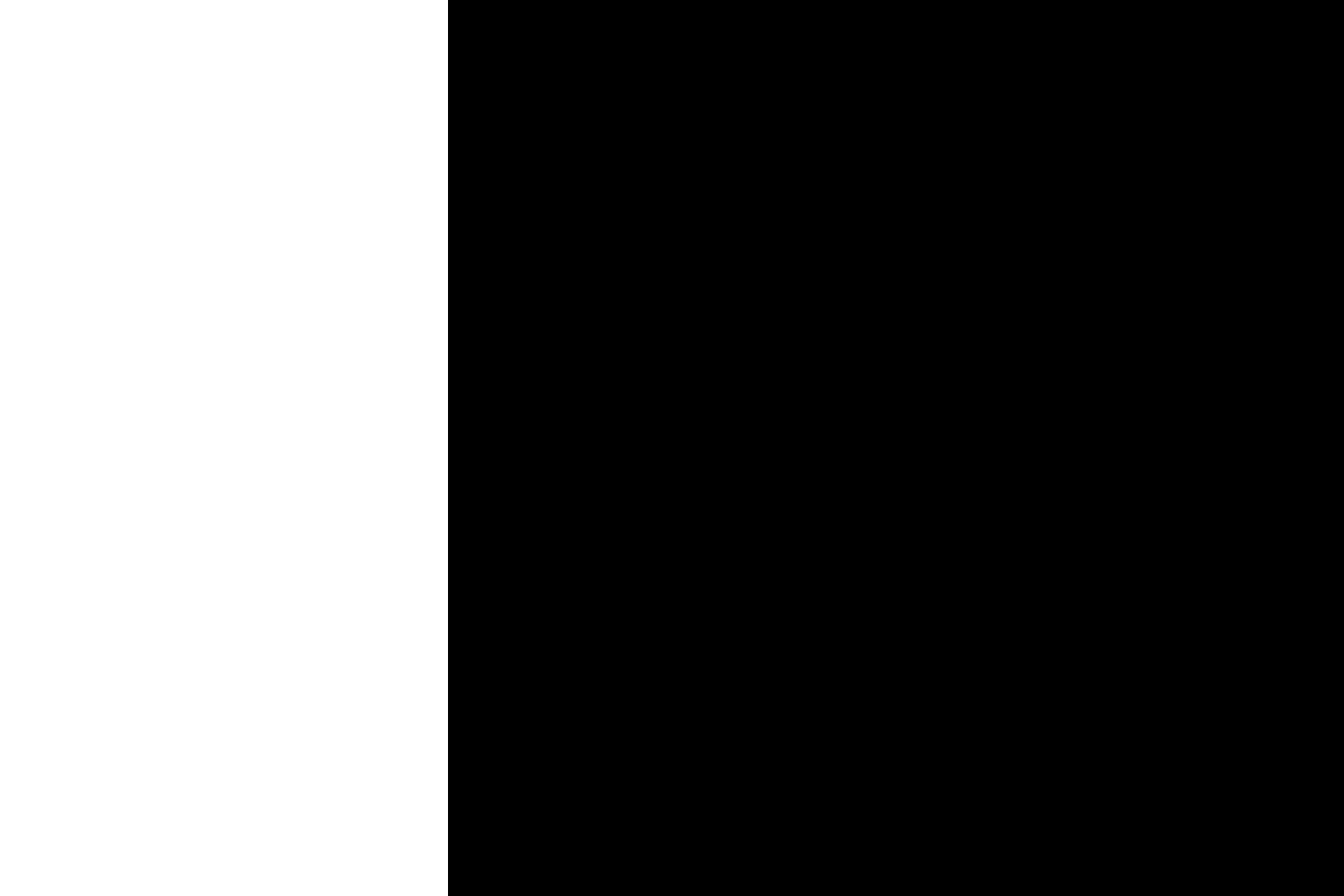}
	\includegraphics[width=0.19\linewidth]{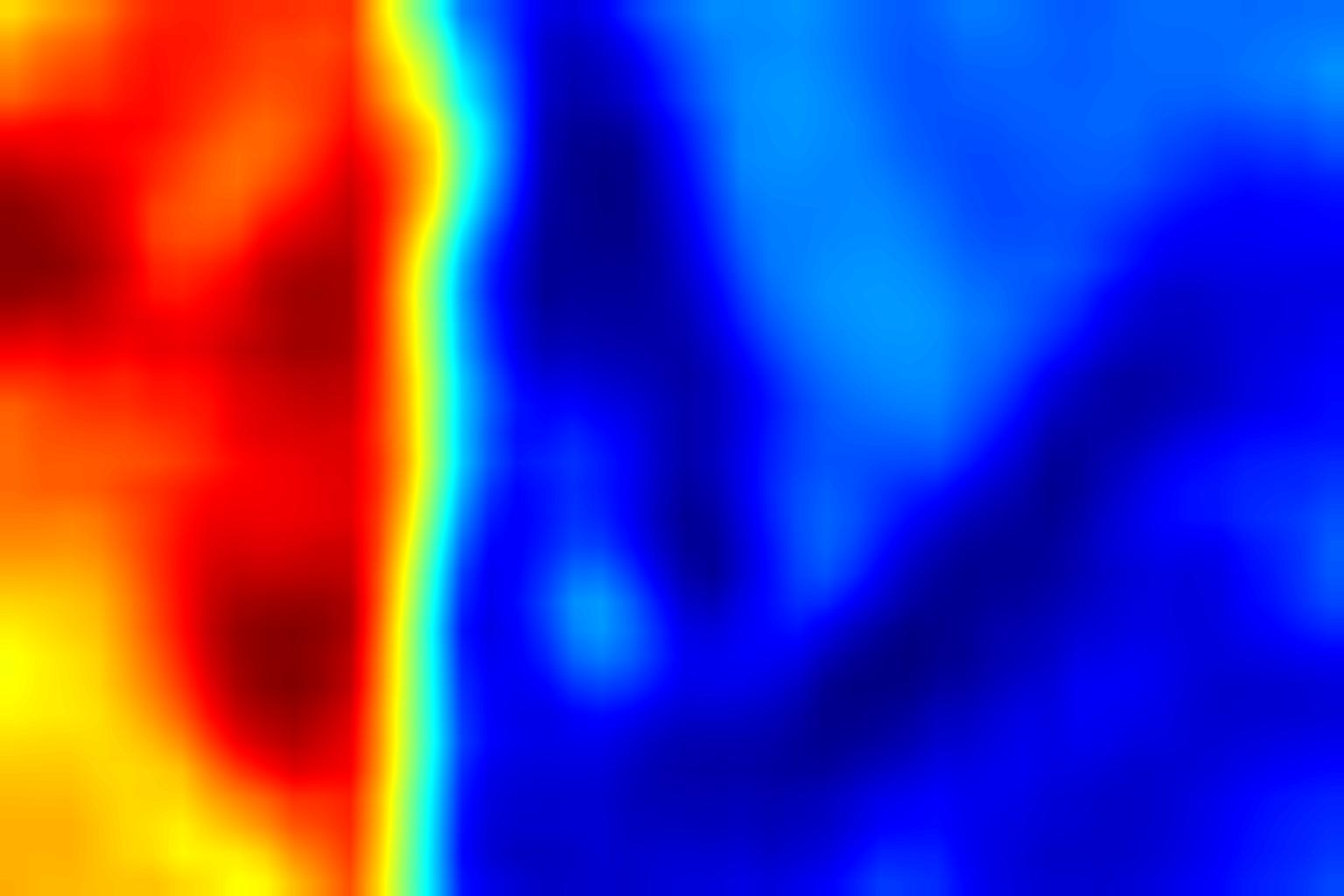}
	\includegraphics[width=0.19\linewidth]{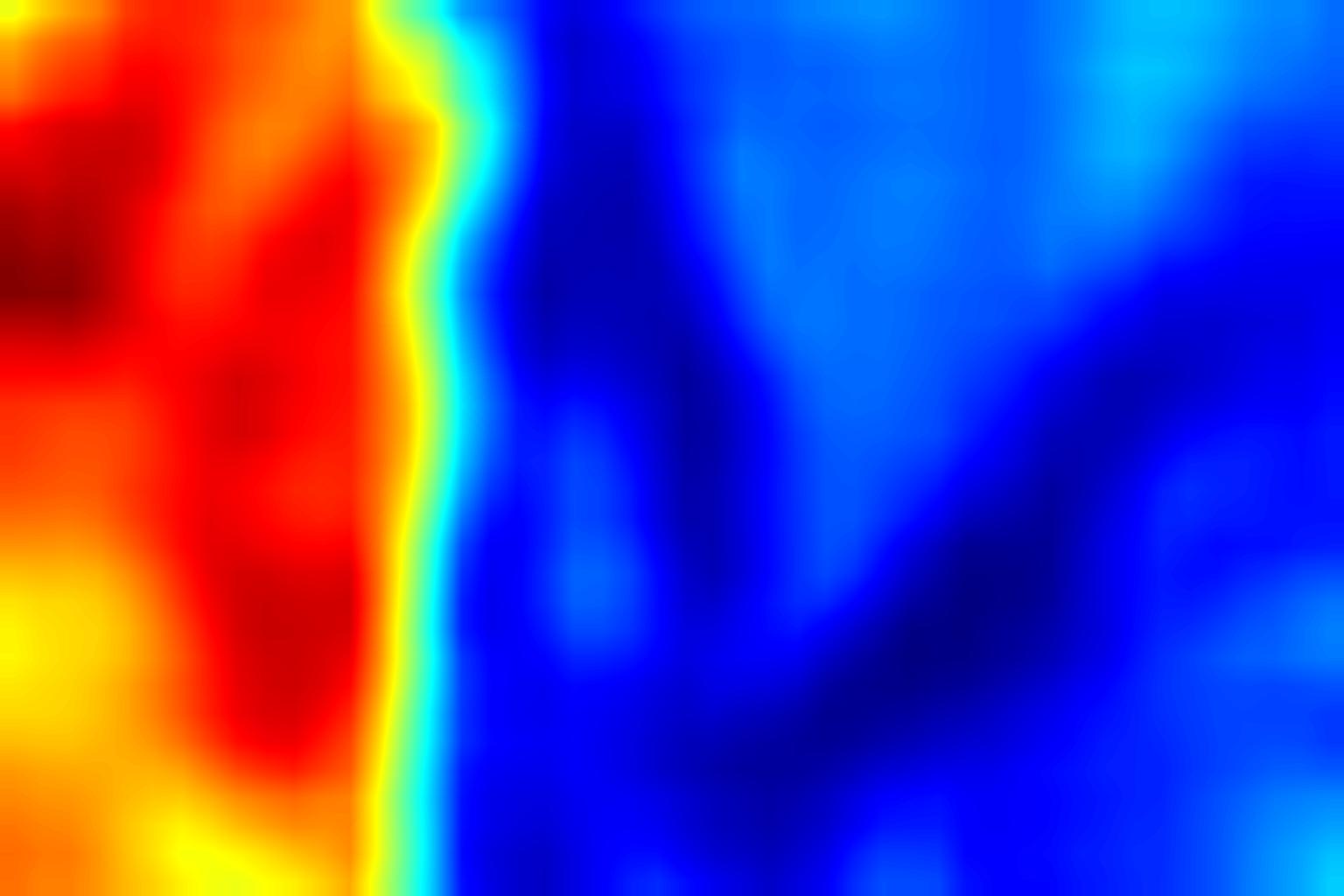}
	\includegraphics[width=0.19\linewidth]{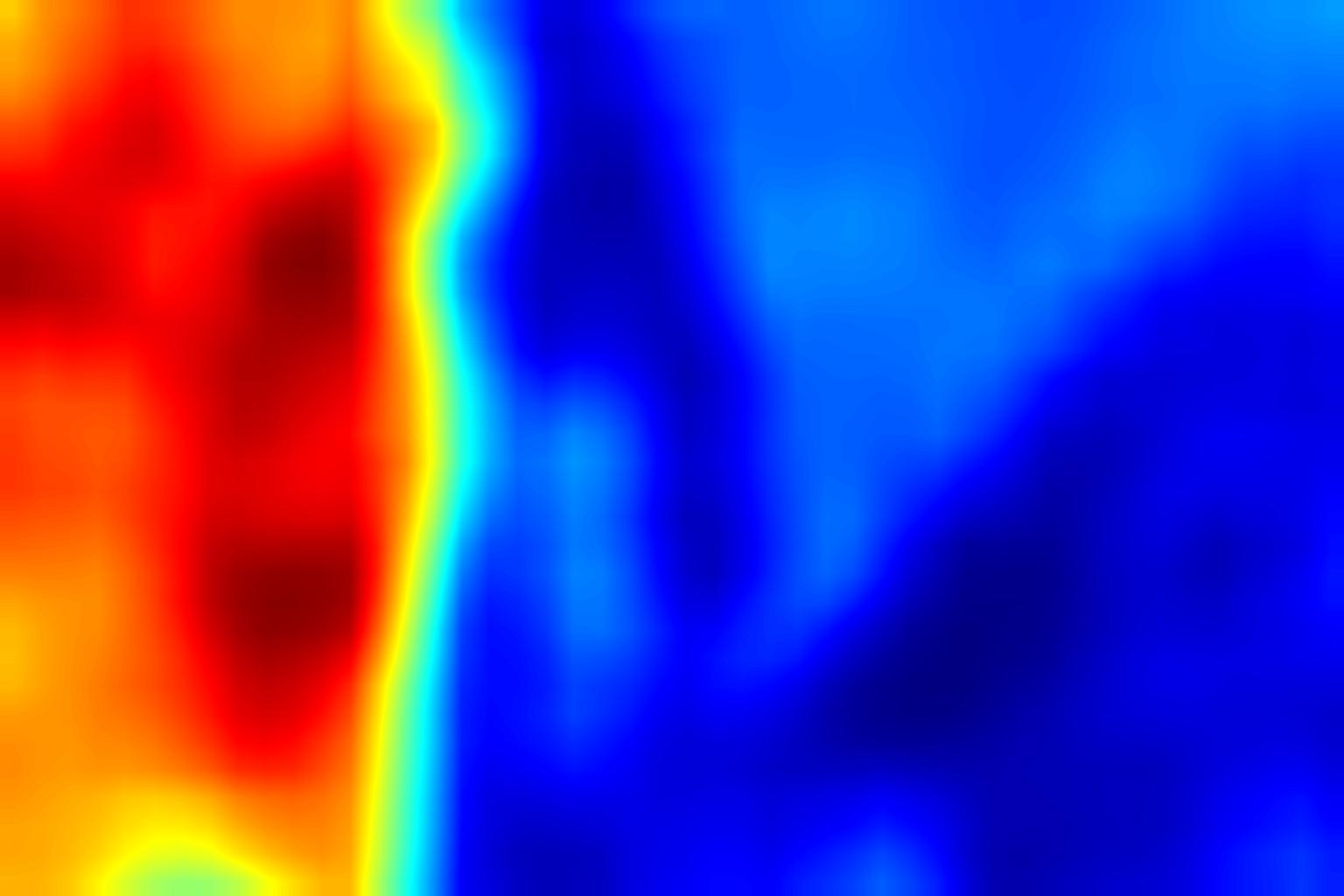}\\
	
	\caption{Qualitative results for manipulation localization using MeanShift. Results barely deteriorate from high-quality (HQ), medium quality (MQ) to low-quality (LQ) images.}
	% First column: test images. Second column: ground truth. Third column: Heatmaps for the original test image. Fourth resp. fifth column: (Upscaled) heatmaps for test images resized to $1200$, resp. $800$ pixels in larger dimension and JPEG compressed with quality level $75$, resp. $50$ (except for last image: no resizing, JPEG qualities $70$, resp. $50$). First three images from In-The-Wild, fourth and fifth image from \newDatasetName{}, sixth image from Columbia, seventh image from Aligned Scenes.
	\label{fig:loc_qual_examples}
\end{figure}

\begin{figure}[tbh]
%	\vspace{-0.5cm}
	\centering
	\begin{tabular}{>{\centering}p{0.16\linewidth}@{}>{\centering}p{0.21\linewidth}@{}>{\centering}p{0.20\linewidth}@{}>{\centering}p{0.22\linewidth}@{}>{\centering\arraybackslash}p{0.14\linewidth}}
		& &  \multicolumn{3}{c}{\small{Results on}} \\\cline{3-5}
		\small{Image\phantom{m}} & \small{Ground truth} & \small{HQ} & \small{MQ} & \small{LQ} \\
	\end{tabular}
	\includegraphics[width=0.19\linewidth]{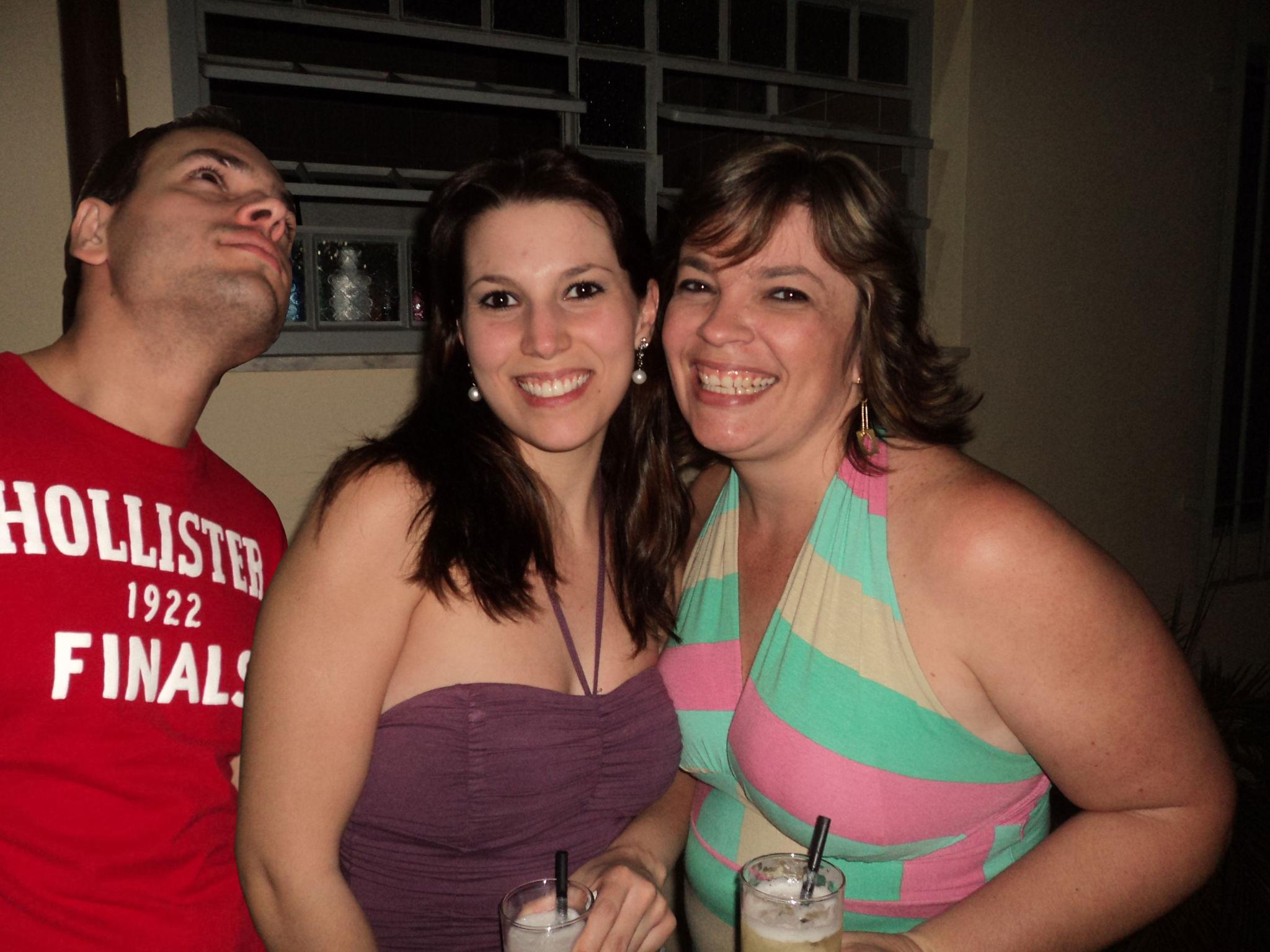}
	\includegraphics[width=0.19\linewidth]{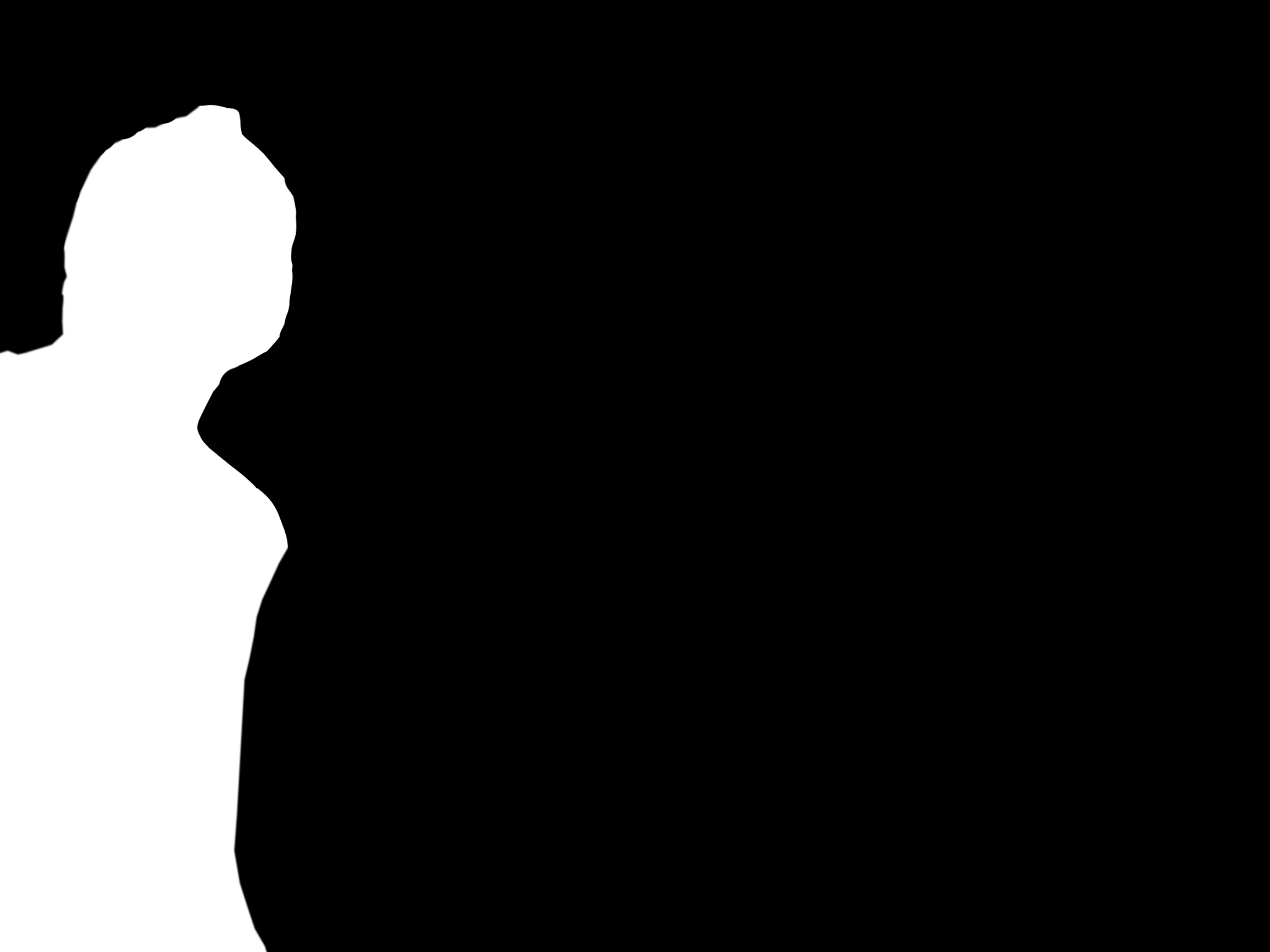}
	\includegraphics[width=0.19\linewidth]{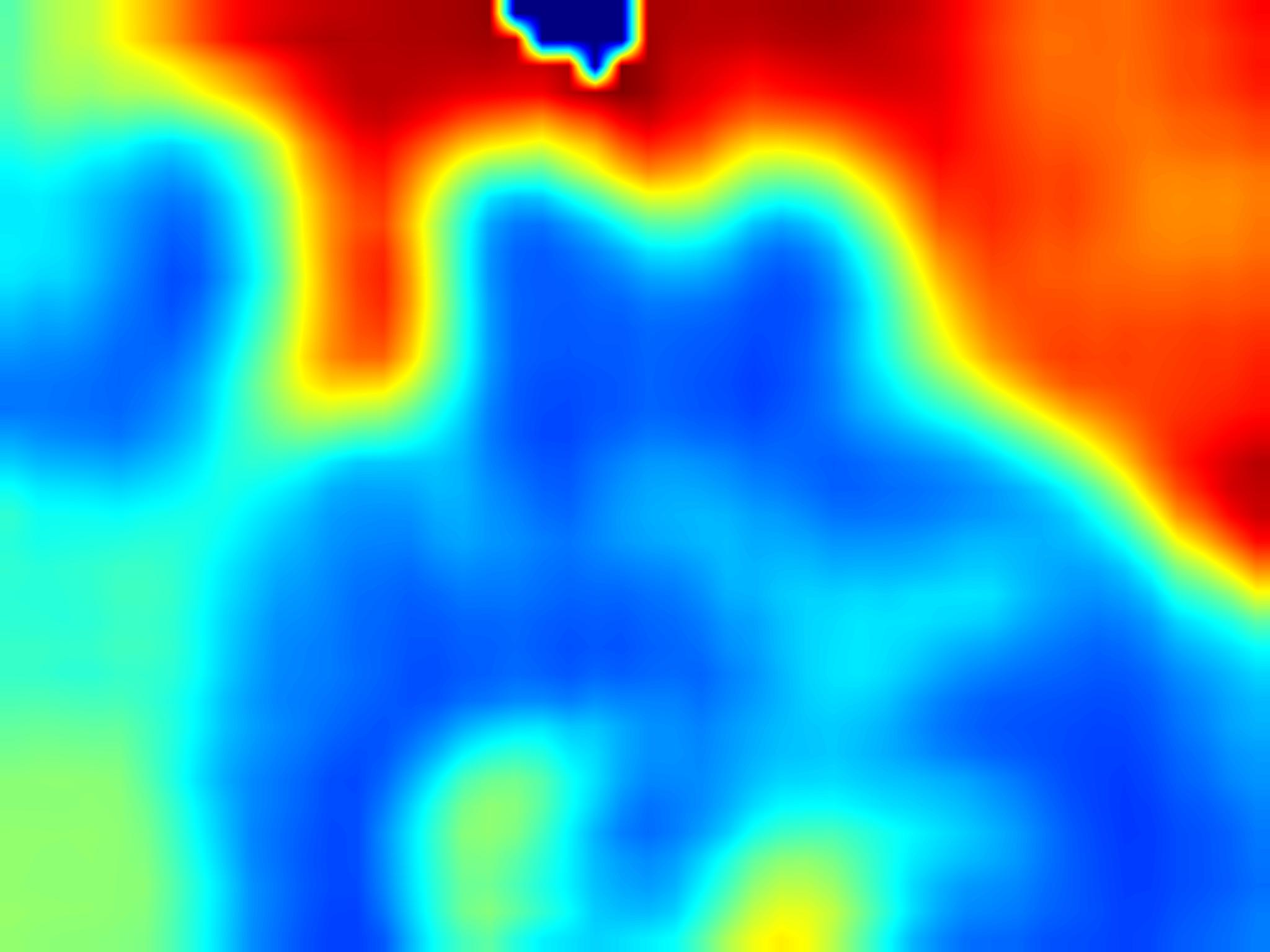}
	\includegraphics[width=0.19\linewidth]{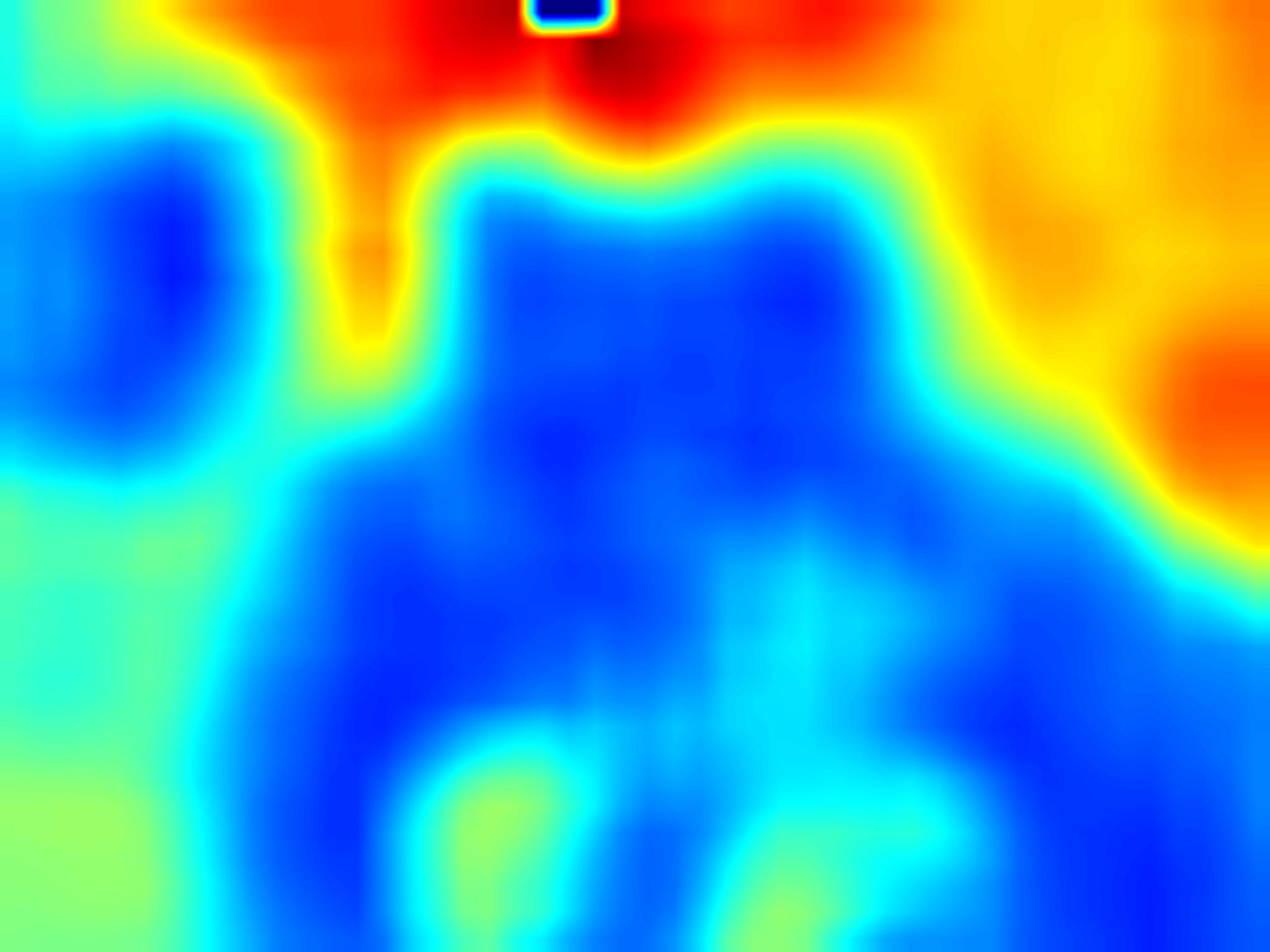}
	\includegraphics[width=0.19\linewidth]{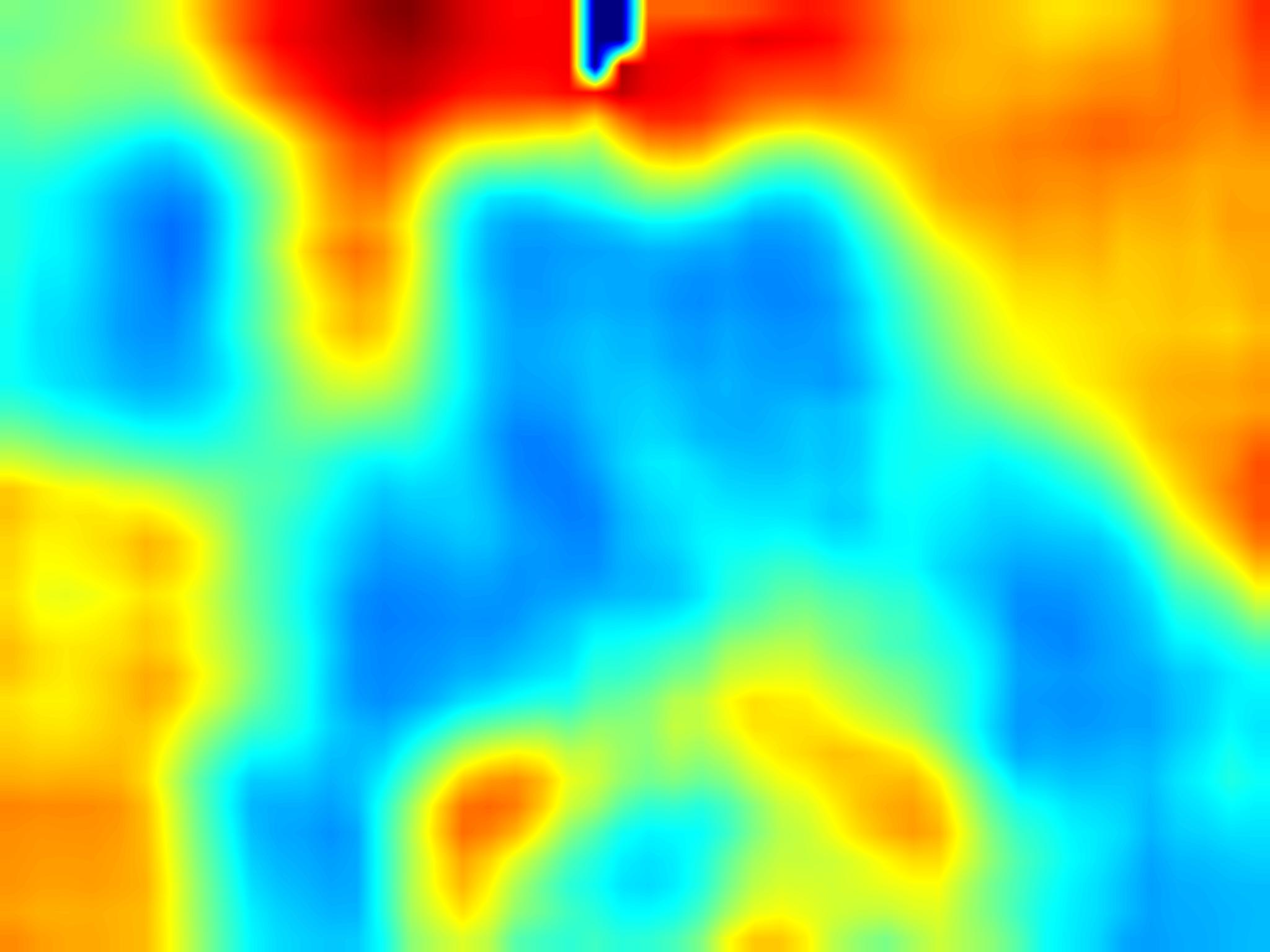}\\
	\vspace{0.1cm}
	\includegraphics[width=0.19\linewidth]{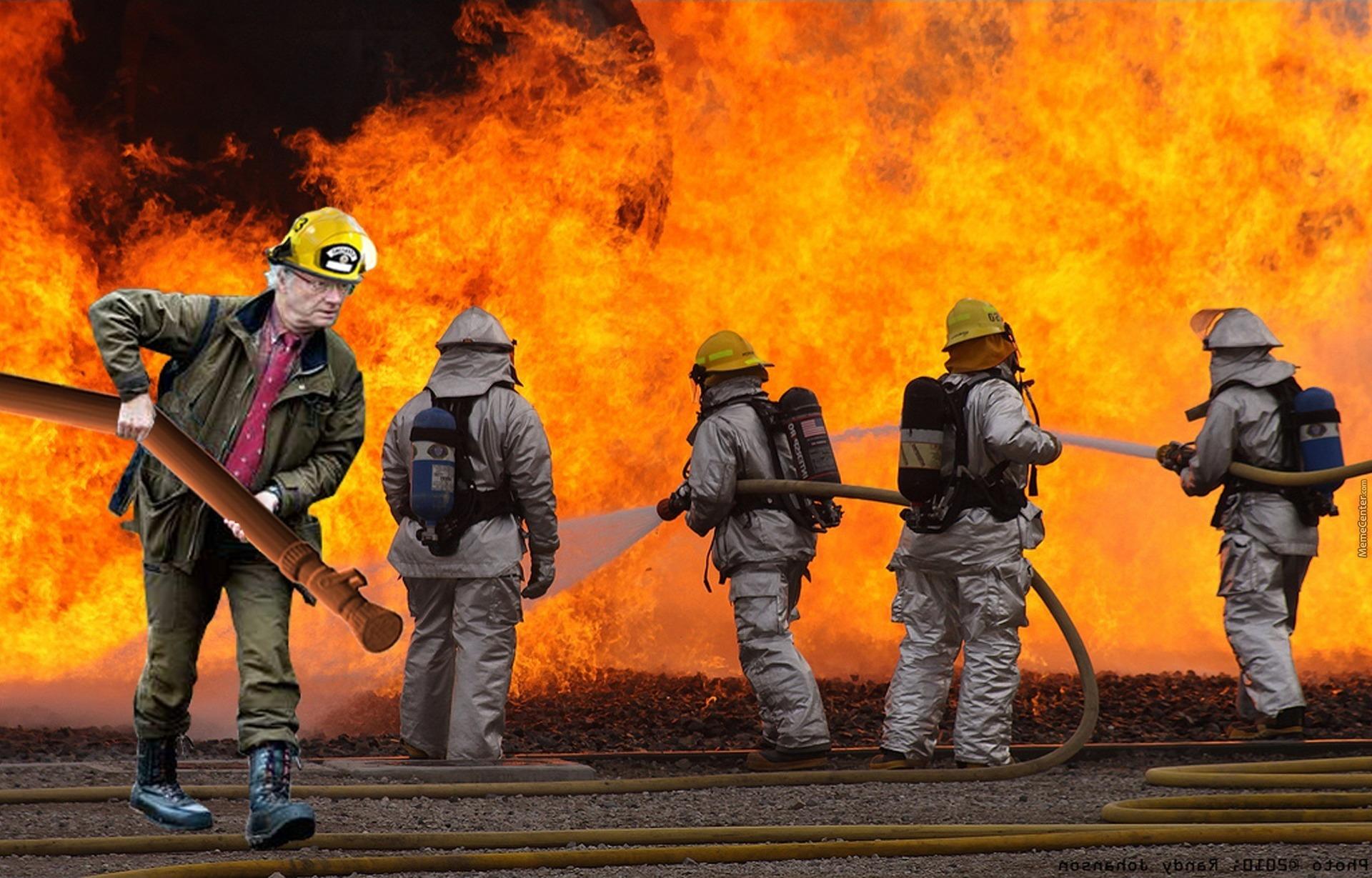}
	\includegraphics[width=0.19\linewidth]{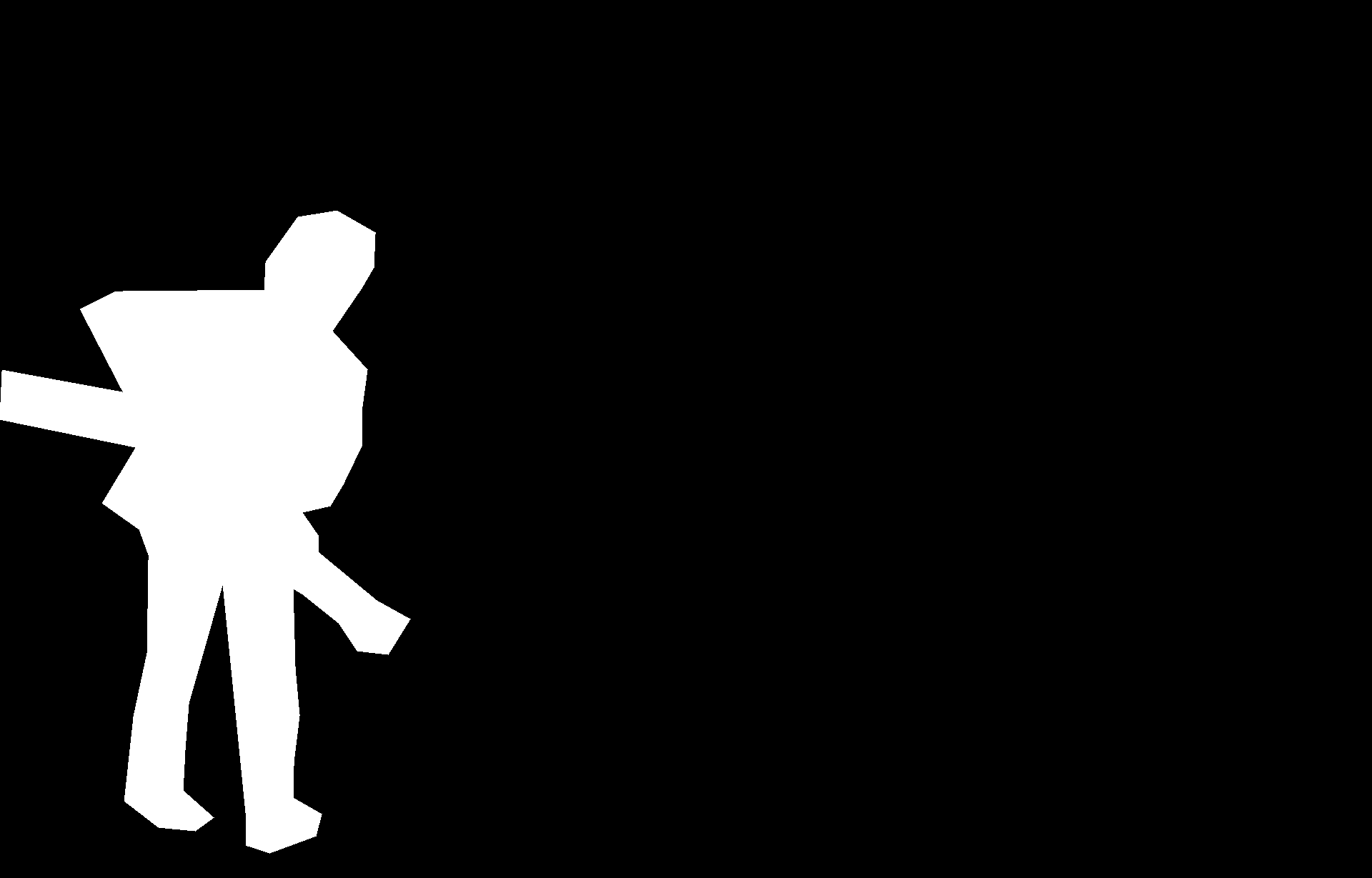}
	\includegraphics[width=0.19\linewidth]{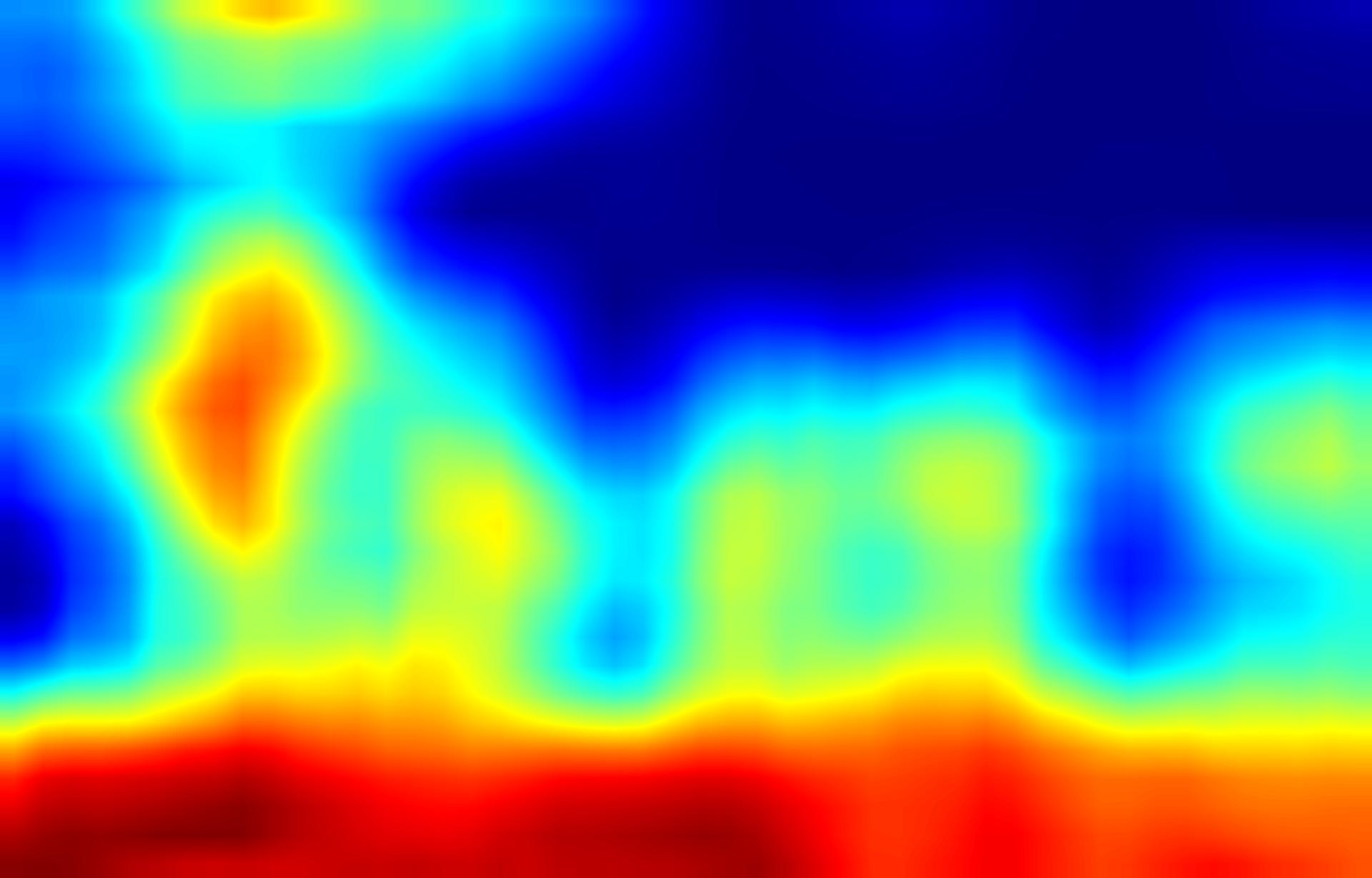}
	\includegraphics[width=0.19\linewidth]{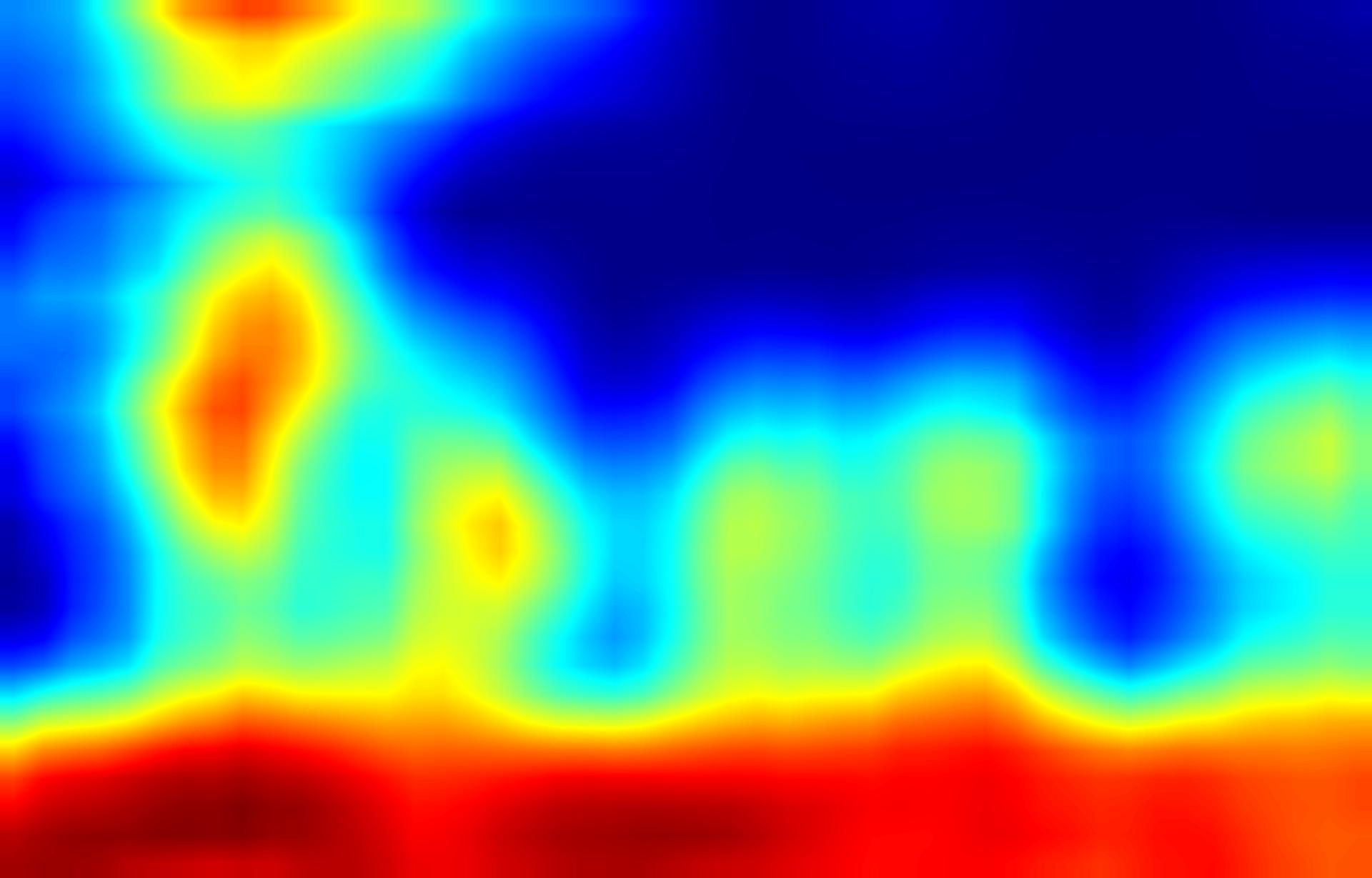}
	\includegraphics[width=0.19\linewidth]{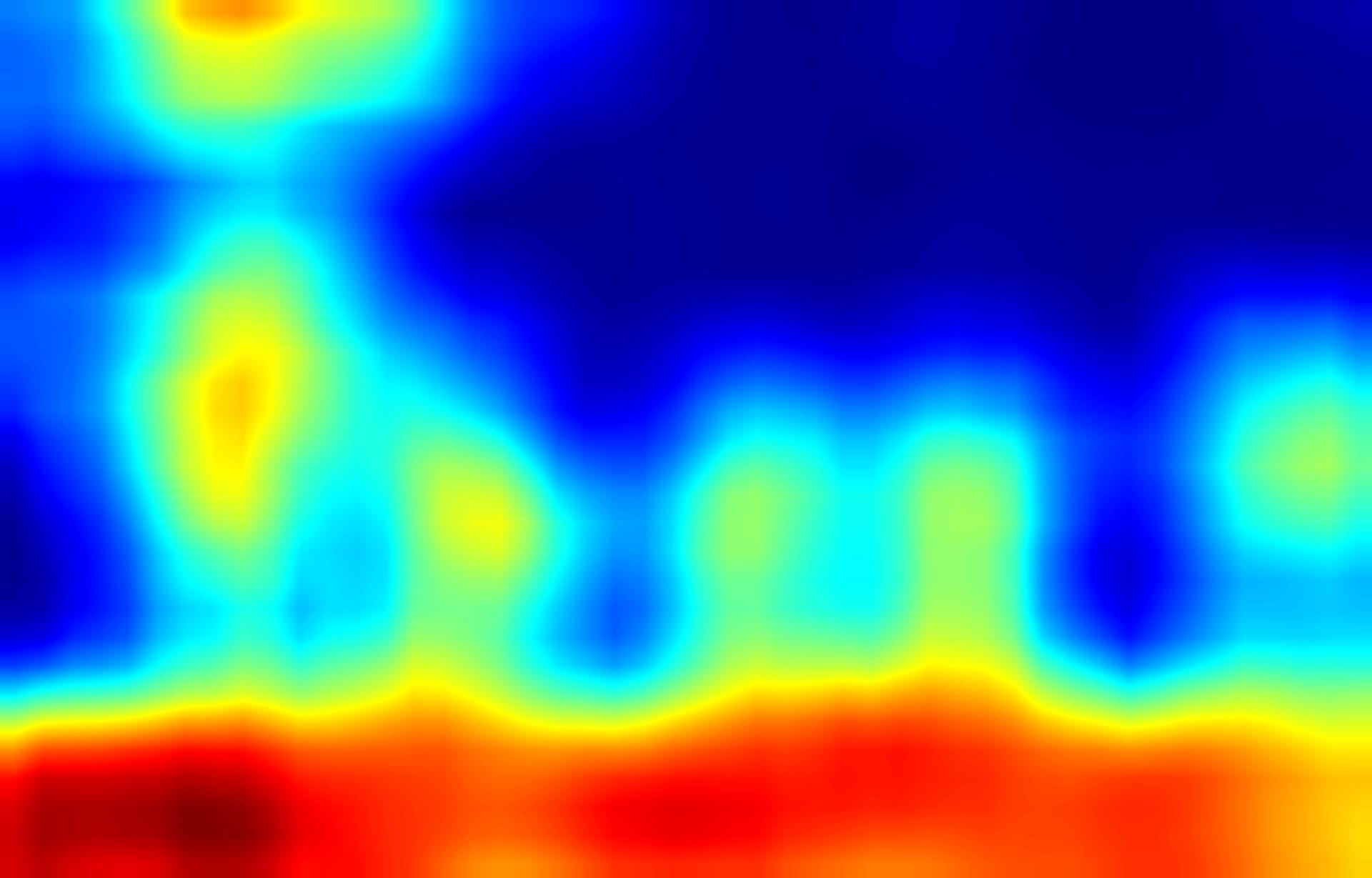}\\
	\vspace{0.1cm}	
	\includegraphics[width=0.19\linewidth]{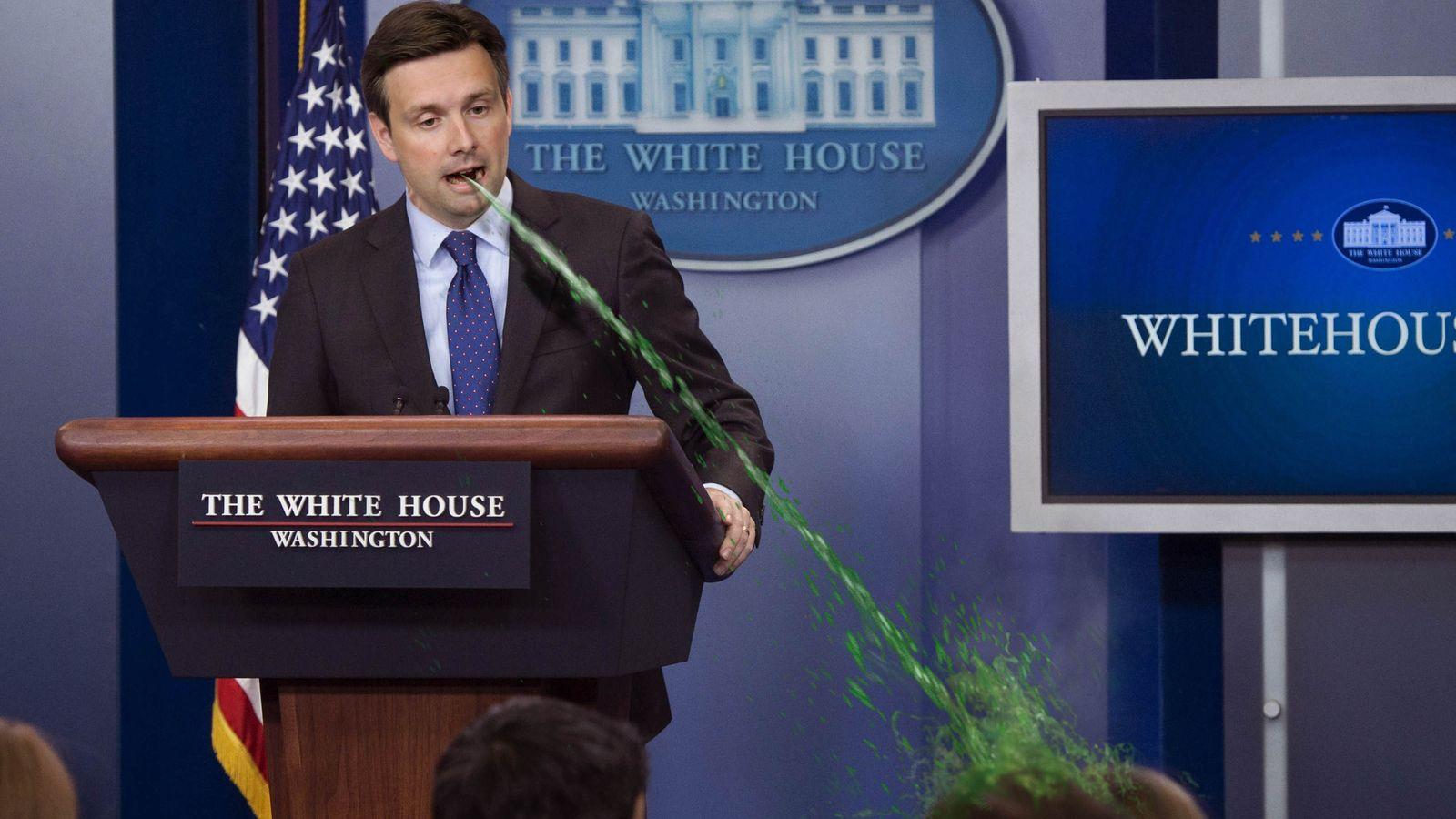}
	\includegraphics[width=0.19\linewidth]{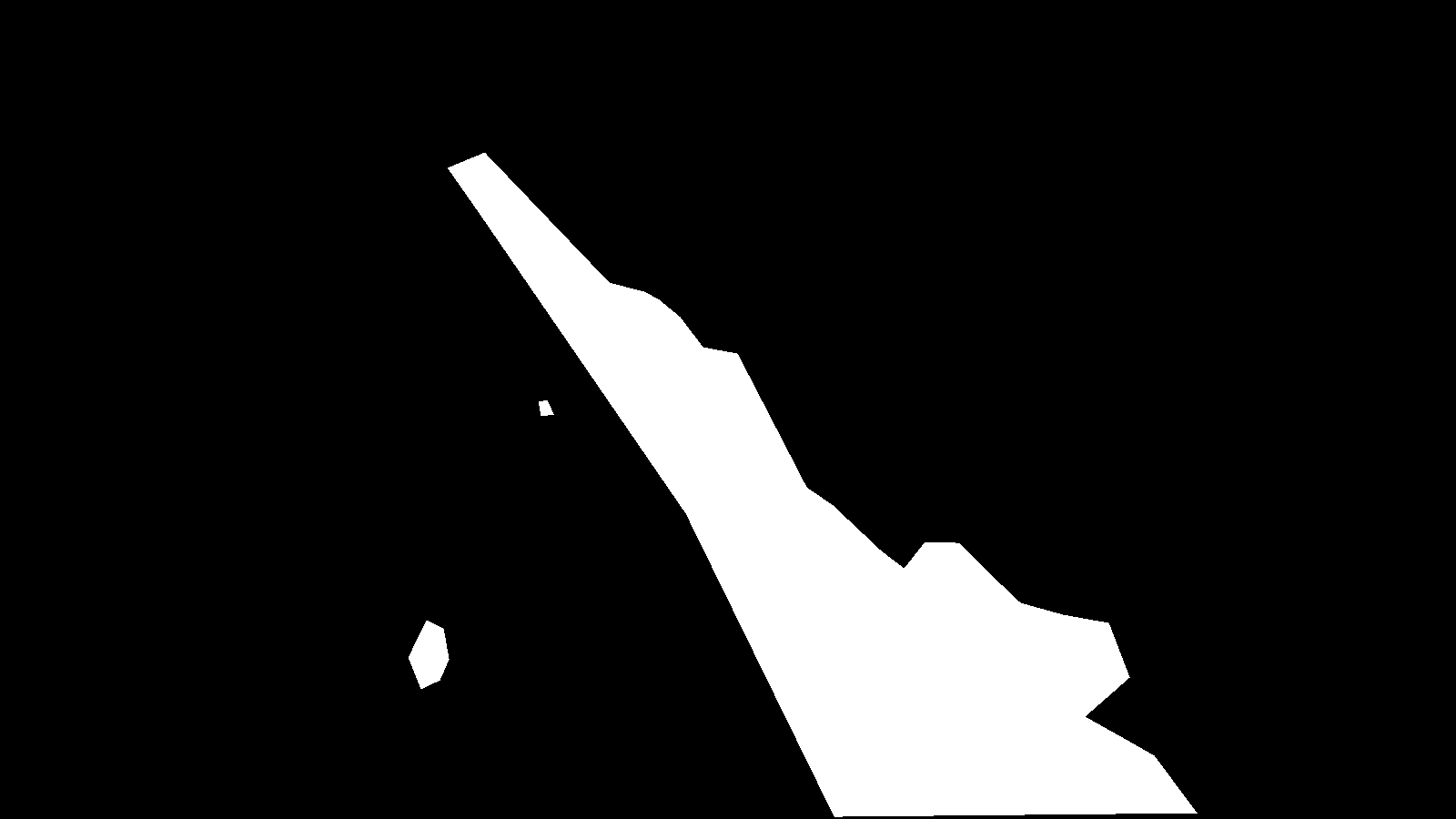}
	\includegraphics[width=0.19\linewidth]{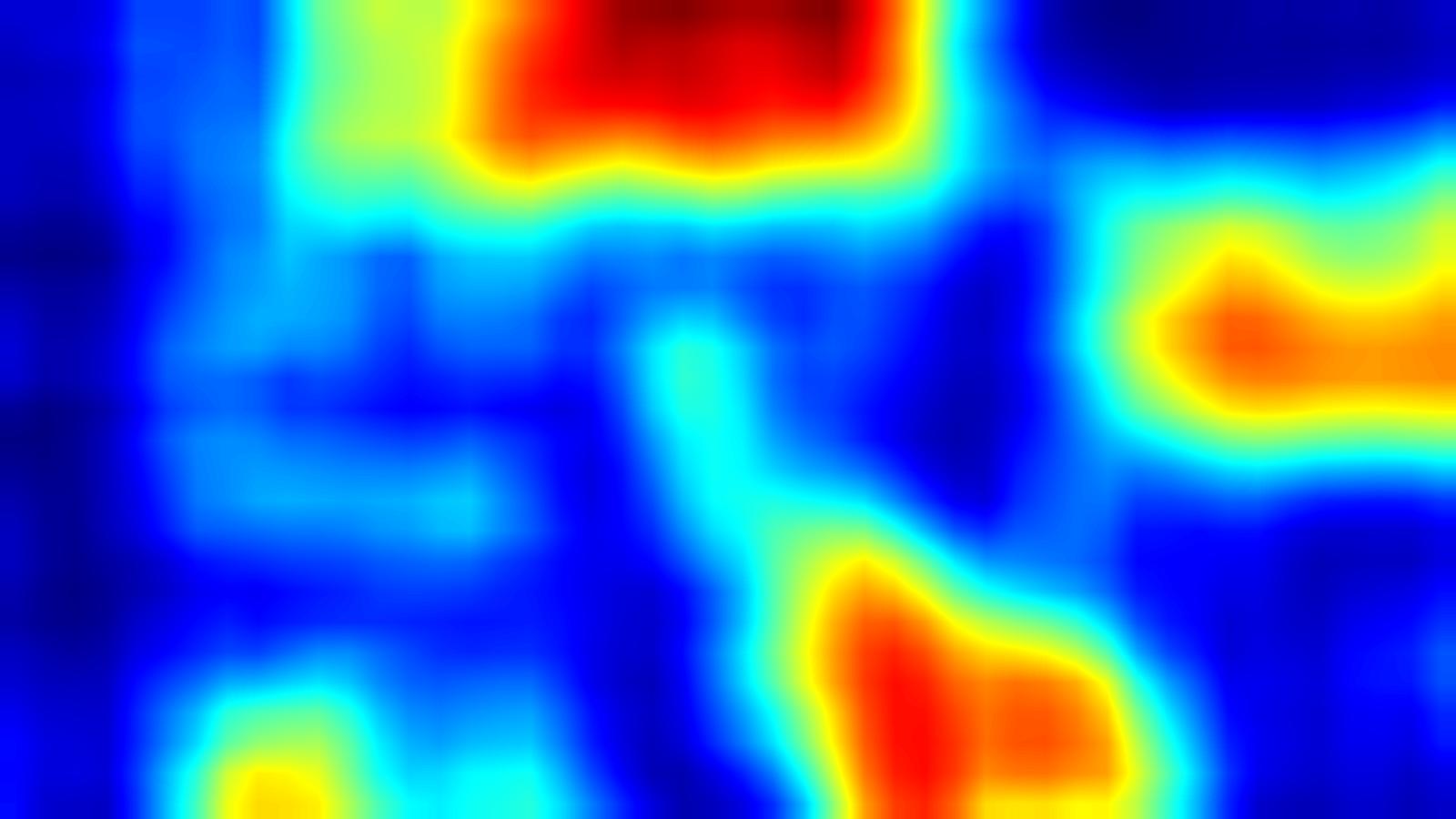}
	\includegraphics[width=0.19\linewidth]{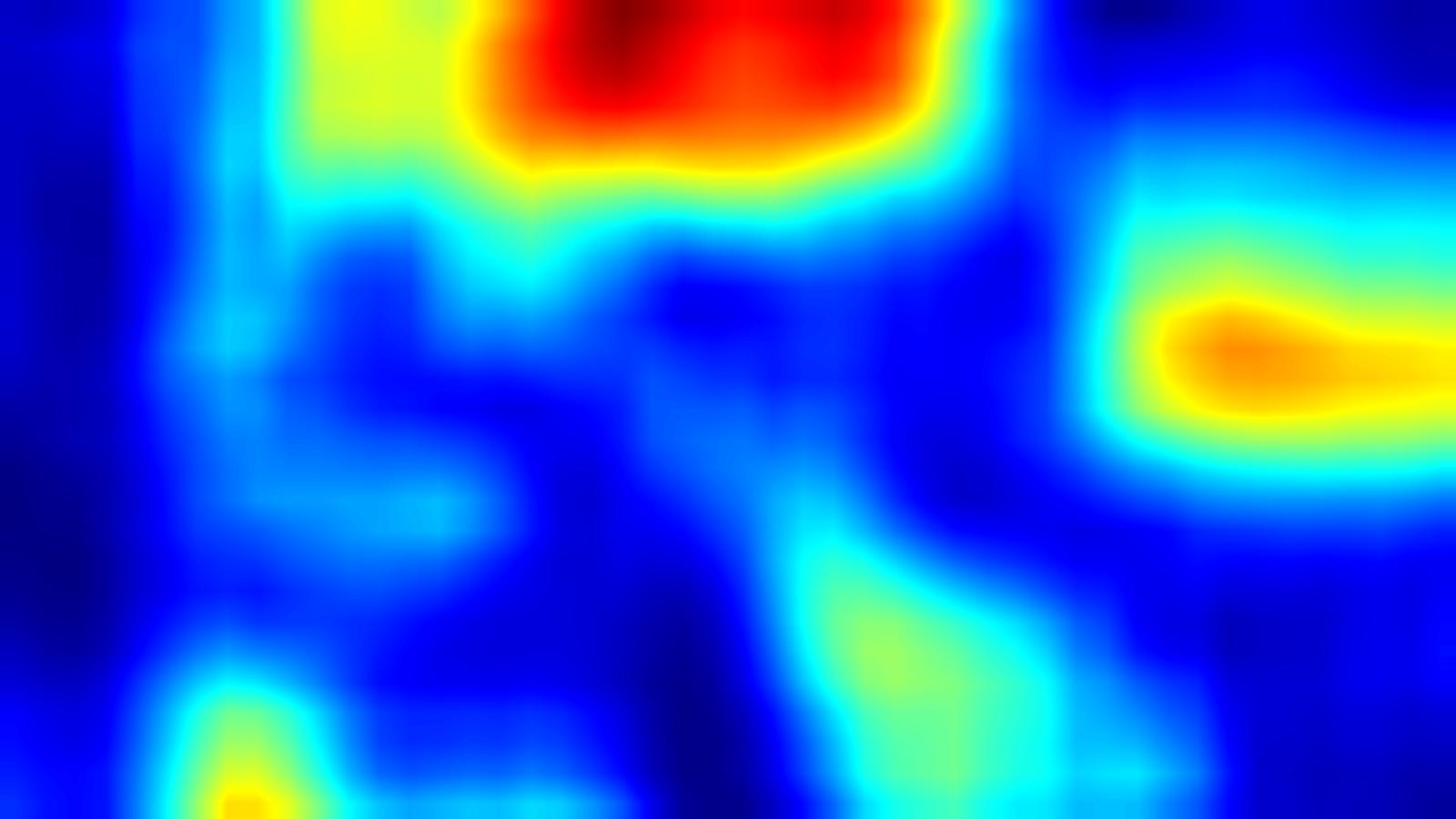}
	\includegraphics[width=0.19\linewidth]{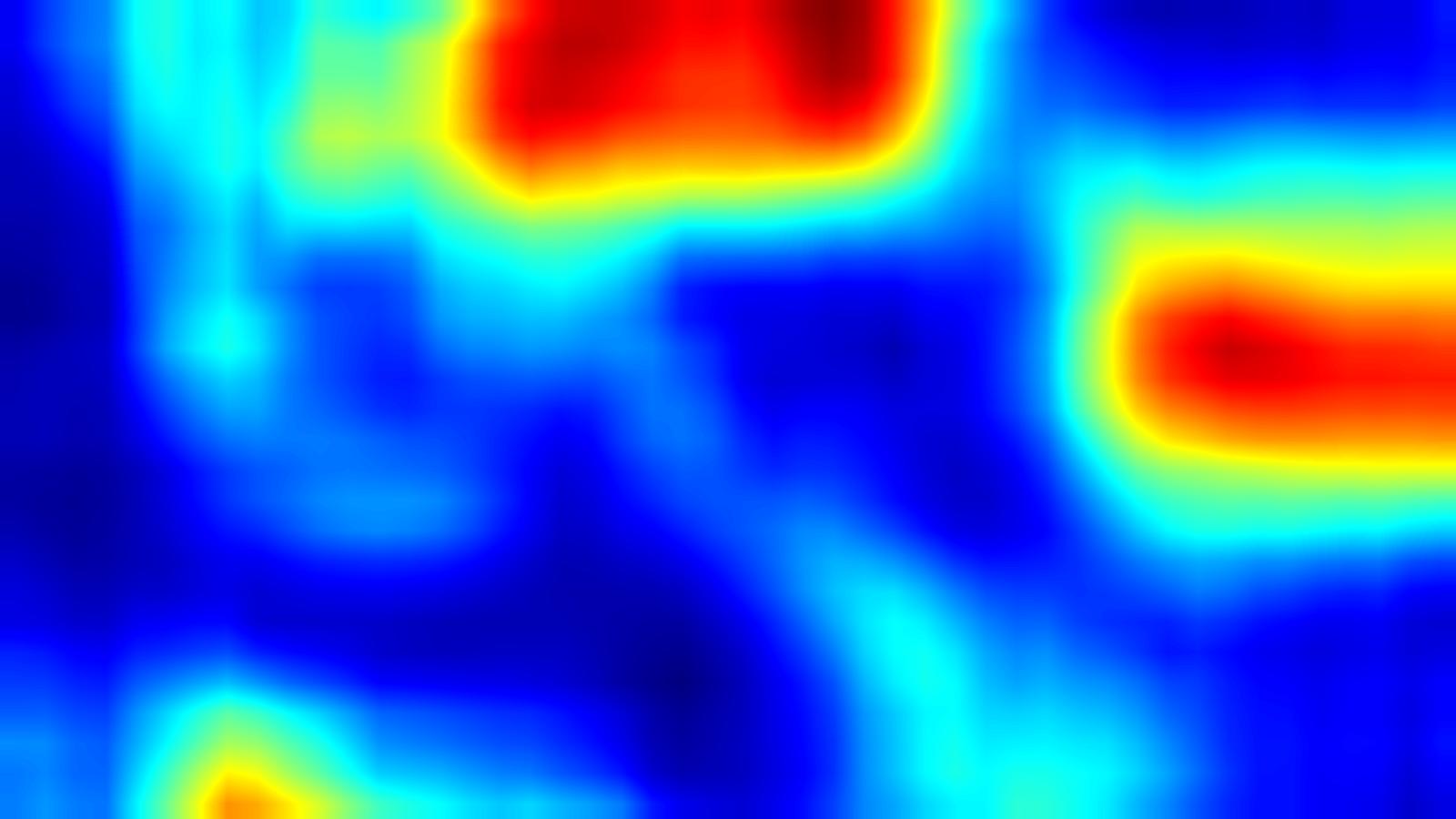}\\
	\caption{Failure cases with significant background contrast, evaluated for
		manipulation localization with MeanShift aggregation.}
	\label{fig:loc_failure_cases}
\end{figure}

%performs quite robust on all datasets, with the best results for Columbia, \newDatasetName{} and Aligned Scenes.
%The performances on the In-the-Wild dataset are overall lower than on the other datasets, suggesting that this dataset is more challenging.
%On DSO-1, statistical methods perform superior for HQ images, but rapidly lose performance for decreasing qualities.
%For localization with the proposed method, MeanShift aggregation generally performs better than medoid scores.  

\paragraph{Qualitative Localization Results} Figure~\ref{fig:loc_qual_examples} shows
qualitative results for the proposed method with MeanShift aggregation on
images from various datasets. In each row, results are shown for increasing
image degradations. Although the manipulations are diverse in size, color, and degree of texture, the proposed method successfully localizes the manipulated
regions throughout all qualities.

Figure~\ref{fig:loc_failure_cases} shows qualitative examples of representative
failure cases. Here, the color discrepancies between the inserted regions and
parts of the background are smaller than within-background differences.

\edited{Table~\ref{tab:mcc_qualitative_examples} lists the localization performance for the examples in Figs.~\ref{fig:loc_qual_examples} and~\ref{fig:loc_failure_cases} in terms of $\mathrm{MCC}$. In the top six rows, corresponding to Fig.~\ref{fig:loc_qual_examples}, the $\mathrm{MCC}$ is high, while for failure cases in the bottom three rows (corresponding to Fig.~\ref{fig:loc_failure_cases}), the $\mathrm{MCC}$ is rather low.}

% \begin{figure*}[t]
% 	\centering
% 	\begin{subfigure}[t]{0.19\textwidth}
% 		\centering
% 		\vspace{-3.2cm}
% 		\includegraphics[width=\textwidth,trim={17.5cm 0 0 0},clip]{figures/SplicingDetection_SyntheticSplices.pdf}
% 	\end{subfigure}
% 	\begin{subfigure}[t]{0.2\textwidth}
% 		\includegraphics[width=\textwidth,trim={0 0 13.5cm 0},clip]{figures/SplicingDetection_Columbia.pdf}
% 		\caption{Columbia\cite{DBLP:conf/icmcs/HsuC06}}
% 		\label{fig:det_columbia}
% 	\end{subfigure}
% 	\begin{subfigure}[t]{0.19\textwidth}
% 		\includegraphics[width=\textwidth,trim={1cm 0 13.5cm 0},clip]{figures/SplicingDetection_DSO-1.pdf}
% 		\caption{DSO-1\cite{DBLP:journals/tifs/CarvalhoRAPR13}}
% 		\label{fig:det_dso1}
% 	\end{subfigure}
% 	\begin{subfigure}[t]{0.19\textwidth}
% 		\includegraphics[width=\textwidth,trim={1cm 0 14.5cm 0},clip]{figures/SplicingDetection_SyntheticSplices.pdf}
% 		\caption{\newDatasetName}
% 		\label{fig:det_synthspl}
% 	\end{subfigure}
% 	\begin{subfigure}[t]{0.19\textwidth}
% 		\includegraphics[width=\textwidth,trim={1cm 0 13.5cm 0},clip]{figures/SplicingDetection_AlignedScenesDIDB.pdf}
% 		\caption{Aligned Scenes\cite{DBLP:conf/icassp/HadwigerBPR19}}
% 		\label{fig:det_alignedscenes}
% 	\end{subfigure}
% 	\caption{{\color{red}{hmm linien dicker, schrift etwas groesser?}} Comparison of splicing detection performance (ROC AUC) on various datasets. Color coding identical for all subfigures.}
% 	\label{fig:results_detection}
% \end{figure*}

\begin{figure*}[t]
	\centering
	\begin{subfigure}[t]{\linewidth}
		\includegraphics[width=\textwidth,trim={0 14.5cm 0 0},clip]{figures/Legend.pdf}
	\end{subfigure}\\
	\vspace{0.2cm}
%	\begin{subfigure}[t]{0.24\textwidth}
%		\includegraphics[width=\textwidth]{figures/SplicingDetection_Columbia.pdf}
%		\caption{Columbia\cite{DBLP:conf/icmcs/HsuC06}}
%		\label{fig:det_columbia}
%	\end{subfigure}
	\begin{subfigure}[t]{0.24\textwidth}
		\includegraphics[width=\textwidth]{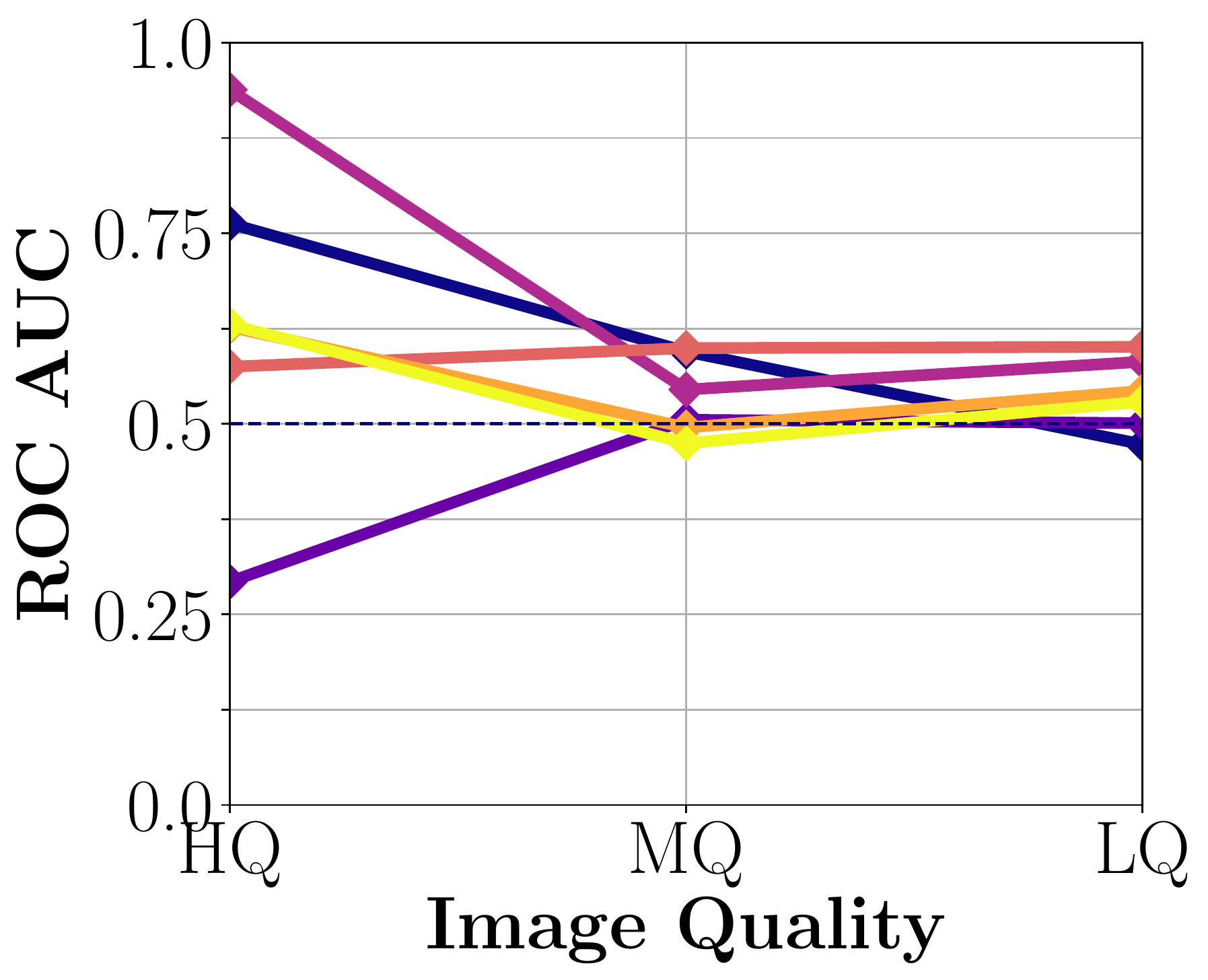}
		\caption{DSO-1\cite{DBLP:journals/tifs/CarvalhoRAPR13}}
		\label{fig:det_dso1}
	\end{subfigure}
	\begin{subfigure}[t]{0.24\textwidth}
		\includegraphics[width=\textwidth]{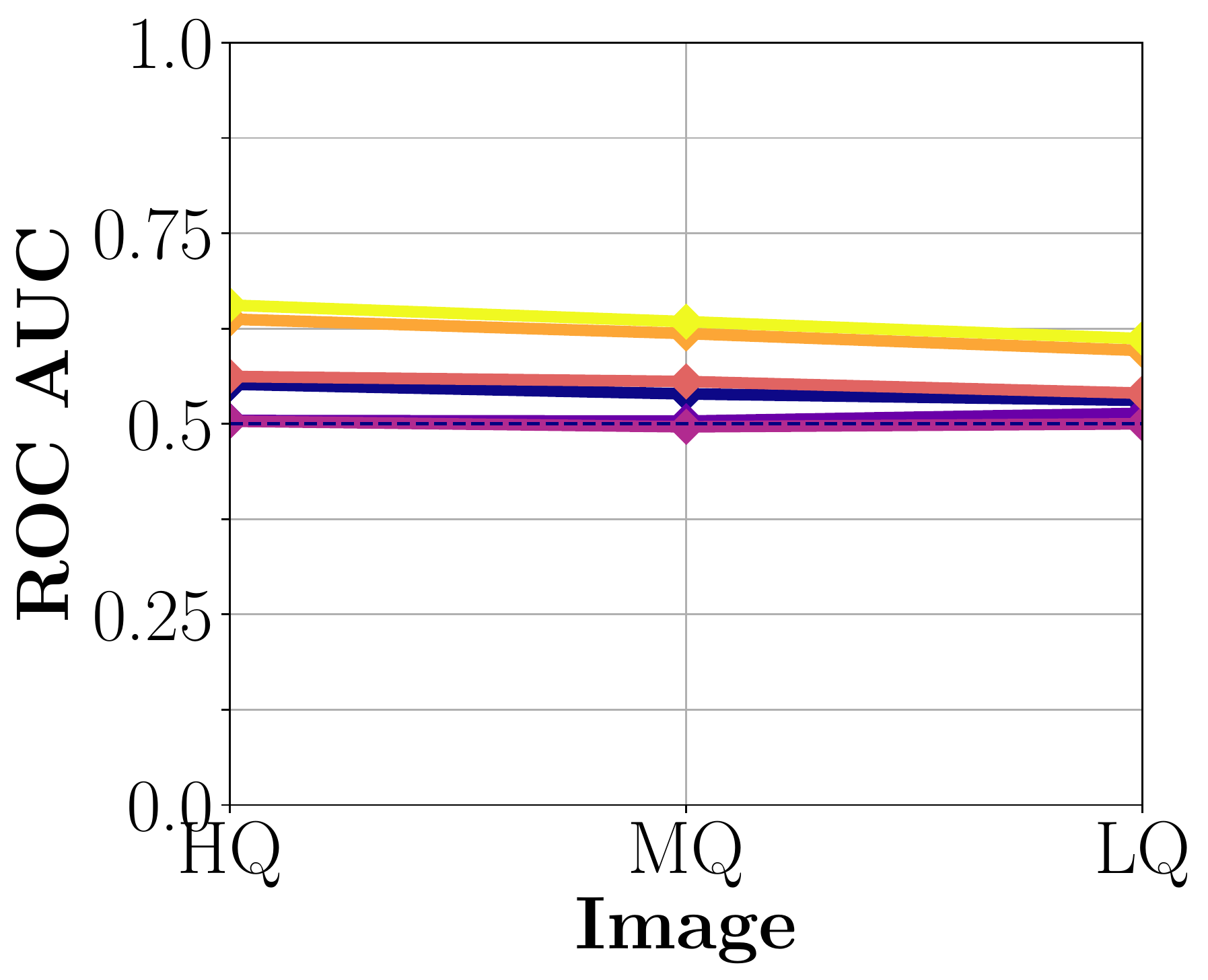}
		\caption{\newDatasetName}
		\label{fig:det_synthspl}
	\end{subfigure}
	\begin{subfigure}[t]{0.24\textwidth}
		\includegraphics[width=\textwidth]{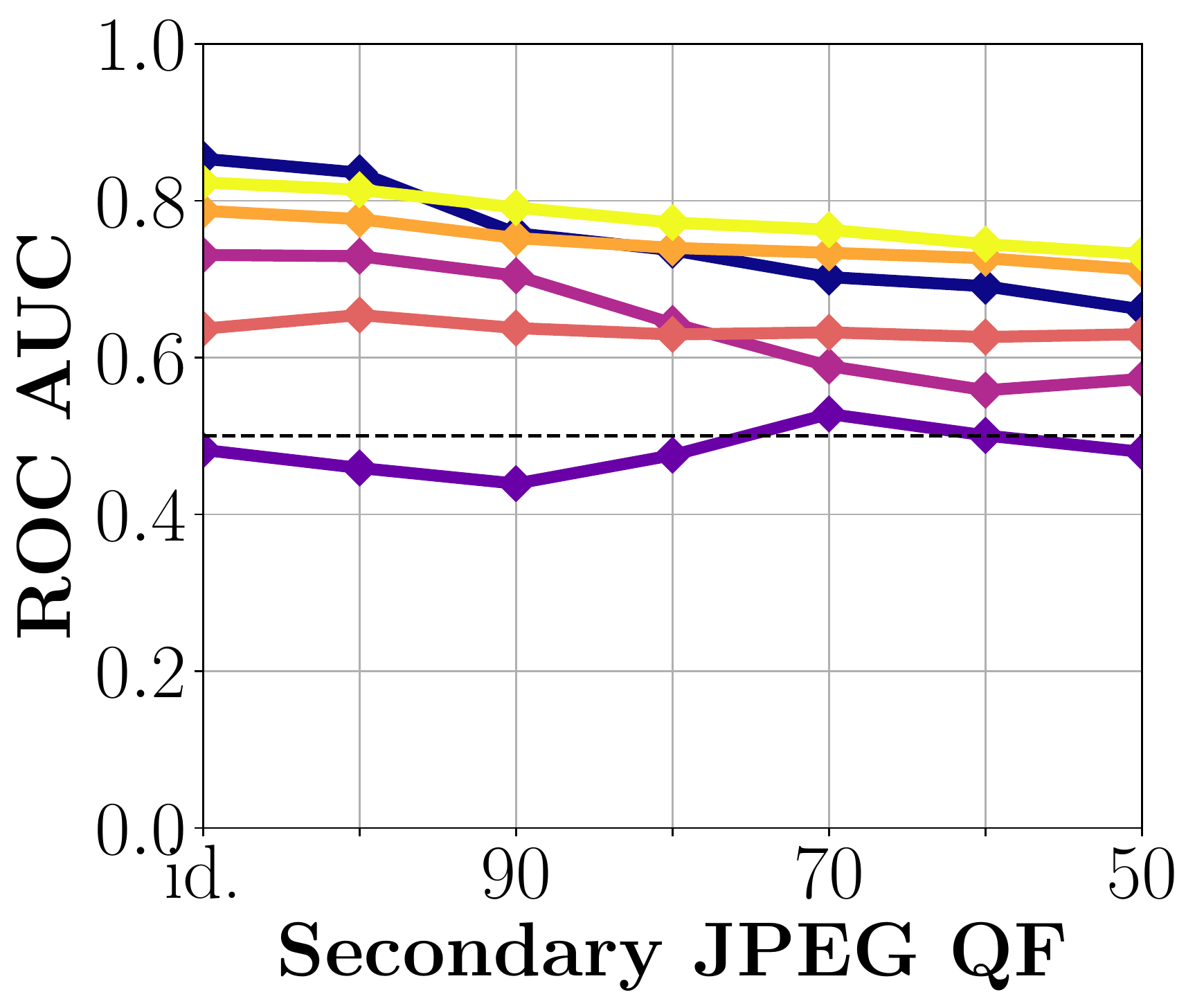}
		\caption{Aligned Scenes\cite{DBLP:conf/icassp/HadwigerBPR19}}
		\label{fig:det_alignedscenes}
	\end{subfigure}
	\caption{Comparison of splicing \textit{detection} ROC AUC on various datasets. The proposed method is specialized on \emph{localization}. Yet, also for detection it can contribute to the analysis of very low-quality images (see text for details).}
	\label{fig:results_detection}
\end{figure*}

\edited{
	\begin{table}
%		\vspace{-1cm}
		\caption{\edited{Performance for splicing localization for the qualitative examples in Figs.~\ref{fig:loc_qual_examples} and \ref{fig:loc_failure_cases} (from top to bottom) in terms of Matthews Correlation Coefficient.}}
		\label{tab:mcc_qualitative_examples}
		\centering
		\resizebox{\columnwidth}{!}{%
			\begin{tabular}{|c|c|c|c|c|}
				\hline
				\textbf{Dataset} & \textbf{Image Basename} & \textbf{HQ} & \textbf{MQ} & \textbf{LQ}\\
				\hhline{|=|=|=|=|=|}
				\multirow{3}{*}{In-The-Wild\cite{DBLP:conf/eccv/HuhLOE18}} & 
				im31\_edit6 & 0.959 & 0.958 & 0.964\\
				\cline{2-5}
				& im30\_edit2 & 0.825 & 0.831 & 0.853\\
				\cline{2-5}
				& 2251 & 0.570 & 0.583 & 0.620\\
				\hline
				\multirow{2}{*}{SplicedColorPipeline} & 210uxdvkfw3j0si\_1 & 0.943 & 0.927 & 0.946\\
				\cline{2-5}
				& 2gcbbedjgxy9v7j\_1 & 0.970 & 0.958 & 0.837\\	
				\hline
%				Columbia~\cite{DBLP:conf/icmcs/HsuC06} & \scriptsize{canonxt\_kodakdcs330\_sub\_02} & 0.973 & 0.828 & 0.753\\
%				\hline
				Align. Sce.~\cite{DBLP:conf/icassp/HadwigerBPR19} & n8tn8fkkx29a & 0.998 & 0.994 & 0.998\\
				\hhline{|=|=|=|=|=|}
				DSO-1~\cite{DBLP:journals/tifs/CarvalhoRAPR13} & splicing-02 & 0.276 & 0.248 & 0.228\\
				\hline
				\multirow{2}{*}{In-The-Wild\cite{DBLP:conf/eccv/HuhLOE18}} & 
				im32\_edit7 & 0.263 & 0.256 & 0.232\\
				\cline{2-5}
				& 5482 & 0.367 & 0.217 & 0.210\\
				\hline
		\end{tabular}}%
	\end{table}
}

\paragraph{Splicing Detection} The proposed method, and the methods for
comparison, are primarily designed for manipulation localization.
However, for completeness, we also report results for splicing detection,
although we acknowledge that localization-based methods can be expected to
perform worse on this task than specialized methods for splicing detection.

For each algorithm, we compute the average over the heatmap according to
Eqn.~(\ref{eq:detection_score}) as a per-image manipulation score. \reedited{As it is common for splicing detection~\cite{DBLP:conf/wifs/CozzolinoPV15, DBLP:conf/cvpr/StantonHM19, DBLP:journals/tifs/MaternRS20}, for each quality level the performance is measured via the ROC AUC}. The
``In-the-Wild'' dataset is not part of this experiment, as it does not contain
any pristine images. The results are shown in Fig.~\ref{fig:results_detection}. 
%Figure~\ref{fig:det_columbia} shows the performance on the Columbia dataset.
%\ffn{} and \forsim{} perform best on high-quality images with AUCs of $0.975$
%and $0.924$. However, their performance decreases with decreasing image
%quality. The proposed method performs slightly worse on HQ images, with AUCs of
%$0.888$ and $0.878$ for MED and MSA.  However, it is barely affected by the
%image quality, such
%that it eventually performs best on low-quality images.  \lcr{} is also robust
%to quality degradations, but scores overall lower than the proposed approach
%(LQ: $0.761$).
We found that \noisep{} exhibits large variations in the values across
heatmaps, such that we could not find a competitive detection threshold for
this localization method.

%best
%performance for HQ images is obtained by \ffn{} ($0.975$), followed by
%\forsim{} ($0.924$), the proposed approach with AUC $0.888$ and $0.878$, for
%MED and MSA, and \lcr{} ($0.776$).
%For decreasing qualities, the performance of \forsim{} decreases steeply (MQ: $0.527$), and \ffn{} also loses performance to some extent (LQ: $0.828$).
%The proposed method, however, is robust with practically identical performance for degraded images (LQ MED: $0.862$, LQ MSA: $0.851$).
%\noisep{} yields heatmaps where value ranges vary with several orders of magnitude between images, thus, we normalize all heatmaps to $[0,1]$. This potentially causes a loss in information otherwise useful for detection, which could explain why this method in general performs poorly for detection in our evaluation.

Figure~\ref{fig:det_dso1} shows the results for the DSO-1 dataset.
The best performance for HQ images is obtained for the statistics-based
\forsim{} (HQ: $0.938$), where the proposed method achieves an AUC of $0.63$.

Figure~\ref{fig:det_synthspl} shows the results for the SplicedColorPipeline
dataset.
Here, the proposed method performs best throughout all quality levels (HQ MED:
$0.656$, LQ MED: $0.612$).  The second best results are obtained for the
color-based \lcr{}, followed by \ffn{} that uses various traces. The
statistics-based \noisep{} and \forsim{} only achieve random guessing
performance, which is expected since the different camera pipelines only
modulate color properties, but do not affect other statistical traces. \edited{The overall rather low AUC of the proposed method on this dataset can be attributed to difficulty of finding a global threshold for the detection score. The fact that the size of the spliced image regions is not limited during creation of the dataset leads to a large variance in the region sizes, and hence detection scores.}

Figure~\ref{fig:det_alignedscenes} shows the results for the Aligned Scenes dataset.
For single compressed and double compressed images with secondary quality
$100$, the best results are obtained for \ffn{} (``$\mathrm{id.}$'': $0.854$).
For all stronger compressions, the proposed method yields the best results
medoid-based scores MED, e.g., for secondary compression quality $90$ an AUC of
$0.791$.

In summary, for manipulations that involve the camera pipeline, the proposed
algorithm performs best for non-ideal images with some degradations.  For high
quality images and other manipulations, the proposed algorithm is outperformed
by algorithms that include statistical traces (\ffn{}, \forsim{}).  For
increasing compression, the performance of the proposed approach remains
remarkably robust on all datasets except on DSO-1. Also, medoid-based scores
generally perform better than MeanShift for detection.
\reedited{We also calculated the true positive rate at a fixed 5\% false alarm
rate, $\mathrm{TPR}_{5\%}$. The relative performances are almost identical to the AUCs in
Fig.~\ref{fig:results_detection}. At LQ and JPEG 50
settings, example $\mathrm{TPR}_{5\%}$ for our method with medoid-based scores are 3.0\%, 10.5\%,
and 19.3\% on the three datasets, which again shows that method works better for
splicing localization than detection.} 

\subsection{Application of the Learned Embeddings to Illuminant Estimation and Camera Identification}
\label{sec:information_content_features}

This experiment shows that the proposed metric space contains information on
camera properties and scene colors. More specifically,
Sec.~\ref{sec:related_tasks} and Fig.~\ref{fig:unification_camera_illuminant}
indicate that the space can be marginalized for illuminant estimation and for
camera identification.

%for related tasks, and . For this, we employ the learned embeddings as features for other tasks. The performance for a specific task then indicates to which extent the CNN relies on the information required for this task, rather than discarding it as unnecessary.

\paragraph{Illuminant Estimation}

%We investigate the illuminant information contained in the learned embeddings by employing them as features for illuminant estimation. 
We apply the learned embeddings as features for illuminant estimation. The performance is taken as \edited{a} surrogate measure for the illuminant information contained in the embeddings. 
The experiment is performed on the Gehler-Shi dataset~\cite{DBLP:conf/cvpr/GehlerRBMS08, shi_reprocessed} using threefold cross-validation. We train a 3-layer neural network, with all hyperparameters tuned on the first validation fold. The images are upscaled to $1,536$ pixels in the larger dimension, and gamma-corrected with $\gamma=0.5$ before patch extraction and computation of the embeddings. The network consists of three dense layers with $64$, $32$ and $2$ units, respectively, to regress the illuminant color in $L_uL_v$-space, as used in~\cite{barron2015convolutional}. The weights are initialized with Glorot Uniform initialization~\cite{DBLP:journals/jmlr/GlorotB10}. Each layer uses Scaled Exponential Linear Units (SeLU)~\cite{DBLP:conf/nips/KlambauerUMH17} with Keras default values, except for the output layer with no nonlinearity. We use Mean Squared Error loss, and train with the Adam~\cite{kingma2014adam} optimizer with moments $\beta_1 = 0.9$ and $\beta_2 = 0.999$ and batches of size $64$. The learning rate is set to $\alpha=10^{-4}$ and halved if there is no improvement for two consecutive epochs, and we apply Early Stopping. 

\begin{table}[t]
	\centering
	\caption{Statistics of the angular error (in degrees) for illuminant
	estimation. While the proposed method is outperformed by several dedicated
	illuminant estimators, it exhibits sensitivity to the scene illuminant (see text for details).}
	\label{tab:illuminant_estimation}
	\begin{tabular}{|c||c|c|c|c|c|c|}
		\hline
		\textbf{Algorithm} & \textbf{Mean} & \textbf{Median} & \textbf{\thead{Tri-\\Mean}} & \textbf{\thead{Best\\$25\%$}} & \textbf{\thead{Worst\\$25\%$}} & \textbf{\thead{Geom.\\Mean}}\\
		\hhline{|=#=|=|=|=|=|=|}
		SVR~\cite{funt2004estimating} & 8.08 & 6.73 & 7.19 & 3.35 & 14.89 & 7.21\\
		% GW~\cite{} & 6.36 & 6.28 & 6.28 & 2.33 & 10.58 & 5.73\\
		\hline
		EBG~\cite{gijsenij2010generalized} & 6.52 & 5.04 & 5.43 & 1.90 & 13.58 & 5.40\\
		\hline
		FOGE~\cite{van2007edge} & 5.33 & 4.52 & 4.73 & 1.86 & 10.03 & 4.63\\		
		\hline
		SOGE~\cite{van2007edge} & 5.13 & 4.44 & 4.62 & 2.11 & 9.26 & 4.60\\
		\hline
		NIS~\cite{gijsenij2010color} & 4.19 & 3.13 & 3.45 & 1.00 & 9.22 & 3.34\\
		%\vdots & \vdots & \vdots & \vdots & \vdots & \vdots & \vdots\\
		\hline
		CCC~\cite{barron2015convolutional} & 1.95 & 1.22 & 1.38 & 0.35 & 4.76 & 1.40\\
		\hline
		CNN~\cite{shi2016deep} & 1.90 & 1.12 & 1.33 & 0.31 & 4.84 & 1.34\\
		\hline
		FFCC~\cite{barron2017fast} & \textbf{1.61} & \textbf{0.86} & \textbf{1.02} & \textbf{0.23} & \textbf{4.27} & \textbf{1.07}\\
		%\hhline{|=#=|=|=|=|=|=|}
		\hline
		Ours & 5.39 & 3.21 & 3.96 & 1.02 & 13.31 & 3.92\\
		\hline
	\end{tabular}
\end{table}

%The ground truth for this dataset is subject to ongoing
%debate~\cite{hemrit2018rehabilitating, banic2019past}. For our purpose, it is
%preferable to operate on the white-balanced images {\color{red}{double-check}}
%to use the camera output after in-camera processing.  We determine the
%RGB-space ground truth $w_j^{\star}\in[0,1]^3$ by median averaging the central
%region of the white patch of the color chart in each image $j$.
The $L_uL_v$-space predictions $\tilde{w}_i\in[0,1]^2$ for patch $i$ are converted 
to RGB-space, $w_i\in[0,1]^3$, see~\cite[Eqn. 6]{barron2015convolutional}. Then, the RGB-space prediction $w$ of an image is computed as median over all patch predictions, and compared to the ground truth $w^\star$ using angular error
\begin{equation}
	\varepsilon = \frac{180^{\circ}}{\pi}\cdot\cos^{-1}\left(\frac{w^{\top}w^{\star}}{\lVert{w}\rVert\lVert{w^{\star}}\rVert}\right)\enspace.
\end{equation}
As a reference, we compare against selected dedicated illuminant estimation methods: Support Vector Regression (SVR)~\cite{funt2004estimating}, Edge-based Gamut (EBG)~\cite{gijsenij2010generalized}, First (FOGE) and Second (SOGE) Order Grey Edge~\cite{van2007edge}, Natural Image Statistics (NIS)~\cite{gijsenij2010color}, a deep learning-based method (CNN)~\cite{shi2016deep}, Convolutional Color Constancy (CCC)~\cite{barron2015convolutional} and Fast Fourier Color Constancy~\cite{barron2017fast}. Table~\ref{tab:illuminant_estimation} lists statistics on the angular errors over all test images and folds, where the last column is the geometric mean of the other five statistics, compare~\cite{barron2015convolutional, barron2017fast}. While the best results are achieved by the recent method FFCC, the learned embeddings outperform dedicated  model-based approaches (SVR, EBG, FOGE, SOGE). This suggests that our CNN indeed \edited{extracts} color information to separate different imaging pipelines.

\paragraph{Camera Identification}

\begin{figure}
%    \centering
	\begin{tabular}{C{0.2\linewidth}C{0.2\linewidth}C{0.2\linewidth}C{0.2\linewidth}}
		%{C{0.195\linewidth}C{0.195\linewidth}C{-2\tabcolsep}C{0.195\linewidth}C{0.192\linewidth}}
%		{>{\centering}C{0.2475\linewidth-2\tabcolsep}@{}
%			        >{\centering}C{0.2475\linewidth}@{}
%			        >{\centering}C{0.01\linewidth}@{}
%			        >{\centering}C{0.2475\linewidth}
%			        >{\centering\arraybackslash}C{0.2475\linewidth}@{}}
%	a & b & c & d & e\\
%	\begin{tabular}{p{\dimexpr 0.24\linewidth-2\tabcolsep}
%			        p{\dimexpr 0.24\linewidth-2\tabcolsep}
%			        p{\dimexpr 0.24\linewidth-2\tabcolsep}
%			        p{\dimexpr 0.24\linewidth-2\tabcolsep}}\\
%
	\multicolumn{2}{c}{\small{Samsung Galaxy A6}} & \multicolumn{2}{c}{\small{Huawei P9 lite}}\\
    %\cline{1-2}\cline{4-5}	
    \hline
	\small{Device 1} & \small{Device 2} & \hspace{0.2cm}\small{Device 1} & \hspace{0.2cm}\small{Device 2}\\
%	\vspace{0.1cm}
%	& &  \multicolumn{3}{c}{\small{Results on}} \\\cline{3-5}
%	\small{Image\phantom{m}} & \small{Ground truth} & \small{HQ} & \small{MQ} & \small{LQ} \\
	\end{tabular}
	%\vspace{0.4cm}
	\begin{subfigure}[t]{\linewidth}
	\centering
	\includegraphics[width=0.2405\linewidth]{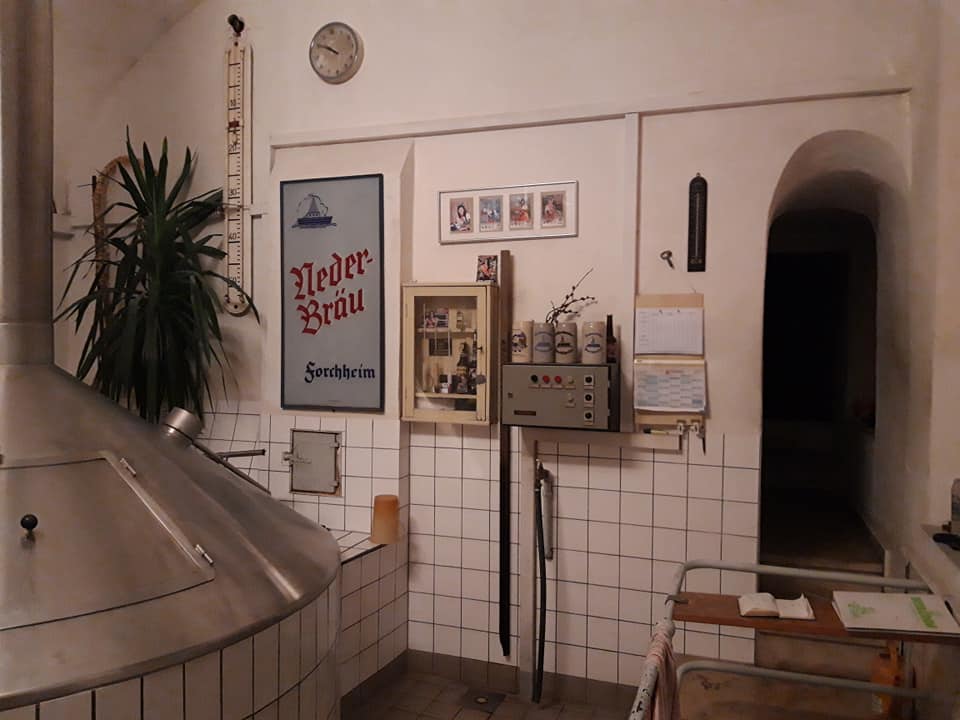}
	\includegraphics[width=0.2405\linewidth]{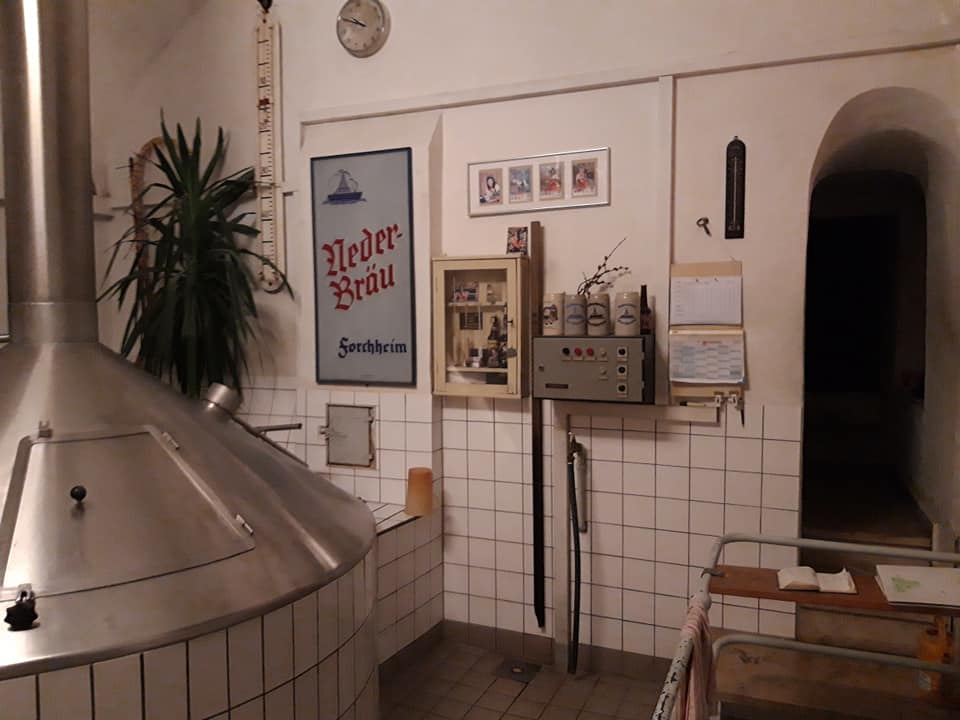}
	\includegraphics[width=0.2405\linewidth]{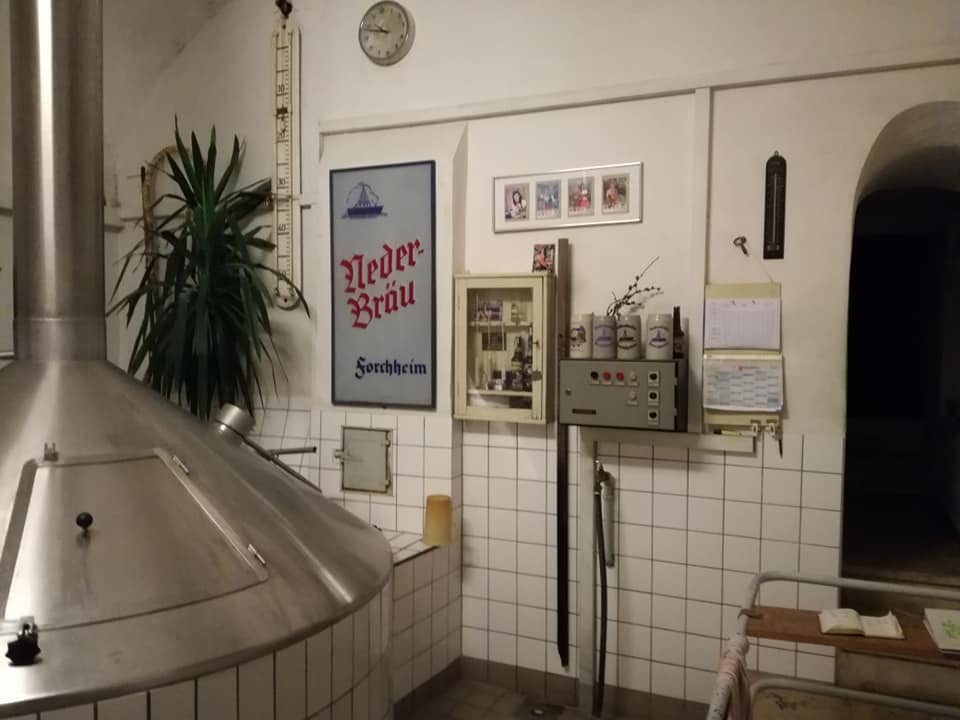}
	\includegraphics[width=0.2405\linewidth]{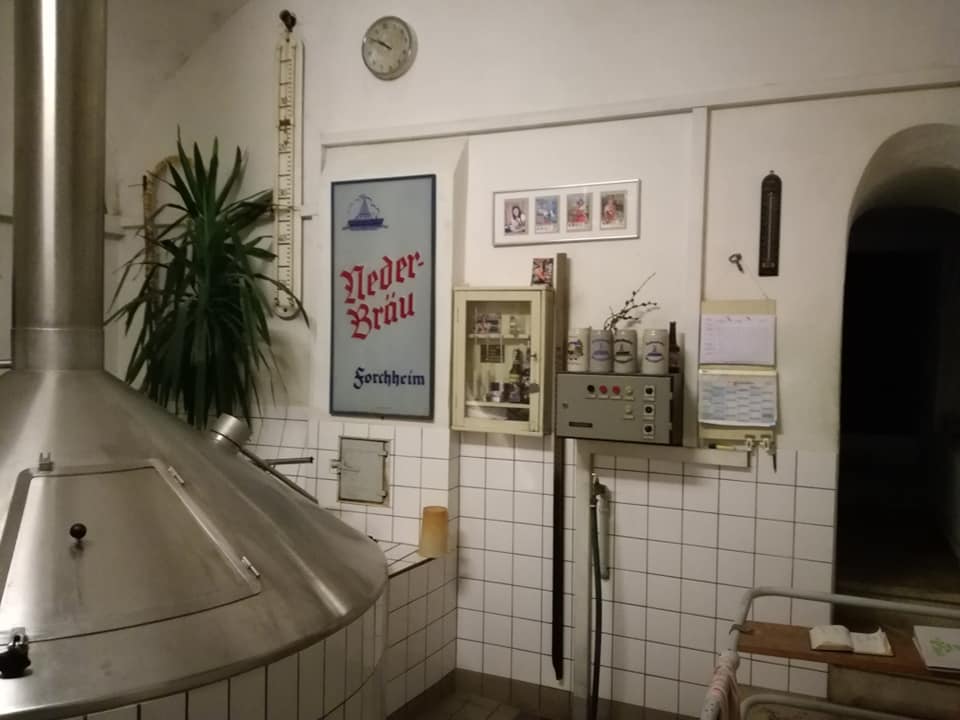}\\
	\vspace{0.1cm}
%	\includegraphics[width=0.2405\linewidth]{figures/D15_img_facebook_0095.jpg}
%	\includegraphics[width=0.2405\linewidth]{figures/D16_img_facebook_0095.jpg}
%	\includegraphics[width=0.2405\linewidth]{figures/D23_img_facebook_0095.jpg}
%	\includegraphics[width=0.2405\linewidth]{figures/D25_img_facebook_0095.jpg}\\
%	\vspace{0.1cm}
	\includegraphics[width=0.2405\linewidth]{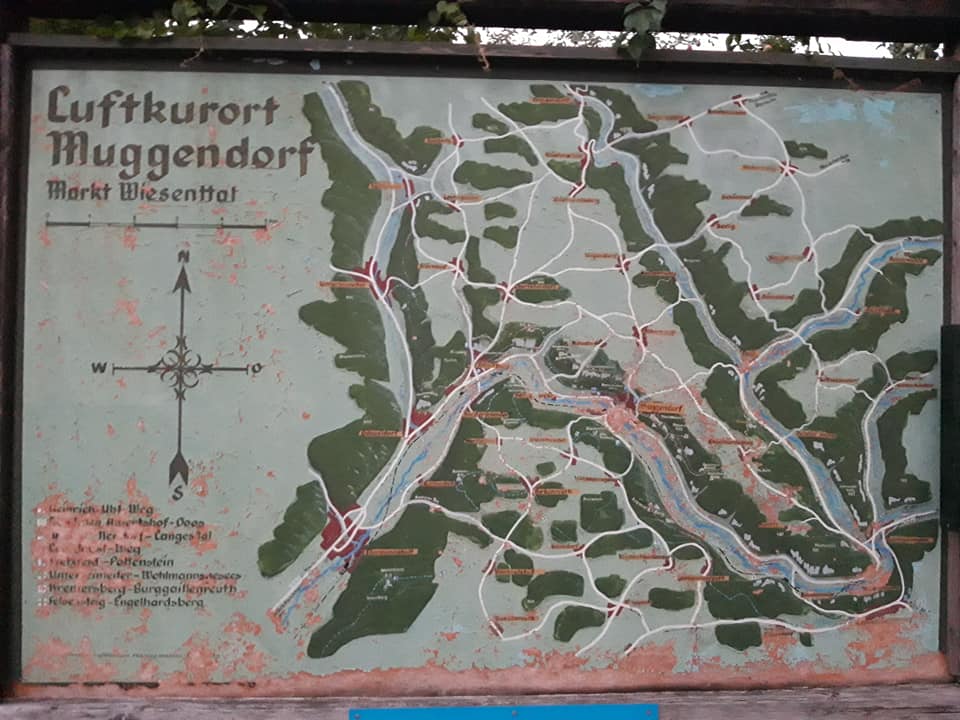}
	\includegraphics[width=0.2405\linewidth]{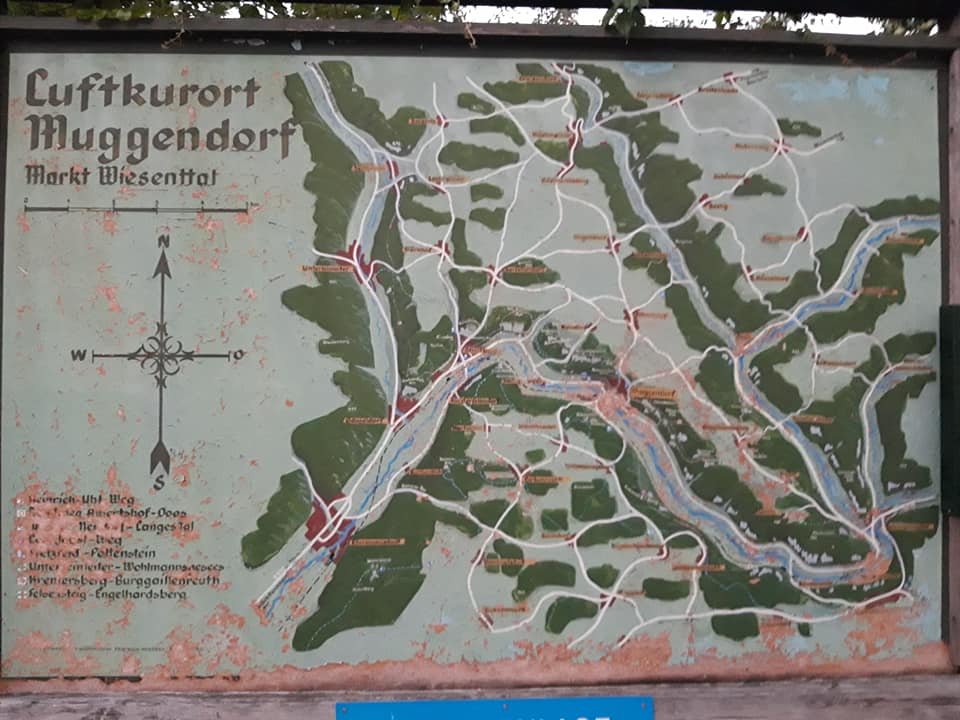}
	\includegraphics[width=0.2405\linewidth]{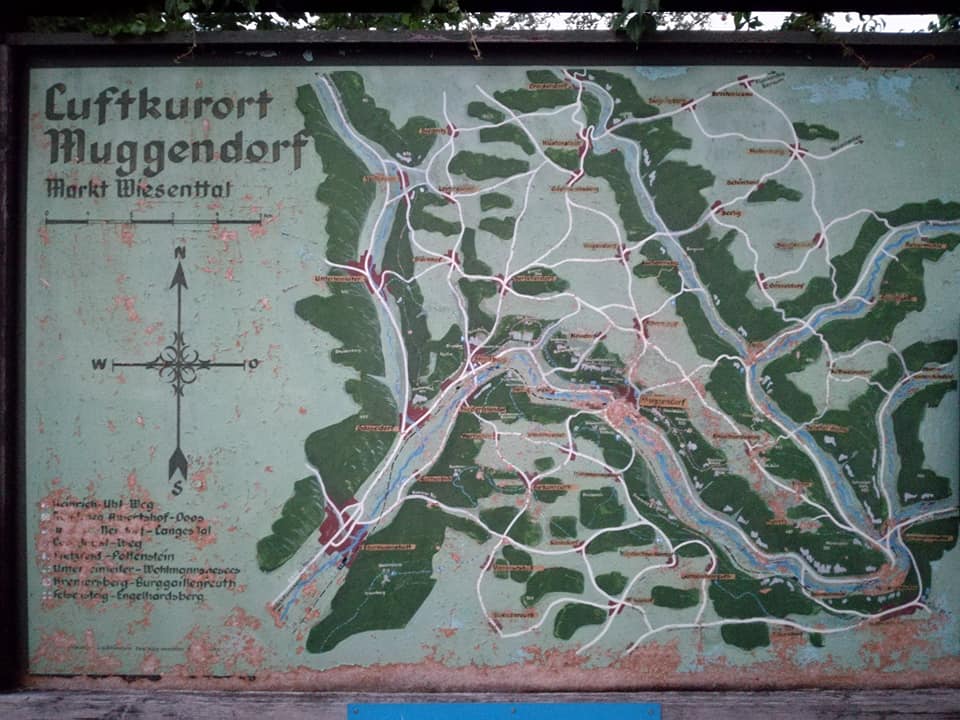}
	\includegraphics[width=0.2405\linewidth]{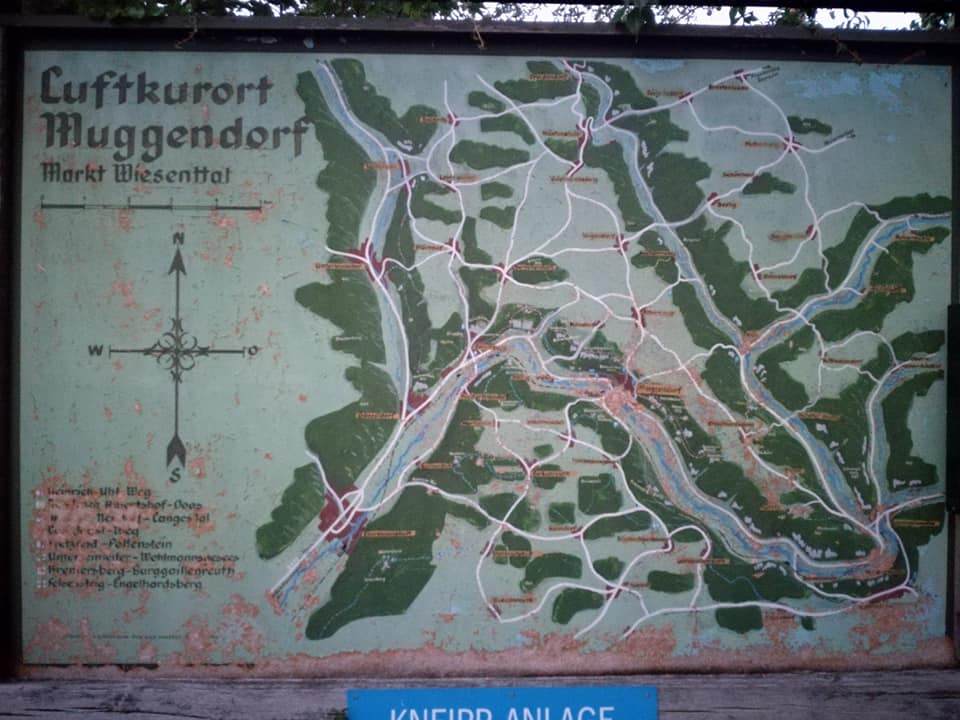}\\	
	\vspace{0.1cm}	
	\includegraphics[width=0.2405\linewidth]{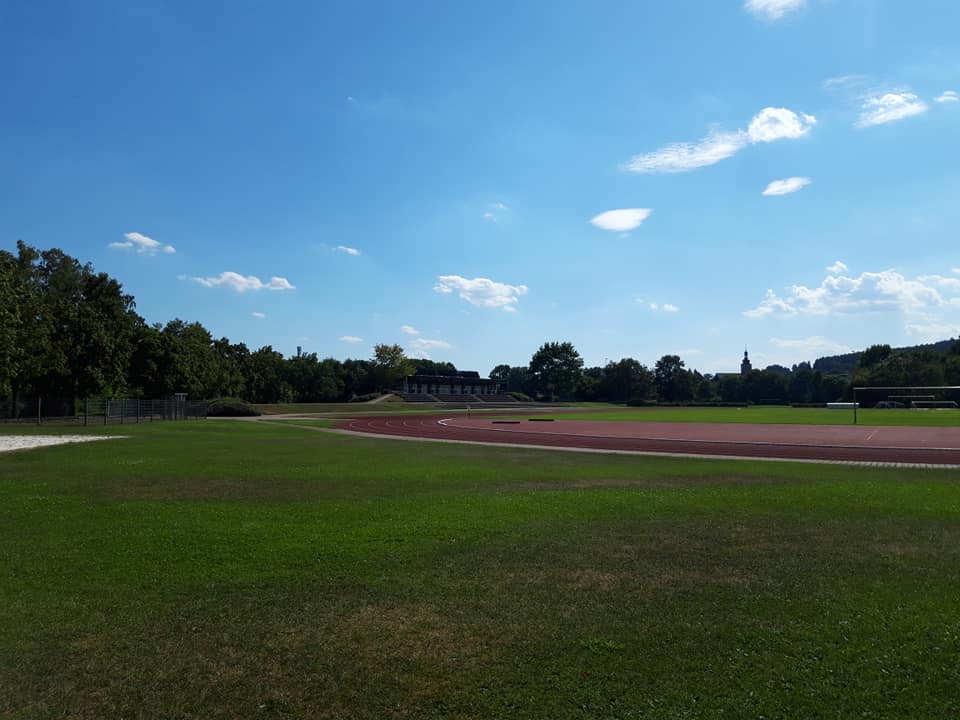}
	\includegraphics[width=0.2405\linewidth]{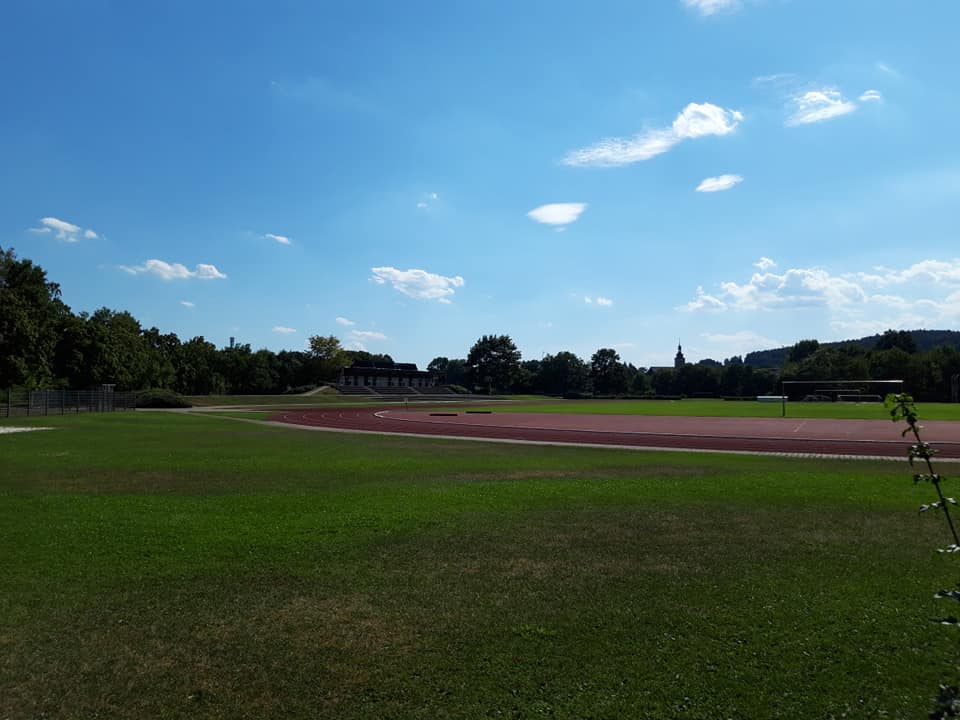}
	\includegraphics[width=0.2405\linewidth]{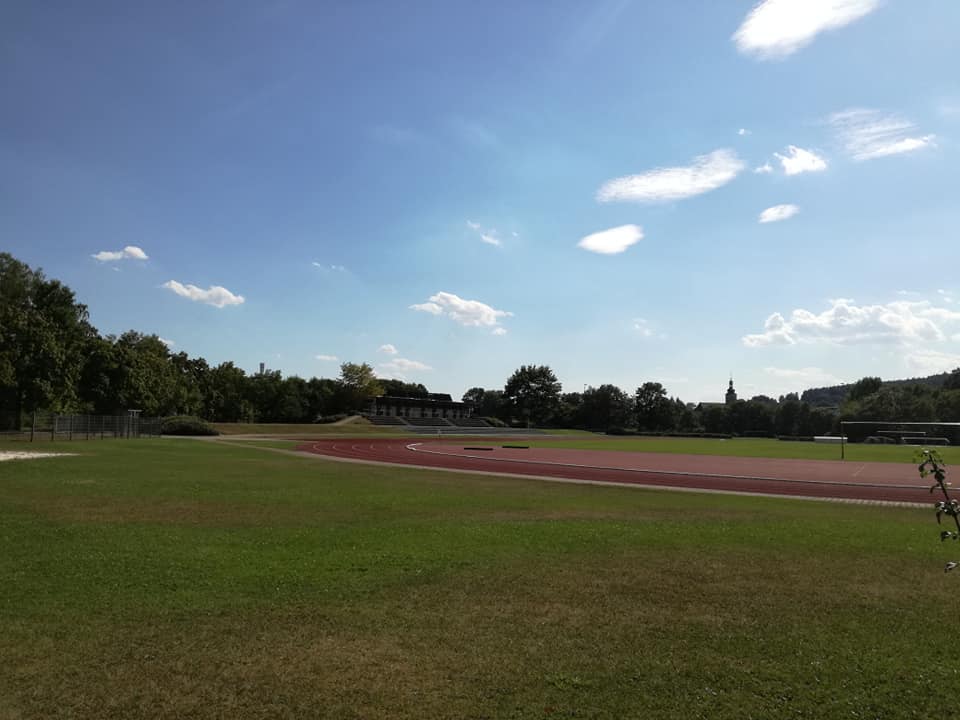}
	\includegraphics[width=0.2405\linewidth]{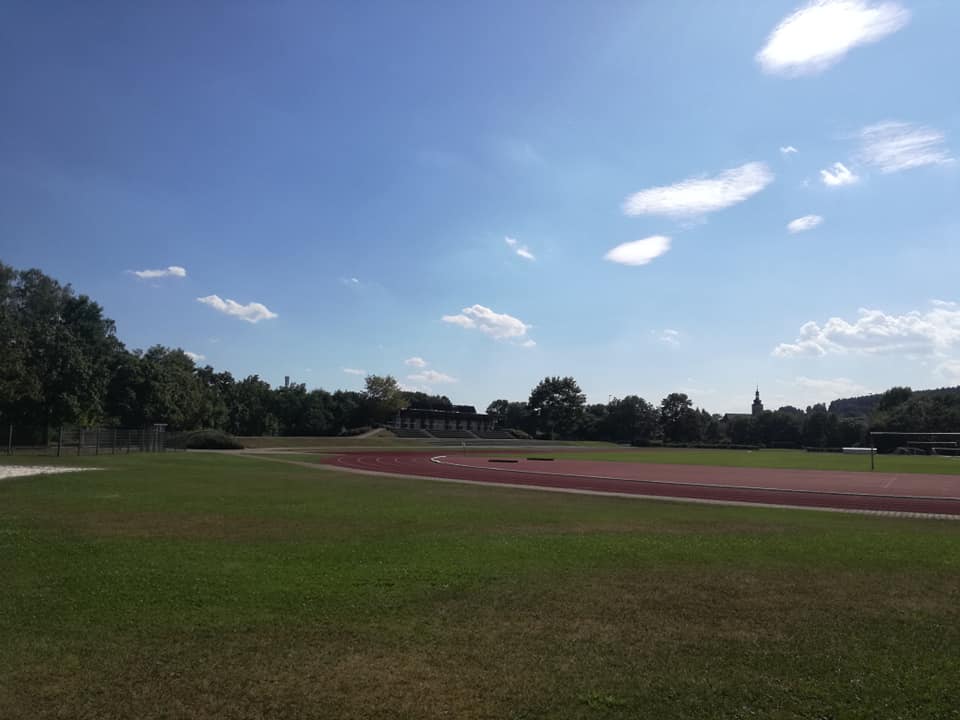}\\
%	\includegraphics[width=0.2405\linewidth]{figures/D15_img_facebook_0060.jpg}
%	\includegraphics[width=0.2405\linewidth]{figures/D16_img_facebook_0060.jpg}
%	\includegraphics[width=0.2405\linewidth]{figures/D15_img_facebook_0066.jpg}
%	\includegraphics[width=0.2405\linewidth]{figures/D16_img_facebook_0066.jpg}\\	
%	\includegraphics[width=0.2405\linewidth]{figures/D23_img_facebook_0060.jpg}
%	\includegraphics[width=0.2405\linewidth]{figures/D25_img_facebook_0060.jpg}
%	\includegraphics[width=0.2405\linewidth]{figures/D23_img_facebook_0066.jpg}
%	\includegraphics[width=0.2405\linewidth]{figures/D25_img_facebook_0066.jpg}\\
%	\vspace{0.1cm}
	\caption{\reedited{Images of the same scene recorded by different devices in FODB~\cite{DBLP:conf/icpr/HadwigerR20}. Scene IDs in FODB from top to bottom: 60, 45, 1.}}
	\label{fig:camera_images}
	\end{subfigure}
%	\vspace{0.1cm}
	\begin{subfigure}[t]{\linewidth}	
	\centering
	\includegraphics[width=0.7\linewidth,trim={0 13.5cm 0 0},clip]{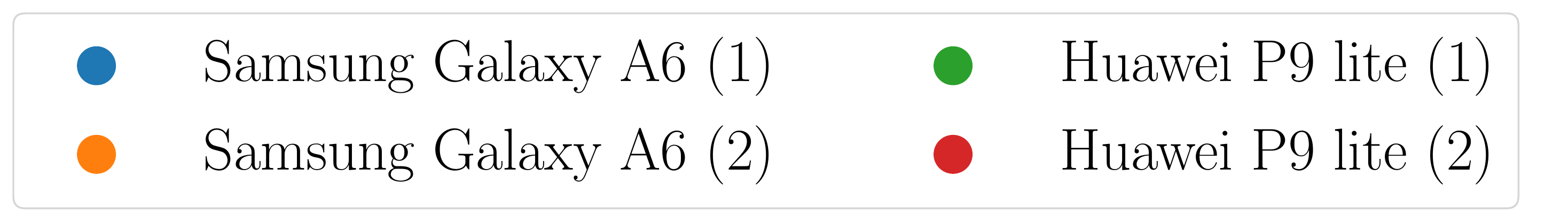}\\	
	\includegraphics[width=0.32\linewidth]{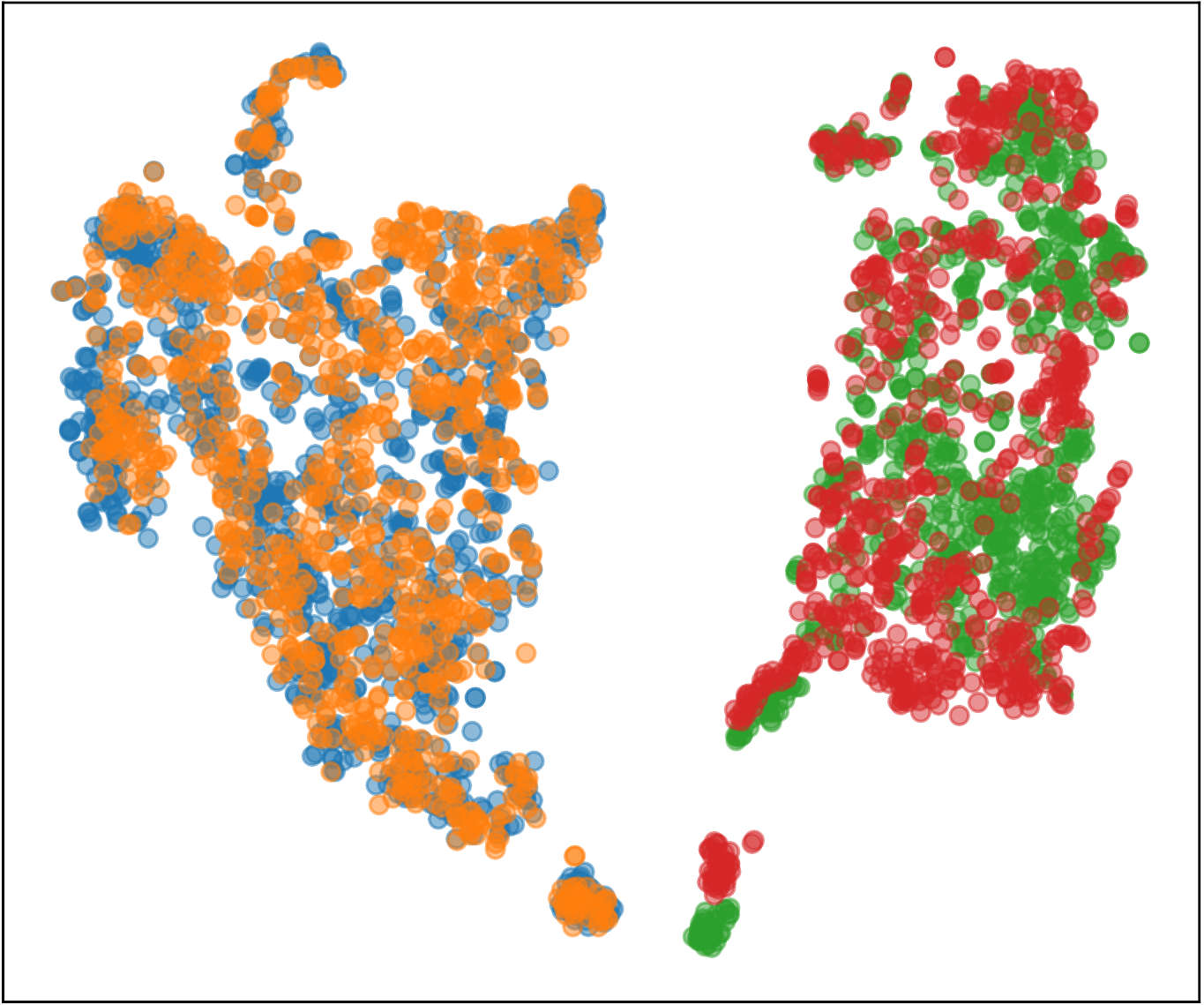}
	\includegraphics[width=0.32\linewidth]{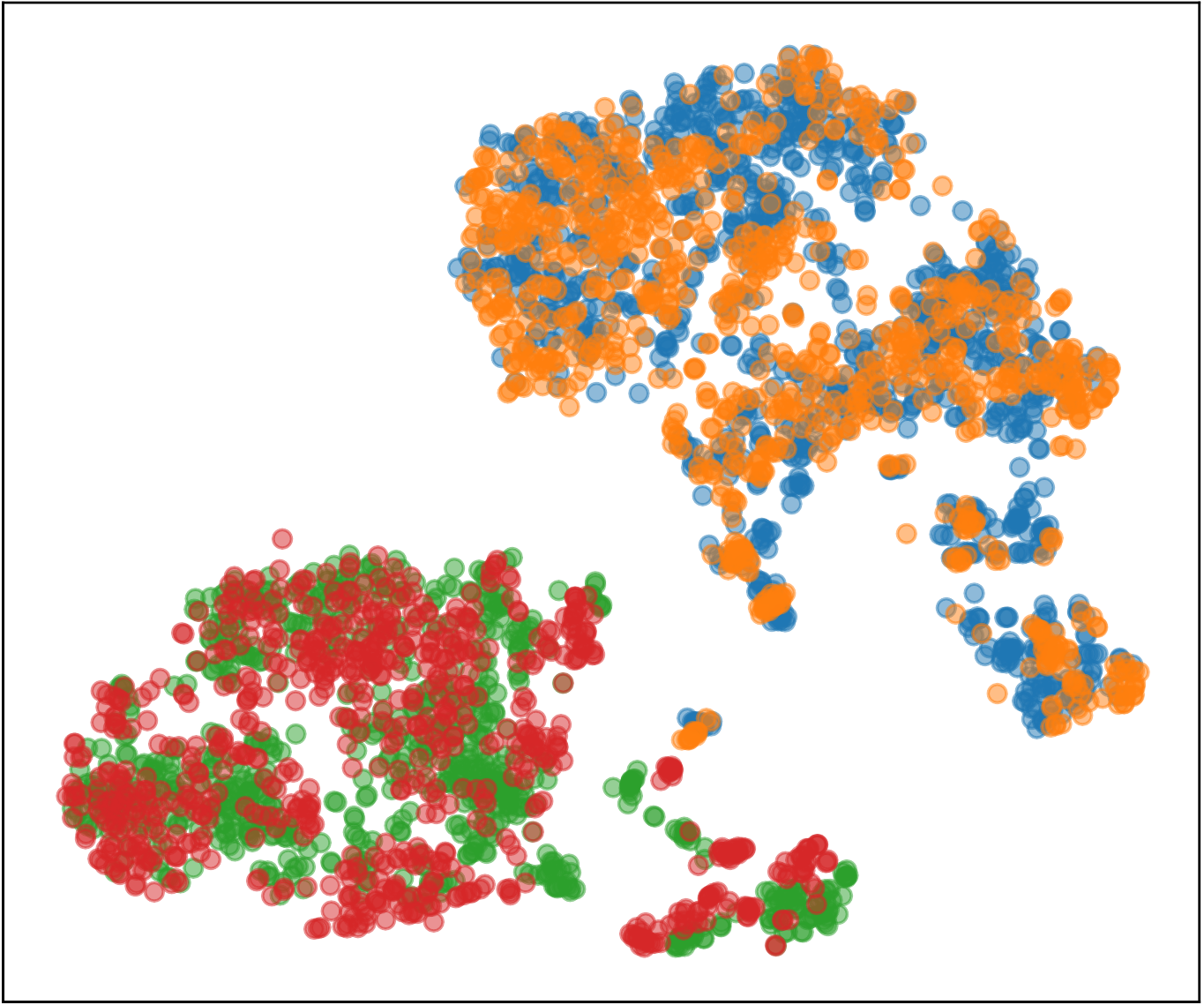}	
	\includegraphics[width=0.32\linewidth]{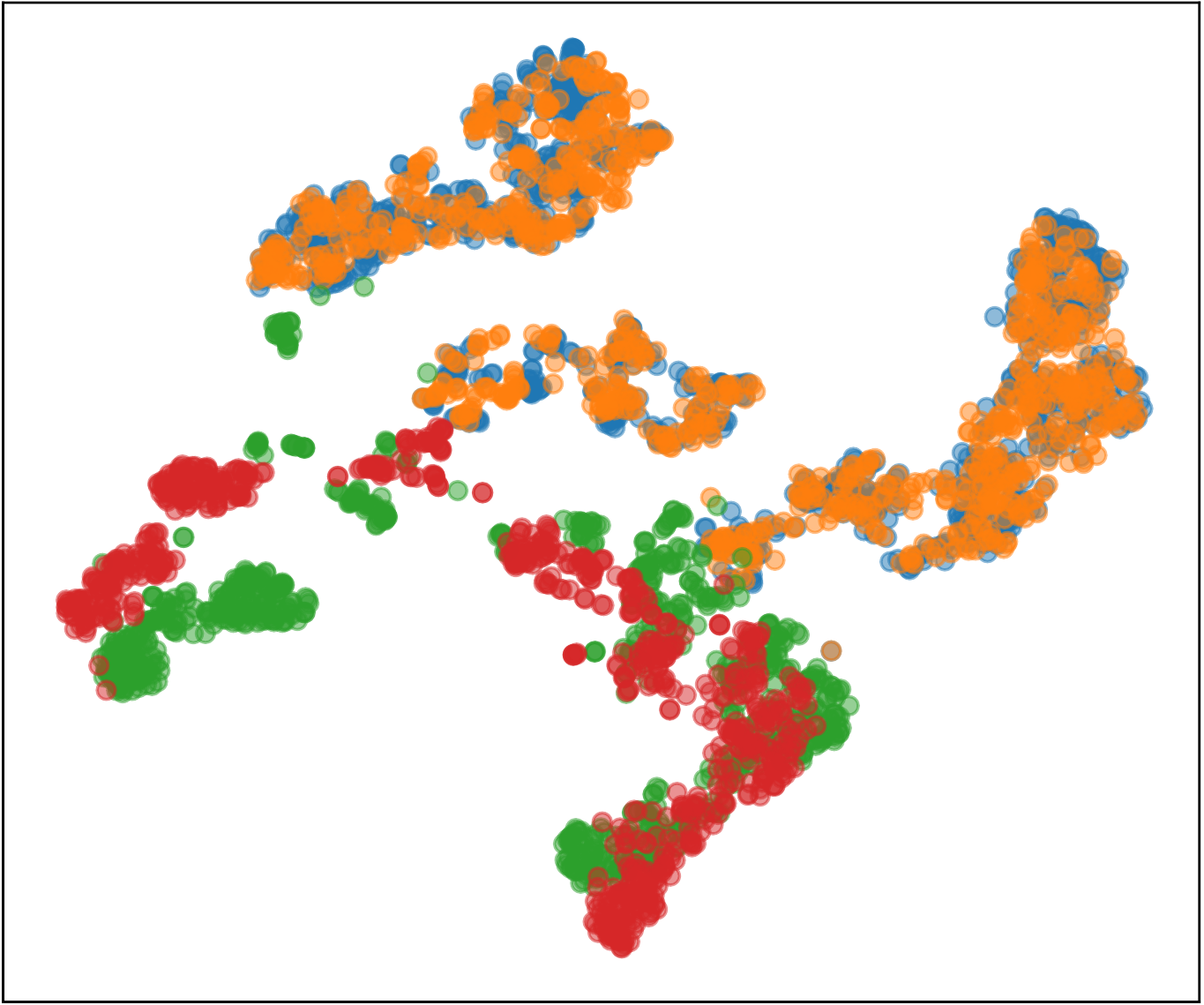}
	\caption{\reedited{Scatter plots of the embeddings for patches extracted from the above images, with one color per device, and one plot (from left to right) per scene, i.e., per row (from top to bottom) in Fig.~\ref{fig:camera_images}.}}
	\label{fig:camera_spread}
	\end{subfigure}
	\caption{\reedited{Illustration of the sensitivity of the learned metric space to camera model traces. a) Each row shows images of the same scene photographed by two devices per model. Images from FODB~\cite{DBLP:conf/icpr/HadwigerR20}. b) Scatter plots of the learned embeddings for patches extracted from the images in (a), with dimensionality reduction by t-SNE~\cite{van2008visualizing}.}}
	\label{fig:camera_experiment}
\end{figure}

\reedited{
We illustrate the sensitivity of the learned embeddings to camera model traces. For this task, we use images from the recent Forchheim Image Database~\cite{DBLP:conf/icpr/HadwigerR20}. This dataset contains images of various scenes, each imaged by different smartphone devices. 
We compare the learned embeddings for images of identical scenes. These scenes are recorded with two Samsung Galaxy A6 devices and two Huawei P9 lite devices, and are shown in Fig.~\ref{fig:camera_images}. Embeddings are calculated for non-overlapping patches from images in \textit{original} quality (confer~\cite{DBLP:conf/icpr/HadwigerR20}), i.e., without any post-processing. %From these images of FODB in \textit{original} quality, we extract non-overlapping patches and compute the learned embeddings. 
Then, we apply \mbox{t-SNE}~\cite{van2008visualizing} to visualize the high-dimensional embeddings in 2-D, which is shown in Fig.~\ref{fig:camera_spread}.

Throughout these scenes, the embeddings from devices of the same model tend to cluster, while embeddings from different models are to some extent separated. This distribution indicates that the learned metric space exhibits sensitivity to camera \textit{model} traces, while largely suppressing \textit{device}-specific variations. This is in agreement with our color imaging model leveraged for training, since color formation is in general common to all devices from a model, which underlines the relation of our approach to camera identification. 
}
%We further performed a quantitative evaluation for camera identification, which is not reported in this paper. However, we found that the performance achieved for the learned embeddings as features is not competitive with specifically designed camera identification features. This can be attributed to two facts. First, our approach is trained to focus on robust low-frequent traces, while most descriptive features for camera identification are high-frequent~\cite{DBLP:conf/sswmc/LukasFG06, DBLP:journals/tifs/CozzolinoV20}. Further, First, we synthesize training data by finishing raw images, which neglects certain properties of real-world camera pipelines. This opens up several directions for future work.}

\section{Conclusion}
\label{sec:conclusion}

We propose a method to learn versatile metric color embeddings that
characterize the color consistency of images.
Manipulated images can then be exposed by locating image regions with large
distances in the learned metric space.
We experimentally demonstrate the robustness against the two frequently
encountered post-processing measures resizing and JPEG recompression.
We further show that the learned embeddings contain information
transferable to illuminant estimation and information characterizing discrepancies of camera pipelines.

The proposed method is particularly well-suited for low quality images that
have been subject to heavy post-processing. In this case, the color-based
analysis is barely affected by the loss of high-frequency image content that is
central to many other forensic algorithms. As such, we hope that the method can
complement a forensic tool set for cases where other algorithms are difficult
to apply.

The limitations of our approach include cases where the color variations within
pristine parts of images are large, manipulation types that do not affect
color distributions, \edited{or forgeries involving image-global color adaptations that do not introduce spatial inconsistencies}. 

%\begin{itemize}
%	\item Discuss results: for HQ images, statistical approaches perform better, for LQ images better use color-based approaches. 
%	\item Discuss limitations, mention future work packages
%\end{itemize}

%\appendices
%\section{Proof of the First Zonklar Equation}
%Appendix one text goes here.

%% you can choose not to have a title for an appendix
%% if you want by leaving the argument blank
%\section{}
%Appendix two text goes here.

% use section* for acknowledgment
\section*{Acknowledgment}

This material is based on research sponsored by the Air Force Research
Laboratory and the Defense Advanced Research Projects Agency under agreement
number FA8750-16-2-0204. The U.S. Government is authorized to reproduce and
distribute reprints for Governmental purposes notwithstanding any copyright
notation thereon.

The views and conclusions contained herein are those of the authors and should
not be interpreted as necessarily representing the official policies or
endorsements, either expressed or implied, of the Air Force Research Laboratory
and the Defense Advanced Research Projects Agency or the U.S. Government.

% Can use something like this to put references on a page
% by themselves when using endfloat and the captionsoff option.
\ifCLASSOPTIONcaptionsoff
  \newpage
\fi

% trigger a \newpage just before the given reference
% number - used to balance the columns on the last page
% adjust value as needed - may need to be readjusted if
% the document is modified later
%\IEEEtriggeratref{8}
% The "triggered" command can be changed if desired:
%\IEEEtriggercmd{\enlargethispage{-5in}}

% references section

% can use a bibliography generated by BibTeX as a .bbl file
% BibTeX documentation can be easily obtained at:
% http://mirror.ctan.org/biblio/bibtex/contrib/doc/
% The IEEEtran BibTeX style support page is at:
% http://www.michaelshell.org/tex/ieeetran/bibtex/
\bibliographystyle{IEEEtran}
% argument is your BibTeX string definitions and bibliography database(s)
\bibliography{IEEEabrv,references}
%\bibliography{references}
%
% <OR> manually copy in the resultant .bbl file
% set second argument of \begin to the number of references
% (used to reserve space for the reference number labels box)
%\begin{thebibliography}{1}

%\bibitem{IEEEhowto:kopka}
%H.~Kopka and P.~W. Daly, \emph{A Guide to \LaTeX}, 3rd~ed.\hskip 1em plus
%  0.5em minus 0.4em\relax Harlow, England: Addison-Wesley, 1999.
%
%\end{thebibliography}

% biography section
% 
% If you have an EPS/PDF photo (graphicx package needed) extra braces are
% needed around the contents of the optional argument to biography to prevent
% the LaTeX parser from getting confused when it sees the complicated
% \includegraphics command within an optional argument. (You could create
% your own custom macro containing the \includegraphics command to make things
% simpler here.)
%\begin{IEEEbiography}[{\includegraphics[width=1in,height=1.25in,clip,keepaspectratio]{mshell}}]{Michael Shell}
% or if you just want to reserve a space for a photo:

\vspace{-0.1cm}

\begin{IEEEbiography}[{\includegraphics[width=1in,height=1.25in,clip,keepaspectratio]{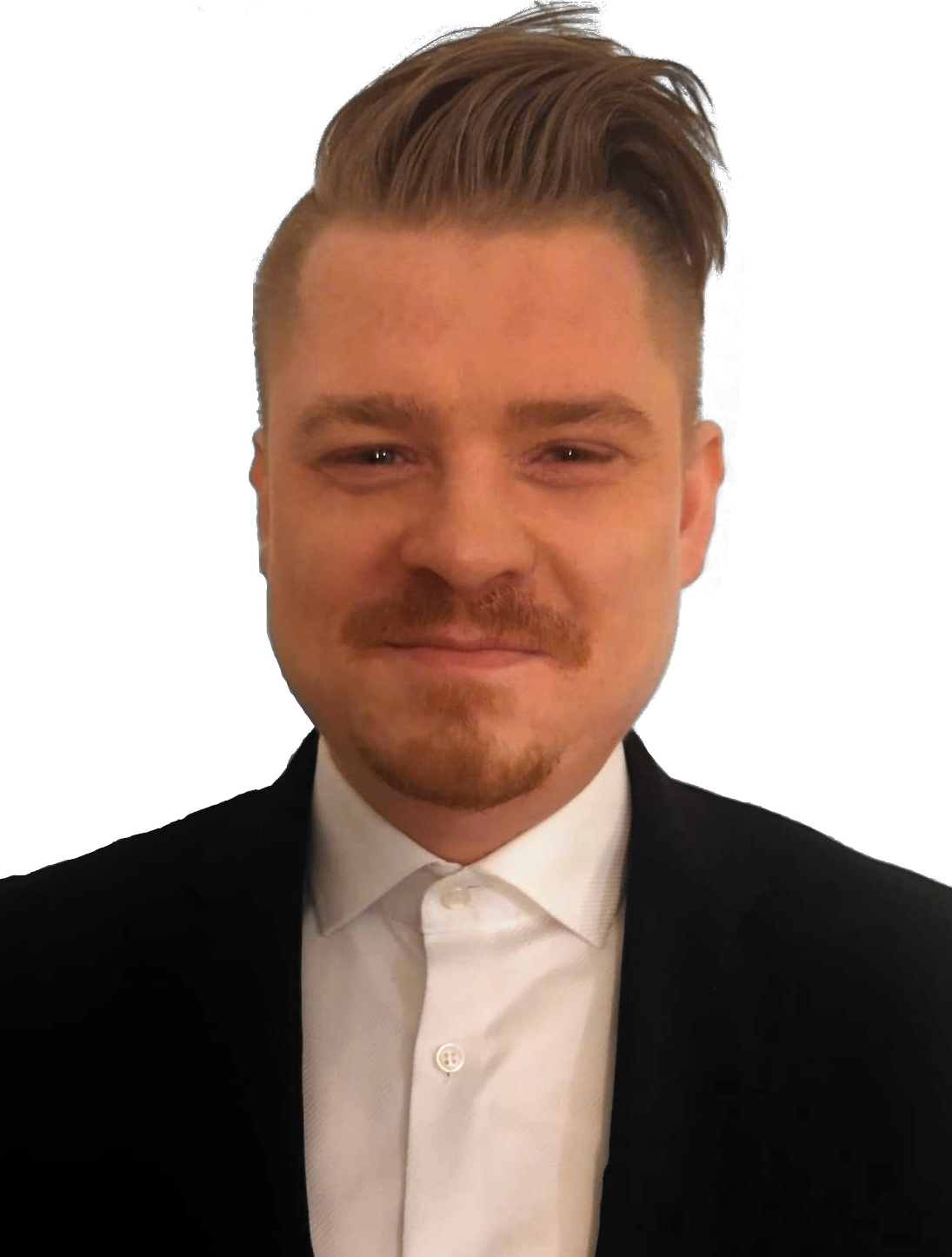}}]{Benjamin Hadwiger}
received the B.Eng. degree in electrical engineering from the N{\"u}rnberg Institute of Technology Georg Simon Ohm, N{\"u}rnberg, Germany, in 2014. In 2016, he received the M.Sc.\ degree in computational engineering from the Friedrich-Alexander University Erlangen-N{\"u}rnberg (FAU), Erlangen, Germany. In 2021, he received the Ph.D. degree in computer science from FAU. His research interests include image forensics, computer vision and machine learning.
\end{IEEEbiography}
%
%%\begin{IEEEbiographynophoto}{Marc Stamminger}
%%Biography text here.
%%\end{IEEEbiographynophoto}
%
%\begin{IEEEbiography}[{\includegraphics[width=1in,height=1.25in,clip,keepaspectratio]{authors/Marc_small}}]{Marc Stamminger}
%is a full professor for Visual Computing at the 
%Friedrich-Alexander University Erlangen-N{\"u}rnberg. His research covers 
%various areas of visual computing, in particular real-time rendering and 
%visualization, 3d reconstruction of static and dynamic objects, and 
%virtual and augmented reality, as well as image forensics. He is 
%associate editor of the Journal on Computer Graphics Tools, was co-chair 
%of the program committee of Eurographics 2009 and the Eurographics 
%Symposium on Rendering in 2010, and member of the program committee of 
%all major computer graphics conferences, such as Siggraph, Siggraph 
%Asia, Eurographics, ACM I3D, or ACM High Performance Graphics. Marc 
%Stamminger is member of the Eurographics Executive Committee and 
%currently serves as the Eurographics Professional Board chair.
%\end{IEEEbiography}
%
\begin{IEEEbiography}[{\includegraphics[width=1in,height=1.25in,clip,keepaspectratio]{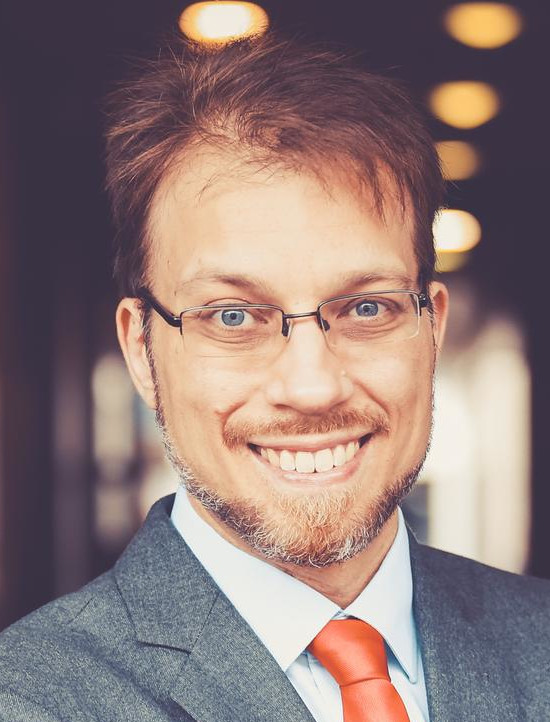}}]{Christian Riess}
received the Ph.D. degree in computer science from the Friedrich-Alexander
University Erlangen-N{\"u}rnberg (FAU), Erlangen, Germany, in 2012.  From 2013
to 2015, he was a Postdoc at the Radiological Sciences Laboratory, Stanford
University, Stanford, CA, USA. Since 2015, he is the head of the Phase-Contrast
X-ray Group at the Pattern Recognition Laboratory at FAU. Since 2016, he is
senior researcher and head of the Multimedia Security Group at the IT
Infrastructures Lab at FAU, and he received his habilitation from FAU in 2020.
He served on the IEEE Information Forensics and Security Technical Committee
2017-2019, and received the IEEE Signal Processing Award in 2017.  His research
interests include all aspects of image processing and imaging, particularly
with applications in image and video forensics, X-ray phase contrast, and
computer vision.
\end{IEEEbiography}

% You can push biographies down or up by placing
% a \vfill before or after them. The appropriate
% use of \vfill depends on what kind of text is
% on the last page and whether or not the columns
% are being equalized.

\vfill

% Can be used to pull up biographies so that the bottom of the last one
% is flush with the other column.
%\enlargethispage{-5in}

% that's all folks
\end{document}